\documentclass[preprint,12pt]{elsarticle}

\usepackage{amssymb}
\usepackage{url}
\usepackage{subfigure}
\usepackage{amsmath}
\usepackage{algorithm}
\usepackage{algorithmic}
\usepackage{epstopdf}
\usepackage{bm}
\usepackage{cases}
\usepackage{subeqnarray}
\usepackage{threeparttable}
\usepackage{amsfonts}
\usepackage{cases}
\usepackage[dvipsone]{epsfig}
\usepackage{graphicx}
\usepackage{booktabs}
\usepackage{multirow}
\usepackage{booktabs}

\usepackage{graphics}
\usepackage{color}
\usepackage{amsfonts}
\usepackage{multirow}
\usepackage{amsmath}
\usepackage{cases}       
\usepackage{graphicx}
\usepackage{subfigure}
\usepackage{url}
\usepackage{lineno}
\usepackage{amssymb}
\usepackage{booktabs}
\usepackage{threeparttable}
\usepackage{epstopdf}
\usepackage{bm}
\usepackage{geometry}
\usepackage{txfonts}
\usepackage{float}

\biboptions{sort&compress}

\newtheorem{theorem}{Theorem}

\begin{document}

\begin{frontmatter}



\title{Single and union non-parallel support vector machine frameworks}{}


\author[1]{Chun-Na Li}
\address[1]{Management School, Hainan University, Haikou, 570228, P.R.China}

\author[1]{Yuan-Hai Shao \corref{cor}}

\author[2]{Huajun Wang}
\address[2]{School of Science, Beijing Jiaotong University, Beijing, 100044, P.R.China}

\author[3]{Yu-Ting Zhao}
\address[3]{School of Economics, Hainan University, Haikou, 570228, P.R.China}

\author[3]{Ling-Wei Huang}

\author[2]{Naihua Xiu}

\author[4]{Nai-Yang Deng}
\address[4]{College of Science, China Agricultural University, Beijing, 100083, P.R.China}

\cortext[cor]{Corresponding author.}

\begin{abstract}
Considering the classification problem, we summarize the nonparallel support vector machines with the nonparallel hyperplanes to two types of frameworks. The first type constructs the hyperplanes separately. It solves a series of small optimization problems to obtain a series of hyperplanes, but is hard to measure the loss of each sample. The other type constructs all the hyperplanes simultaneously, and it solves one big optimization problem with the ascertained loss of each sample. We give the characteristics of each framework and compare them carefully. In addition, based on the second framework, we construct a max-min distance-based nonparallel support vector machine for multiclass classification problem, called NSVM. It constructs hyperplanes with large distance margin by solving an optimization problem. Experimental results on benchmark data sets show the advantages of our NSVM.
\end{abstract}

\end{frontmatter}



\section{Introduction}
For binary classification problem, the generalized eigenvalue proximal support vector machine (GEPSVM) was proposed by Mangasarian and Wild \cite{GEPSVM} in 2006, which is the first nonparallel support vector machine. It aims at generating two nonparallel hyperplanes such that each hyperplane is closer to its own class and as far as possible from the other class. GEPSVM is effective, particularly when dealing with the ``Xor"-type data \cite{GEPSVM}. This leads to extensive studies on nonparallel support vector machines (NSVMs) \cite{TWSVMbook17}.

\begin{figure}[htbp]
\centering
\subfigure[Original data]{
\label{Fig.sub.2}
\includegraphics[width=0.18\textwidth]{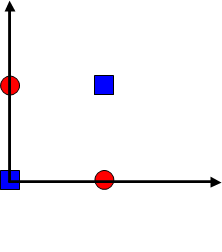}}
\subfigure[NN (1)]{
\label{Fig.sub.2}
\includegraphics[width=0.20\textwidth]{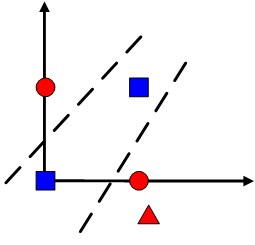}}
\subfigure[NN (2)]{
\label{Fig.sub.1}
\includegraphics[width=0.21\textwidth]{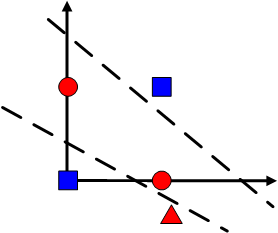}}
\subfigure[GEPSVM]{
\label{Fig.sub.1}
\includegraphics[width=0.21\textwidth]{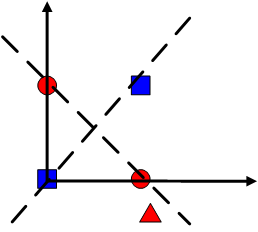}}
\caption{Classification results of two-layer NN and GEPSVM on a ``Xor" data set.}
\label{fig1}
\end{figure}

To see the advantage of the characteristic of NSVMs on ``Xor"-type data, we now consider a simplest two-class ``Xor" data set with one class containing two red circle samples, the other class containing two blue square samples, as shown in Fig.\ref{fig1}(a). Further, suppose there is a new red triangular sample which belongs to the red circle class as plotted in Fig.\ref{fig1}(b). Obviously, this is a typical "Xor" data that cannot be separately linearly, and one can resort to two layer neural networks (NN) \cite{Nitta03}, but classical support vector machine (SVM) \cite{SVM,SVMbook12} will not work.
By applying the two-layer neural networks with some ReLU activation \cite{LiYuang17}, we obtain the classification result in Fig.\ref{fig1} (b). However, by choosing a different initial weighter vector of two-layer NN, we may obtain an essentially different separating result, as presented in Fig.\ref{fig1} (c). This phenomenon will affect the generalization performance of NN on such type of data. In fact, by observing Fig.\ref{fig1} (b) and Fig.\ref{fig1} (c), we already seen opposite NN predicting results for the new triangular sample.
We now apply GEPSVM on this data, and the classification result is given in Fig. \ref{fig1}(d). We see GEPSVM generates two nonparallel hyperplanes such that each hyperplane is close to one of the class and at the same time far away from the other class. Different from NN, GEPSVM classifies these two classes from a proximal point of view with stable generalization performance.

To now, there are several dozens of NSVMs models. However, there are little studies on general forms and characteristics of NSVMs. In this paper, by observing the existing NSVMs, we find that they can be mainly categorized into two types.
The first type constructs two hyperplanes separately, where a series of small optimization problems are solved. There are many NSVMs belong to this type, including GEPSVM.
In GEPSVM, the optimization problems were reduced to generalized eigenvalue problems, and its improvement IGEPSVM \cite{IGEPSVM} replaced them by two standard eigenvalue problems. Following GEPSVM, twin support vector machine (TWSVM) \cite{TWSVM,TBSVM} solved two small-scale QPPs, each of them was similar to that of SVM. Nonparallel support vector machine (NPSVM) \cite{NPSVM} and improved NPSVM \cite{INPSVM} extended TWSVM by considering sparseness. Best fitting hyperplane classifier (BFHC) \cite{BFHC} used the ramp distance to measure the dissimilarity of two hyperplanes. On the other hand, L1NPSVM \cite{L1NPSVM} considered the L1-norm-based GEPSVM, and $L_1$-GEPSVM \cite{L1GEPSVM} and LpNPSVM \cite{LpNPSVM} imposed the $L_2$-norm and $L_p$-norm ($p>0$) regularization terms on L1NPSVM, respectively. More NSVMs of this type also include least squares twin support vector machines (LSTWSVM) \cite{LSTSVM,ILSTSVM}, TWSVM with the pinball loss (Pin-TWSVM) \cite{PinTWSVM}, Ramp loss TWSVM \cite{RNPSVM}, etc.

For the above NSVMs, there is a common characteristic that each hyperplane is constructed separately. In contrast, a different type of NSVMs finds all hyperplanes simultaneously by some union models. A typical representation of this second type of NSVMs is
nonparallel hyperplane support vector machine (NHSVM) \cite{NHSVM} that solved one single quadratic programming problem to construct two nonparallel hyperplanes in binary classification. Robust nonparallel hyperplane SVM (RNH-SVM)\cite{NHSVMSOC} extended NHSVM to second-order cones, and its least square version was studied in \cite{LSNHSVM}. Another representation is the proximal classifier with consistency (PCC) \cite{PCC}, which is also the extension of the GEPSVM. Different from GEPSVM, PCC was based on comparing two distances between a point and two hyperplanes.
Compared to the first type of NSVMs, the second type of NSVMs has the advantage that each sample has an ascertained loss \cite{NHSVM}. However, it needs to solve a large scale optimization problem and hence is time consuming.



We should note that most of the above NSVMs focus on the binary classification problem. In fact, NSVMs for binary classification problem have been also extended to many powerful variants and applied in many machine learning problems, including multiclass NSVMs \cite{MBSVM13,BSTWSVM13,MNHSVM15,MLSTMSVMDAG16,MTSVM16,
MIBSVM17,WLMBSVM17,MILSTWSVM18,RastogiSaigal18,SaigalRastogi19,LiuGong20}, projection NSVMs \cite{MVSVM11,PTSVM11,RLSPTSVM12,RPTSVM13,RLSPTSVMn14,MPTSVM16,RichhariyaTanveer20},
regression \cite{Peng10,ShaoZhangYang13,BalasundaramTanveer13,YangHuaShao16,WangShiNiu17,
YeShaoDeng17,LiuShaoWang18,TangTianYang18,Carrasco19},
semi-supervised learning \cite{QiTianShi12,ChenShaoDeng14,SunXie16,YangXu16,RastogiPal19}, clustering \cite{WangShaoBai15,KhemchandaniPal16,YeZhaoNaiem17,YeZhaoLi18,WangChenLi18,BaiShaoWang19}, multilabel learning \cite{ChenShaoLi16,Azad-Manjiri19}, tensor learning
\cite{KhemchandaniKarpatne13,ZhaoShiLv14,ShiZhaoJing14,ShiZhaoZhen16,XiangJiangHe18},
and multiview problem \cite{XieSun14,TangLiTian18,SunXieDong18,Xie18,XieSun20}. However, all of the above extensions of NSVMs are similar to those of SVM-type models. For example, for multiclass classification problem, the ``1-vs-1" \cite{Kressel99}, ``1-vs-rest" \cite{Vapnik98} or other SVM strategies \cite{PlattCristianini00,CheongOh04} are usually considered in multiclass NSVMs. 
This indicates these extensions do not use the intrinsic features of NSVMs sufficiently.

This paper is concerned with the NSVMs for multiclass classification problem. Specifically, we have the following contributions:

(i) We first summarize the existing NSVMs into two types, and further establish two frameworks: \textit{The framework of Single Models} and \textit{The framework of Union Models}. The former one constructs each hyperplane separately, and the latter one constructs all the hyperplanes simultaneously in a union model. We also summarize the characteristics of these two frameworks.



(ii) Based on the second type framework, a new max-min distance-based nonparallel support vector machine (NSVM) is proposed, where the loss function is carefully introduced. It not only could separate different classes well, it also could capture the structure of each class effectively.


(iii) The primal problem of NSVM is effectively solved through a modified proximal difference-of-convex algorithm with extrapolation algorithm (MpDCA$_{\hbox{e}}$).

(iv) NSVM can be easily extended to its nonlinear version, and the nonlinear algorithm requires only inner products of data samples. Therefore, the kernel trick is applicable and the algorithm is computationally efficient.

(v) Experimental results on an artificial data set and some benchmark data sets show the advantages of our NSVM.
%
%
%
%

%

%
%


The rest of this paper is organized as follows. Two types of frameworks of NSVMs are described in Section \ref{secframeworks}. Based on the second framework, we propose a max-min distance-based NSVM in Section \ref{DNSVM}. Experiments and conclusions are arranged in Sections \ref{experiments} and \ref{conclusion}, respectively. The proof of the main theorem is presented in the Appendix.

\section{Frameworks of NSVMs}\label{secframeworks}

\subsection{Problem formulation}

We consider the following general multiclass classification problem: given the training data set $T$ with $m$ samples
\begin{align}\label{Dataset}
\begin{array}{cl}
T=\{(x_i,y_i)|i=1,...,m\},
 \end{array}
\end{align}
where $x_i\in \mathcal{X}$ and $y_i \in \mathcal{Y}$, $i=1,...,m$, $\mathcal{X}$ is the input space $\mathbb{R}^n$, and $\mathcal{Y}$ is the output space $\{1,2,...,K\}$.
The goal of the multiclass classification problem is to deduce the output $y\in\mathcal{Y}$ of any given $x\in\mathcal{X}$ from the training set $T$.

In NSVMs, they try to find $K$ hyperplanes
\begin{align}\label{khyperplanes}
\begin{array}{cl}
g(x;w_{y},b_y)=w_y^{\top}x +b_y=0,y=1,... ,K,
 \end{array}
\end{align}
such that for any $x\in\mathbb{R}^n$, its output can be deduced from the values $g(x;w_{y},b_y)$, $y=1,... ,K$. Here, $w_y\in\mathbb{R}^n$ and $b_y\in\mathbb{R}$ are parameters depending on $y\in\{1,... ,K\}$, and $(\cdot)^{\top}$ is the transposition of a matrix.
NSVMs require the samples of the $y$-th class to be close to their corresponding $y$-th hyperplane and far away from other hyperplanes. This leads to the results that most NSVMs have the following characteristics: maximizing some inter-class distance, and at the same time minimizing some  intra-class distance.

Here we first give some notations. Sometimes we write $g(x;w_y,b_y)$ as $g_y(x)$ for simplicity. Under certain circumstances, we also may extend the definition of $g(x;w_y,b_y)$ from a vector $x$ to a matrix $Z$, that is, $g_y(Z)=g(Z;w_{y},b_y)=w_y^{\top} Z +b_ye$,
where $Z\in\mathbb{R}^{n\times m_{_Z}}$ is the data matrix that is composed of $m_{_Z}$ vectors of dimension $n$.  The symbol $e$ is the column vector of all ones of appropriate dimension, and $\textbf{0}$ represents the vector of zeros of an appropriate dimension. $\|\cdot\|_p$ means the $L_p$-norm for $p>0$. When $p=2$, for brevity, we write $\|\cdot\|_2$ as $\|\cdot\|$. The absolute value operation $|\cdot|$ acts on a vector componentwisely.


The task of the rest of this section is to establish the unified frameworks, respectively for the first type and the second type of NSVMs, namely \textit{The framework of Single Models} and \textit{The framework of Union Models}.

\subsection{The framework of Single Models}

Firstly, we consider NSVMs which construct each hyperplane separately.
Let $X=(x_1,x_2,\ldots,x_m)\in\mathbb{R}^{n\times m}$ be the corresponding data matrix of the training set $T$, and we further reorganize the training data as $X=[X_1,...,X_K]$, where $X_y \in R^{n \times m_{y}}$ is the $y$-th class input matrix, $m_{y}$ is the number of the samples in the $y$-th class and $\sum_{y=1}^{K}m_{y}=m$, $y=1,\ldots,K$.
Then, we have \textit{The framework of Single Models} as in Model 1.

\noindent
\rule{15cm}{1pt}\\
\noindent{\bf Model 1}. \textit{The framework of Single Models}.\\
\rule{15cm}{0.6pt}\\
\noindent{\bf Input:} Data $X=[X_1,...,X_K]$, a new coming sample $x$.\\
\noindent{\bf Process:}\\
\textbf{1. Training:}\\

For $y=1,\ldots,K$, solve the following $K$ problems:
\begin{align}\label{UNSVM1}
\begin{split}
\underset{w_y,b_y}{\min} &~\frac{1}{2}\|g_y\|_\mathcal{F}^2+C_1 \underline{\mathcal{D}}(g_{y}(X_y),0)-C_2 \sum_{j\neq y}{\overline{\mathcal{D}}}( {g_y}(X_j),g_y(X_y)),\\
\end{split}
\end{align}
where $\|g_y\|_\mathcal{F}^2:=\|\big(\begin{smallmatrix}
        w_{y} \\
        b_{y} \\
\end{smallmatrix}\big)\|_\mathcal{F}^2$ or $\|w_y\|_\mathcal{F}^2$ or 0, is the induced norm of $g_y(x)$ in the functional space $\mathcal{F}$, $y,j=1,\ldots,K$, $C_1$ and $C_2$ are regularization parameters. $\underline{\mathcal{D}}(\underline{u},\underline{v}):=\mathbb{R}^{n_{\underline{u}}\times n_{\underline{v}}}\rightarrow \mathbb{R}$ and $\overline{\mathcal{D}}(\overline{u},\overline{v}):=\mathbb{R}^{n_{\overline{u}}\times n_{\overline{v}}}\rightarrow \mathbb{R}$ are respectively the intra-class distance and the inter-class distance, where $n_{\underline{u}}$ and $n_{\underline{v}}$ are respectively the dimensions of vectors $\underline{u}$ and $\underline{v}$, and $n_{\overline{u}}$ and $n_{\overline{v}}$ are respectively the dimensions of vectors $\overline{u}$ and $\overline{v}$.

Obtain the solutions $[w_1;b_1]$, $[w_2;b_2],\ldots,[w_K;b_K]$ for the above $K$ problems.\\
\textbf{2. Prediction:}\\
\quad The label of $x$ is given by
$$ \textrm{label} (x) =\underset{y\in\{1,2,\ldots,K\}}{\arg\min} \mathcal{D}(g_y(x)), $$
where $\mathcal{D}(\cdot)$ is the distance from $x$ to the $y$-th hyperplane
$g_y(x)=0$.\\
\rule{15cm}{1pt}\\

Note that the functional space $\mathcal{F}$ can be a reproducing kernel Hilbert space (RKHS) induced by some kernel if we construct a kernel machine. Also, different distances can be used for  $\underline{\mathcal{D}}(\cdot)$, $\overline{\mathcal{D}}(\cdot)$ and $\mathcal{D}(\cdot)$, such as the $L_2$-norm distance, the $L_1$-norm distance or other kernel distances.
Now we list some representative NSVMs that belong to this model.

\subsubsection{GEPSVM}

The first nonparallel support vector machine is GEPSVM \cite{GEPSVM} for binary classification problem, and it formulates as
\begin{align}\label{GEPSVM11}
 \underset{(w_{1},b_{1})\neq 0}{\min}~
 \frac{\|w_{1}^{\top}X_{1}+eb_{1}\|^{2}+\frac{\delta}{2}\|\left(
\begin{smallmatrix}
        w_{1} \\
        b_{1} \\
\end{smallmatrix}
\right)\|^{2}}{\|w_{1}^{\top}X_{2}+eb_{1}\|^{2}}
\end{align}
and
\begin{align}\label{GEPSVM22}
 \underset{(w_{2},b_{2})\neq 0}{\min}~
 \frac{\|w_{2}^{\top}X_{2}+eb_{2}\|^{2}+\frac{\delta}{2}\|\left(
 \begin{smallmatrix}
        w_{2} \\
        b_{2} \\
\end{smallmatrix}
\right)\|^{2}}{\|w_{2}^{\top}X_{1}+eb_{2}\|^{2}},
\end{align}
where $\delta$ is a positive regularization parameter.

Set
\begin{equation}
\begin{cases}
\displaystyle \|g_y\|_\mathcal{F}^2= \|\left(
\begin{smallmatrix}
        w_{y} \\
        b_{y} \\
\end{smallmatrix}
\right)\|^{2},\\
\displaystyle \underline{\mathcal{D}}(g_{y}(X_y),0)= \|g_{y}(X_y)\|^2,\\
\displaystyle {\overline{\mathcal{D}}}( {g}_y(X_j),g_y(X_y))=\|g_y(X_j)\|^2,\\
\mathcal{D}(g_y(x))=\frac{|g_y(x)|}{\|w_{y}\|},\\
C_{1}= \frac{1}{\delta},C_{2}=\frac{1}{\delta},
\end{cases}
\end{equation}
where $y,j=1,2,~y\not=j$.
Then, \eqref{GEPSVM11} and \eqref{GEPSVM22} fall into framework \eqref{UNSVM1} up to $\frac{\delta}{\|g_{y}(X_j)\|^2}$.

That is to say, GEPSVM is a special case of \textit{The framework of Single Models}. By setting $y,j=1,2,\cdots,K,$ and $y\not=j$, then we obtain a multiclass classification GEPSVM model.

Note that all distances $\underline{\mathcal{D}}$, $\overline{\mathcal{D}}$, $\mathcal{D}$ and the regularization term are based on the $L_2$-norm distance. There are also some modified GEPSVM models that use the $L_1$-norm distance \cite{L1NPSVM} or the $L_p$-norm distance \cite{LpNPSVM}.

\begin{table*}[htbp]
\begin{center}
\caption{NSVMs of \textit{The framework of Single Models}.}
\resizebox{6in}{!}
{
\begin{tabular}{l|ccccccccccc}
\toprule
Method &$\|g_y\|_\mathcal{F}^2$  &  $\mathcal{\underline{D}}(g_{y}(X_y),0)$&  $\overline{\mathcal{D}}( {g_y}(X_j),g_y(X_y))$ &$\mathcal{D}(g_y(x))$&Distance\\
\hline
  GEPSVM\cite{GEPSVM,IGEPSVM}  & $\|\left(
\begin{smallmatrix}
        w_{y} \\
        b_{y} \\
\end{smallmatrix}
\right)\|^{2}$ &  $\|g_{y}(X_y)\|^2$ &  $ \|{g_y}(X_j)\|^2$ & $\frac{|g_y(x)|}{\|w_{y}\|}$&   $\|\cdot\|$ \\
\hline
L1NPSVM \cite{L1NPSVM}&--- &  $\|g_{y}(X_y)\|_1$&$\|{g_y}(X_j)\|_1$ &$|g_y(x)|$& $\|\cdot\|_1$ \\
\hline
LpNPSVM \cite{LpNPSVM} &$\|w_y\|_p$ &  $\|g_{y}(X_y)\|_1$&  $\|{g_y}(X_j)\|_1$ & $\frac{|g_y(x)|}{\|w_{y}\|}$ & $\|\cdot\|_p/\|\cdot\|_1$ \\
\hline
TW/BSVM \cite{TWSVM,TBSVM}   & $\|\left(
\begin{smallmatrix}
        w_{y} \\
        b_{y} \\
\end{smallmatrix}
\right)\|^{2}$ &  $\|g_{y}(X_y)\|^2$&  $-e^{\top}\cdot(g_{y}(X_j)-e)_{+}$ &  $\frac{|g_y(x)|}{\|w_{y}\|}$  & $\|\cdot\|$ / $(\cdot)_+$ \\
  \hline
LSTSVM \cite{LSTSVM,ILSTSVM} & $\|\left(
\begin{smallmatrix}
        w_{y} \\
        b_{y} \\
\end{smallmatrix}
\right)\|^{2}$ &  $\|g_{y}(X_y)\|^2$&  $ -\|g_{y}(X_j)-e \|^2$ & $|g_y(x)|$  &   $\|\cdot\|$ \\
  \hline
Pin-TWSVM \cite{PinTWSVM}  &  $\|w_y\|^2$&  $L_{\tau}(X_y,y,g_y(X_y))^*$ &  $ \|g_{y}(X_j)\|_1$& $\frac{|g_y(x)|}{\|w_{y}\|}$  &  $\|\cdot\|$ / $L_{\tau}(\cdot)$/$\|\cdot\|_1$\\
  \hline
NPSVM \cite{NPSVM}&  $\|w_y\|^2$ & $\|g_{y}(X_y)-\epsilon e\|_1$&  $-e^{\top}\cdot(g_{y}(X_j)-e)_{+}$ & $|g_y(x)|$ & $\|\cdot\|$ / $\|\cdot\|_1$/$(\cdot)_+$\\
\hline
RNPSVM \cite{RNPSVM}  &  $\|w_y\|^2$&  $e^{\top}\cdot R_{\epsilon,t}(g_{y}(X_y))$$^{*}$&
   $ -e^{\top}\cdot R_{s}(-g_{y}(X_j))$$^{*}$&  $|g_y(x)|$ &  $\|\cdot\|$/$R_{\epsilon,t}(\cdot)$/$R_{s}(\cdot)$ \\
  \hline
BFHC \cite{BFHC} &  $\|w_y\|^2$ & $e^{\top}\cdot L_{pos}(g_{y}(X_y))^*$  &  $-e^{\top}\cdot L_{neg}(g_{y}(X_j))^*$ & $\frac{|g_y(x)|}{\|w_{y}\|}$  & $\|\cdot\|$ / $L_{pos}(\cdot)$/$L_{neg}(\cdot)$\\
\hline
{NSVM \cite{NSVM2018} } & $\|w_y\|^2$ &  $l_{\epsilon}(g_{y}(X_y))^*$ &  $-e^{\top}\cdot(|g_{y}(X_j)|-e)_+$ & $\frac{|g_y(x)|}{\|w_{y}\|}$ & $\|\cdot\|$/$l_{\epsilon}(\cdot)$/$(\cdot)_+$\\
\hline
{AVIPC \cite{AVIPC2014} }  & $\|w_y\|_1$  &  $\|g_{y}(X_y)\|_1$ & $\|g_{y}(X_j)\|_1$ & $|g_y(x)|$ &  $|\cdot|$ \\
\bottomrule
\end{tabular}
}
\label{Firsttypetable}
\end{center}
\scriptsize{$^{*}$
$L_{\tau}(X_y,y,g_y(X_y))=\begin{cases}e^{\top}(0-yg_{y}(X_y)),~&0-yg_{y}(X_y)\geq0,\\
-\tau e^{\top}(yg_{y}(X_y)-0),&0-yg_{y}(X_y)<0,\end{cases}$
$R_{\epsilon,t}(z)=\begin{cases}t-\epsilon,&|z|>t\\
|z|-\epsilon,&\epsilon\leq|z|\leq t\\0,&|z|<\epsilon,\end{cases}$,
$R_{s}(z)=\begin{cases}0,~~~~~~z>1\\1-z,~s\leq z\leq 1,\\1-s,~~z<s,\end{cases}$
$l_{\epsilon}(z)=\begin{cases}1-|z|,&|z|\leq 1\\0,&1< |z|\leq 1+\epsilon\\|z|-1-\epsilon,&otherwise,\end{cases}$,
$L_{pos}(z)=\min\{1-s,\max\{0,-1+\Delta-z\}\}+\min\{1-s,\max\{0,-1+\Delta+z\}\}$, $L_{neg}(z)=\min\{1+\Delta-s,\max\{0,1+\Delta-z\}\}+\min\{1+\Delta-s,\max\{0,1+\Delta+z\}\}$,
where $-1<s\leq0$ is a parameter, and $\Delta\geq0$ is a user defined constant.}
\end{table*}

\subsubsection{TWSVM}

Twin (bounded) support vector machine (TW/BSVM) \cite{TWSVM,TBSVM} is one of the most applied NSVMs. Its formulation is considered as
\begin{align}\label{TWSVM1}
\begin{array}{cl}
 \underset{w_{1},b_{1},\xi}{\min} & \frac{\lambda_1}{2}\|\left(
\begin{smallmatrix}
        w_{1} \\
        b_{1} \\
\end{smallmatrix}
\right)\|^{2}+ \frac{1}{2}\|w_{1}^{\top}X_{1}+eb_{1}\|^2+\delta_{1}e^{\top}\xi\\
 \mathrm{s.t.\ }  &-(w_{1}^{\top}X_{2}+eb_{1})+\xi \geq e, \xi \geq 0
 \end{array}
\end{align}
and
\begin{align} \label{TWSVM2}
\begin{array}{cl}
 \underset{w_{2},b_{2}, \eta}{\min}& \frac{\lambda_2}{2}\|\left(
\begin{smallmatrix}
        w_{2} \\
        b_{2} \\
\end{smallmatrix}
\right)\|^{2}+ \frac{1}{2}\|w_{2}^{\top}X_{2}+eb_{2}\|^2+\delta_{2}e^{\top}\eta\\
 \mathrm{s.t.\ } & (w_{2}^{\top}X_{1}+eb_{2})+\eta\geq e, \eta \geq0,
 \end{array}
\end{align}
where $\delta_1$ and $\delta_{2}$ are positive parameters.

Set
\begin{equation}
\begin{cases}
\displaystyle \|g_y\|_\mathcal{F}^2=  \|\left(
\begin{smallmatrix}
        w_{y} \\
        b_{y} \\
\end{smallmatrix}
\right)\|^{2},\\
\displaystyle \underline{\mathcal{D}}(g_{y}(X_y),0)= \|g_{y}(X_y)\|^2,\\
\displaystyle {\overline{\mathcal{D}}}({g}_y(X_j),g_y(X_y))= -e^{\top}(g_{y}(X_j)-e)_{+},\\
\mathcal{D}(g_y(x))=\frac{|g_y(x)|}{\|w_{y}\|},\\
C_{1}= \frac{1}{2},C_{2}=\delta_y,
\end{cases}
\end{equation}
in framework \eqref{UNSVM1}, it becomes \eqref{TWSVM1} and \eqref{TWSVM2}.
In TWSVM, the distances for $\underline{\mathcal{D}}$ and $\overline{\mathcal{D}}$ are the $L_2$-norm distance and the $(\cdot)_+$ distance; for decision, the distance $\mathcal{D}$ is the absolute value distance.

TWSVM is a typical case of \textit{The framework of Single Models}. By setting $y,j=1,2,\cdots,K,$ and $y\not=j$, we also obtain a multiclass classification TWSVM model. There are many variants of TWSVM with different distances $\underline{\mathcal{D}}$, $\overline{\mathcal{D}}$ and $\mathcal{D}$. More details could be found in \cite{vTWSVM10,TPMSM11,CDMTSVM12,WLTSVM12,WLSTSVM15}.

\subsubsection{LSTSVM}

Least square twin support vector machine (LSTSVM) \cite{LSTSVM} has the following form
\begin{align}\label{LSTSVM1}
\begin{array}{cl}
 \underset{w_{1},b_{1},\xi}{\min} &  \frac{\lambda_1}{2}\|\left(
\begin{smallmatrix}
        w_{1} \\
        b_{1} \\
\end{smallmatrix}
\right)\|^{2}+
\frac{1}{2}\|w_{1}^{\top}X_{1}+eb_{1}\|^2+ \frac{\delta_1}{2}\xi^{\top}\xi\\
 \mathrm{s.t.\ }  &-(w_{1}^{\top}X_{2}+e_{2}b_{1})+\xi = e
 \end{array}
\end{align}
and
\begin{align} \label{LSTSVM2}
\begin{array}{cl}
 \underset{w_{2},b_{2},\eta}{\min}& \frac{\lambda_2}{2}\|\left(
\begin{smallmatrix}
        w_{2} \\
        b_{2} \\
\end{smallmatrix}
\right)\|^{2}
+\frac{1}{2}\|w_{2}^{\top}X_{2}+eb_{2}\|^2+\frac{\delta_2}{2}\eta^{\top}\eta\\
 \mathrm{s.t.\ } & (w_{2}^{\top}X_{1}+e_{1}b_{2})+\eta = e,
 \end{array}
\end{align}
where $\delta_1$ and $\delta_{2}$ are positive parameters.

Set
\begin{equation}
\begin{cases}
\displaystyle \|g_y\|_\mathcal{F}^2= \|\left(
\begin{smallmatrix}
        w_{y} \\
        b_{y} \\
\end{smallmatrix}
\right)\|^{2},\\
\displaystyle \underline{\mathcal{D}}(g_{y}(X_y),0)= \|g_{y}(X_y)\|^2,\\
\displaystyle {\overline{\mathcal{D}}}({g}_y(X_j),g_y(X_y))= -\|g_{y}(X_j)-e\|^2,\\
\mathcal{D}(g_y(x))=|g_y(x)|,\\
C_{1}= \frac{1}{2},C_{2}=\frac{1}{2}\delta_y,
\end{cases}
\end{equation}
for $y,j=1,2,~y\not=j$ in framework \eqref{UNSVM1}.
%
Then, \eqref{LSTSVM1} and \eqref{LSTSVM2} fit into \textit{The framework of Single Models} \eqref{UNSVM1}.
In LSTSVM, the distances for $\underline{\mathcal{D}}$ and $\overline{\mathcal{D}}$ are the $L_2$-norm distances; for decision, the distance $\mathcal{D}$ is the absolute value distance. By setting $y,j=1,2,\cdots,K,$ and $y\not=j$, then we also obtain a multiclass classification LSTSVM model.

There are also many other single type NSVM models such as \cite{L1NPSVM,LpNPSVM,RNPSVM,BFHC,NSVM2018,AVIPC2014,FanLuXLi16,WangZhangChoi19}. We skip their descriptions and list some of them in Table \ref{Firsttypetable}.
In summary, single models have the following characteristics:

(i) Their $K$ hyperplanes are constructed by $K$ single problems separately. The $y$-th problem tries to minimize the distances from the samples of the $y$-class to the $y$-th hyperplane, and maximize the distances from the samples not in the $y$-class to the $y$-th hyperplane, $y=1,2,\ldots,K$.
Generally, each of the $K$ problems corresponds to one hyperplane and may be easy to solve, or each subproblem has a small scale.

(ii) The distance metric used for $\underline{\mathcal{D}}(g_{y}(X_y),0)$, $\overline{\mathcal{D}}( {g_y}(X_j),g_y(X_y))$ and $\mathcal{D}(g_y(x))$ is flexible and could be different from each other.

However, there are also some disadvantages of single models:

(i) In the training procedure of \textit{The framework of Single Models}, the distances from samples of different classes to one hyperplane are characterized. However, in the predicting procedure, the label of a new coming sample is determined by the distances from this sample to different hyperplanes. That is to say, the training and predicting procedures are inconsistent.

(ii) Due to the first type NSVMs characterize the distances from one class to its center hyperplane by using all samples together and the first drawback, it is hard to define a loss function for each training sample.

\subsection{The framework of Union Models}

Rather than constructing each hyperplane separately, the second type NSVMs construct $K$ hyperplanes at the same time, yielding the union model.
The union model could measure the distances between each sample to $K$ hyperplanes together, and measure the misclassification rate for each sample. \textit{The framework of Union Models} could be written as the following:

\noindent
\rule{15cm}{1pt}\\
\noindent{\bf Model 2}. \textit{The framework of Union Models}.\\
\rule{15cm}{0.6pt}\\
\noindent{\bf Input:} Data set $T=\{(x_i,y_i)|i=1,...,m\}$, a new coming sample $x$.\\
\noindent{\bf Process:}\\
\textbf{1. Training:}

Solve the following union problem:
\begin{align}\label{UNSVM2}
\begin{split}
\underset{w_1,b_1,\ldots,w_K,b_K}{\min} &~\frac{1}{2}\sum_{y=1}^K\|g_y\|_\mathcal{F}^2+C_1 \sum_{i=1}^{m}\underline{\mathcal{D}}(g_{y_i}(x_i),0)\\
&-C_2 \sum_{i=1}^{m}\sum\limits_{j\neq y_{i}}{\overline{\mathcal{D}}}( {g_j}(x_i),g_{y_{i}}(x_i)).
\end{split}
\end{align}
Obtain the solution $(w_1;b_1)$, $(w_2;b_2),\ldots,(w_K;b_K)$ for the above problem.\\
\textbf{2. Prediction:}\\
\quad The label of $x$ is given by
$$ \textrm{label} (x) =\underset{y\in\{1,2,\ldots,K\}}{\arg\min} \mathcal{D}(g_y(x)).$$
\rule{15cm}{1pt}\\

Proximal classifier with consistency (PCC) \cite{PCC} and nonparallel hyperplane SVM (NHSVM) \cite{NHSVM} are two representative NSVMs belonging to this model.

\begin{table*}[htbp]
\begin{center}
\caption{NSVMs of \textit{The framework of Union Models}.}
\resizebox{6in}{!}
{
\begin{tabular}{l|ccccccccccc}
\toprule
Method & $\|g_y\|_\mathcal{F}^2$ &  ${\mathcal{\underline{D}}}(g_{y_i}(x_i),0)$ &  ${\overline{\mathcal{D}}}( {g_j}(x_i),g_{y_i}(x_i))$ &$\mathcal{D}(g_{y}(x))$& Distance \\
   \hline
  {PCC\cite{PCC}}  &   $\|\left(
\begin{smallmatrix}
        w_{y} \\
        b_{y} \\
\end{smallmatrix}
\right)\|^{2}$  & $ \frac{\|g_{y_i}(x_{i})\|^2}{\|u_{y_i}\|}$ &  $ \frac{\|g_{j}(x_{i})\|^2}{\|u_{j}\|}$ & $\frac{|g_y(x)|}{\|u_{y}\|}$ & $\|\cdot\|$  \\
  \hline
  NHSVM \cite{NHSVM}  & $\|\left(
\begin{smallmatrix}
        w_{y} \\
        b_{y} \\
\end{smallmatrix}
\right)\|^{2}$ & $\|g_{y_i}(x_i)\|^2$&$(g_{y_i}(x_i)-g_j(x_i)-1)_+$ &$\frac{|g_y(x)|}{\|w_{y}\|}$  & $\|\cdot\|$/$(\cdot)_+$\\
  \hline
 RNH-SVM \cite{NHSVMSOC} & $\|\left(
\begin{smallmatrix}
        w_{y} \\
        b_{y} \\
\end{smallmatrix}
\right)\|^{2}$& $\|g_{y_i}(x_i)\|^2$& $\text{Pr}\{x_i\in\{x_i: |g_{y_i}(x_i)-g_j(x_i)|\geq 1\}\}^*$& $\frac{|g_y(x)|}{\|w_{y}\|}$ & $\|\cdot\|$/$(\cdot)_+$  \\
  \hline
{LSNHSVM \cite{LSNHSVM}} &  $\|\left(
\begin{smallmatrix}
        w_{y} \\
        b_{y} \\
\end{smallmatrix}
\right)\|^{2}$&  $\|g_{y_i}(x_{i})\|^2$ &  $\|g_{j}(x_i)-g_{y_i}(x_i)-e\|^2$ & $\frac{|g_y(x)|}{\|w_{y}\|}$ & $\|\cdot\|$ \\
\bottomrule
\end{tabular}
}
\label{Secondtypetable}
\end{center}
\scriptsize{~$^{*}$
$\text{Pr}\{\cdot\}$ is the probability.}
\end{table*}

\subsubsection{PCC}

The formulation of PCC \cite{PCC} is
\begin{align}\label{PCC1}
\begin{array}{cl}
\underset{w_{1},b_{1},w_{2},b_{2}}{\min}
&{\sum_{j=1}\limits^{N_1}\left((w_1^{\top}x_{1j}+b_1)^2
+\delta (w_1^2+b_1^2)\right)}/{(w_1^2+b_1^2)}\\
&-\nu{\sum_{j=1}\limits^{N_1}\left((w_2^{\top}x_{1j}+b_2)^2
+\delta (w_2^2+b_2^2)\right)}/{(w_2^2+b_2^2)}\\
&+{\sum_{j=1}\limits^{N_2}\left((w_2^{\top}x_{2j}+b_2)^2
+\delta (w_2^2+b_2^2)\right)}{(w_2^2+b_2^2)}\\
&-\nu{\sum_{j=1}\limits^{N_2}\left((w_1^{\top}x_{2j}+b_1)^2
+\delta (w_1^2+b_1^2)\right)}{(w_1^2+b_1^2)},\\
\end{array}
\end{align}
where $u_y=\left(
\begin{smallmatrix}
        w_{y} \\
        b_{y} \\
\end{smallmatrix}
\right)$, $y=1,2$, $\nu$ and $\delta$ are positive parameters.

Set
\begin{equation}
\begin{cases}
\displaystyle \|g_y\|_\mathcal{F}^2=\|u_{y}\|^{2},\\
\displaystyle \underline{\mathcal{D}}(g_{y_i}(x_i),0)= \frac{\|g_{y_i}(x_i)\|^2}{\|u_{y_i}\|},\\
\displaystyle {\overline{\mathcal{D}}}({g}_j(x_i),g_{y_i}(x_i))= \frac{{\|g_j}(x_i)\|^2}{\|u_{j}\|},\\
\displaystyle \mathcal{D}(g_y(x))=\frac{|g_y(x)|}{\|u_{y}\|},\\
C_{1}= 1, C_{2}=\nu,
\end{cases}
\end{equation}
for $y,j=1,2,~y\not=j$ in framework \eqref{UNSVM2}.
Then, \eqref{PCC1} falls into framework \eqref{UNSVM2}, and PCC is a special case of \textit{The framework of Union Models}. By setting $y,j=1,2,\cdots,K,$ and $y\not=j$, then we obtain a multiclass classification PCC model.

All three distances $\underline{\mathcal{D}}$, $\overline{\mathcal{D}}$, $\mathcal{D}$ and the regularization terms are based on the $L_2$-norm distance.
There are also some extensions of PCC. Local proximal classifier (LPC) \cite{LPC} divided the feature space into positive local regions and negative local regions in PCC. Locality sensitive proximal classifier with consistency (LSPCC) \cite{LSPCC} was also proposed to capture the local geometric structure of the underlying manifold.

\subsubsection{NHSVM}
NHSVM \cite{NHSVM} for binary classification problem could be expressed as
\begin{align}
\begin{array}{lll}\label{NHSVM}
\underset{w_y,b_y,\xi_y}{\min} &\displaystyle \frac{1}{2}\sum_{y=1}^{2}(\|w_y\|^{2}+b_{y}^{2})+\displaystyle \frac{\delta_{1}}{2}\sum_{y=1}^{2}\sum_{j=1}^{N_y}(w_{y}^{\top}x_{yj}+b_{y})^{2}\\
&+\delta_{2}\displaystyle \sum_{y=1}^{2}\sum_{j=1}^{N_y}\xi_{yj}\\
~~\mathrm{s.t.} &w_{1}^{\top}x_{1j}+b_{1}-w_{2}^{\top}x_{1j}-b_{2} \geq 1 -\xi_{1j}, \xi_{1j} \geq 0,\\
& w_{2}^{\top}x_{2j}+b_{2}-w_{1}^{\top}x_{2j}-b_{1}\geq 1-\xi_{2j}, \xi_{2j} \geq 0,
\end{array}
\end{align}
where $\delta_1$ and $\delta_2$ are positive parameters, $x_{yj}$ is the $j$-th sample belonging to the $y$-th class, $y=1,2$.

Set
\begin{equation}
\begin{cases}
\displaystyle \|g_y\|_\mathcal{F}^2=
\|\left(
\begin{smallmatrix}
        w_{y} \\
        b_{y} \\
\end{smallmatrix}
\right)\|^{2},\\
\displaystyle {\overline{\mathcal{D}}}({g}_j(x_i),g_{y_i}(x_i))= (g_{y_i}(x_i)-g_j(x_i)-1)_+,\\
\displaystyle \underline{\mathcal{D}}(g_{y_i}(x_i),0)= \|g_{y_i}(x_i)\|^2,\\
\mathcal{D}(g_y(x))=\frac{|g_y(x)|}{\|w_{y}\|},\\
C_{1}= \frac{1}{2}\delta_1,C_{2}=\delta_2,
\end{cases}
\end{equation}
for $y,j=1,2,~y\not=j$ in \eqref{UNSVM2}.
Then framework \eqref{UNSVM2} becomes \eqref{NHSVM}. In NHSVM, the distances for $\underline{\mathcal{D}}$ and $\overline{\mathcal{D}}$ are the $L_2$-norm distance and $(\cdot)_{+}$ distance; for decision, the distance $\mathcal{D}$ is also the $L_2$-norm distance.
By setting $y,j=1,2,\cdots,K,$ and $y\not=j$, then we obtain a multiclass classification NHSVM union model. There are also other variants of NHSVM with different distances
\cite{NHSVMSOC,LSNHSVM}.

In summary, there are three characteristics of union NSVMs:

(i) Their $K$ hyperplanes are constructed by a single optimization problem by minimizing the distances from the samples of the $y$-class to the $y$-th hyperplane, and maximizing the distances of the samples of the $y$-class to other classes, $y=1,2,\ldots,K$.

(ii) In predicting, a new sample will be assigned to the class that corresponds to its nearest hyperplane. Remember that they also maximize the distances of each sample to different classes. That is to say, the training and predicting procedures are consistent, and it is easy to define a loss function for training samples in union NSVMs.

(iii) It is natural to extend the union model to multiclass classification problem by solving a single optimization problem.

However, the optimization problems of union NSVMs usually have much larger scale than the single ones.
We should also note that there are some overlap between single and union NSVMs. For example, by observing the formulation of PCC, we see PCC can be easily separated into two subproblems, with each subproblem corresponding to one hyperplane. In fact, it was pointed in \cite{PCC} that the solution of PCC is equal to the solutions of these two subproblems. There are also some other combined models on NSVMs \cite{FTSVM}, which combined both the fist type NSVMs and the second type NSVMs.

\section{Max-min distance-based NSVM}\label{DNSVM}


In this section, based on \textit{The framework of Union Models}, we construct a multiclass max-min distance-based nonparallel support vector machine (NSVM), in which each sample is assigned a loss.


\subsection{Max-min distance-based NSVM model}

For multiclass classification problem, we first define the following max-min distance-based loss function for the sample $(x,y)$
\begin{align}\label{NSSVMOloss}
\begin{split}
\ell(x,y)
=~\underset{j\not=y}{\max~}\{\|\langle w_{y}, x \rangle+b_{y} \|^2-\|\langle w_{j}, x \rangle+b_{j}\|^2,0\}.
\end{split}
\end{align}

We now explain the meaning of this loss function. For the $i$-th sample $(x_i,y_i)$, it is reasonable to require its distance to any hyperplane $w_{j}x+b_{j}=0$ ($j\neq y_i$) is greater than its distance to the $y_i$-th hyperplane $w_{y_i}x+b_{y_i}=0$. In other words, it is reasonable to require
$\|\langle w_{i},x_i \rangle+b_{y_i} \|^2<\|\langle w_{j}, x_i \rangle+b_{j}\|^2$. If the inequality is false for some $j\neq y_i$, then the loss $\ell(x_i,y_i)$ will be penalized. Therefore, we in fact maximize the minimum distance of a sample $(x_i,y_i)$ to the hyperplanes it is not belonging to.

%

Let $\displaystyle \frac{1}{2}\|g\|_\mathcal{F}^2=\frac{1}{2}  \displaystyle \sum_{y=1}^{K} \|w_{y}\|^2$, $\displaystyle \sum_{j\neq y_i}{\overline{\mathcal{D}}}( {g_{j}}(x_i),g_{y_{i}}(x_i))=-\ell(x_i,y_i)$ and
$\displaystyle \underline{\mathcal{D}}(g_{y_{i}}(x_{i}),0)=\displaystyle \|g_{y_{i}}(x_i)\|^2$.
Then, we construct our max-min distance-based NSVM as the following
\begin{align}\label{NSVMLM}
\begin{split}
&\underset{w_1,b_1,\ldots,w_K,b_K}{\min}~\frac{1}{2}\|g\|_\mathcal{F}^2+C_1 \sum_{i=1}^{m}\underline{\mathcal{D}}(g_{y_i}(x_i),0)\\
&-C_2 \sum_{i=1}^{m}\displaystyle \sum_{j\neq y_i}{\overline{\mathcal{D}}}( {g_j}(x_i),g_{y_{i}}(x_i))\\
= &\underset{w_1,b_1,\ldots,w_K,b_K}{\min}~~\frac{1}{2}  \displaystyle \sum_{y=1}^{K} \|w_{y}\|^2+C_1 \sum_{i=1}^{m} \|\langle w_{y_i}, x_i \rangle+b_{y_i} \|^2\\
&+C_2 \sum_{i=1}^{m} \underset{j\not=y_i}{\max}\{\|\langle w_{y_i}, x_i \rangle+b_{y_i} \|^2-\|\langle w_{j}, x_i \rangle+b_{j}\|^2,0\}.\\
\end{split}
\end{align}

This model is easy to explain. The first regularization term avoids the model from over-fitting and improves the generalization ability. Minimizing the second term requires the $i$-th sample in the $y_i$-th class to be as close as possible to the $y_i$-th hyperplane. Minimizing the third term ensures the distance from the $i$-th sample in the $y_i$-th class to the $y_i$-th hyperplane smaller than the distance from the $i$-th sample in the $y_i$-th class to its nearest $j$-th hyperplane, where $j\not=y_i$. By observing the module, we see NSVM not only captures the structure of each class, but also separates different classes well.

Once solving \eqref{NSVMLM}, we obtain $\left(
\begin{smallmatrix}
        w_{y} \\
        b_{y} \\
\end{smallmatrix}
\right)$, $y=1,2,\ldots,K$, and predict the label of an unseen sample $x$ by
\begin{align}\label{assign-l}
\begin{split}
\textrm{label} (x) =\underset{y\in\{1,2,\ldots,K\}}{\arg\min} \mathcal{D}(g_y(x))=\frac{|g_y(x)|}{\|{w}_{y}\|}.
\end{split}
\end{align}

%
%
\subsection{Max-min distance-based NSVM solver}

Before solving our optimization problem \eqref{NSVMLM}, we first transfer it into an equivalent formulation.
%
%
Denote $B_{ij}=\|\langle w_{y_i}, x_i \rangle+b_{y_i} \|^2-\|\langle w_{j}, x_i \rangle+b_{j}\|^2$ and $B_i=(B_{i1},\ldots,B_{iK})^{\top}\in\mathbb{R}^K$. Obviously, $\underset{j=1,2,\ldots,K}{\max~}\{B_{ij}\}\geq0$. Suppose it achieves the maximum value at $j_i$, $j_i\in\{1,2,\ldots,K\}$.
Then define $F_i=(0,\ldots,0,1,0,\ldots,0)^{\top}\in\mathbb{R}^{K}$, where the $j_i$-th component $F_{ij_i}$ of $F_i$ is 1, and the rest components of $F_i$ are 0.
Then \eqref{NSVMLM} becomes
\begin{align}\label{NSSVMHinge4-1}
\begin{split}
\underset{w_1,b_1,\ldots,w_K,b_K}{\min}~~&\frac{1}{2}  \displaystyle \sum_{y=1}^{K} \|w_{y}\|^2+C_1 \sum_{i=1}^{m} \|\langle w_{y_i}, x_i \rangle+b_{y_i} \|^2\\
&+C_2\sum_{i=1}^{m}F_{i}^{\top}B_i.\\
\end{split}
\end{align}
Define $\widetilde{B}_i=(\|\langle w_{1}, x_i \rangle+b_{1}\|^2,\ldots,\|\langle w_{K}, x_i \rangle+b_{K}\|^2)^{\top}\in\mathbb{R}^K$.
Then $F_{i}^{\top}B_i=-F_{i}^{\top}\widetilde{B}_i+\|\langle w_{y_i}, x_i \rangle+b_{y_i} \|^2$.
Then \eqref{NSSVMHinge4-1} is transformed to
\begin{align}\label{NSSVMHinge4-2}
\begin{split}
\underset{w_1,b_1,\ldots,w_K,b_K}{\min}~~&\frac{1}{2}  \displaystyle \sum_{y=1}^{K} \|w_{y}\|^2+ (C_1+C_2)\sum_{i=1}^{m}\|\langle w_{y_i}, x_i \rangle+b_{y_i} \|^2\\
&-C_2\sum_{i=1}^{m} F_i^{\top}\widetilde{B}_i.
\end{split}
\end{align}
Denote the objective function of \eqref{NSSVMHinge4-2} as $h(w_1,b_1,\ldots,w_K,b_K)$.
Let $z_{i}=\begin{pmatrix}x_{i}\\1\end{pmatrix}$,
$\overline{w}_y=\begin{pmatrix}w_y\\b_y\end{pmatrix}$, $\overline{w}=\begin{pmatrix}\overline{w}_1;\cdots; \overline{w}_K\end{pmatrix}^{\top}\in\mathbb{R}^{(n+1)K}$, $D=C_1+C_2$.
Write
$G_y=\frac{1}{2}A+D\sum\limits_{y_i=y}z_iz_i^{\top}$, $A=\begin{pmatrix}\textbf{I}&\textbf{0}\\ \textbf{0}&0\end{pmatrix}$, $\textbf{I}$ is the identity matrix, $\textbf{0}$ is the matrix of zeros of an appropriate size. Let $G=\text{diag}(G_y)$. Here for matrices $Q_i$, $i=1,2,\ldots,N$, $\text{diag}(Q_i)$ is defined as the (block) diagonal matrix with its $(i,i)$-th block $Q_i$, and other $(i,j)$-th block is $\textbf{0}$ for $i\neq j$.


For $F_i$ that corresponds to the $i$-th sample $x_i$, suppose the position of its only nonzero element 1 is $j_i$. Define $M_{j_i}=\begin{pmatrix}\textbf{0},\cdots,\textbf{0},\textbf{I}_{j_i},\textbf{0},\cdots,\textbf{0}\end{pmatrix}
\in\mathbb{R}^{(n+1)\times(n+1)K}$, where $\textbf{0}$ is the zero matrix of size $(n+1)\times (n+1)$, $\textbf{I}_{j_i}$ is the identity matrix of size $(n+1)\times (n+1)$, and the $j_i$-th block of $M_{j_i}$ is $\textbf{I}_{j_i}$. Write $H_i=M_{j_i}^{\top}z_iz_i^{\top}M_{j_i}$ and $H=C_2\sum_{i=1}^mH_i$.
Then \eqref{NSSVMHinge4-2} can be rewritten as
\begin{align}\label{NSSVMHinge7}
\begin{split}
\underset{\overline{w}}{\min}~~h(\overline{w})=\left(\frac{L}{2}\|\overline{w}\|^2+\overline{w}^{\top} G\overline{w}\right)-\left(\frac{L}{2}\|\overline{w}\|^2+\overline{w}^{\top}H \overline{w}\right),
\end{split}
\end{align}
where $L>0$.

We now give some properties of optimization problem \eqref{NSSVMHinge7}.\\
\noindent\textbf{Property 1:}
(i) $G$ is positive definite;

(ii) $H$ is the sum of $m$ rank-1 matrices, and is positive semidefinite;

(iii) The objective of optimization problem \eqref{NSSVMHinge7} is lower bounded, and $\inf~h(\overline{w})\geq 0$.\\
\noindent\textbf{Proof:}
(i) and (ii) are easily obtained by definitions of $G$ and $H$.
For (iii), since each term of \eqref{NSSVMHinge4-1} is nonnegative and therefore the objective of \eqref{NSSVMHinge4-2} or $h(\overline{w})$ is lower bounded by 0.
\hfill$\square$
%

We now solve \eqref{NSSVMHinge7} and present its solving iteration procedure. Suppose $F_{i}$ is obtained from the last $t$-th iteration, and its largest component is $F_{ij_i^t}$.
Write $M_{j_i}^t=M_{j_i^t}$ and $H^t=C_2\sum_{i=1}^m(M_{j_i}^t)^{\top}z_iz_i^{\top}M_{j_i}^t$, $h_t(\overline{w})=\left(\frac{L}{2}\|\overline{w}\|^2+\overline{w}^{\top} G\overline{w}\right)-\left(\frac{L}{2}\|\overline{w}\|^2+\overline{w}^{\top}H^t \overline{w}\right)$.
For the $t$-iteration, note that $h_t(\overline{w})$ is non-convex. Therefore, we first majorize its concave part $-\left(\frac{L}{2}\|\overline{w}\|^2+\overline{w}^{\top}H^t \overline{w}\right)$ by its local linear approximation, and solve the resulting convex optimization subproblem. Inspired by the proximal difference-of-convex algorithm with extrapolation algorithm (pDCA$_e$) \cite{WenChenPong18}, to further accelerate the procedure, we also incorporate the extrapolation technique into the above approximation process.
In specific, we approximate $-\left(\frac{L}{2}\|\overline{w}\|^2+\overline{w}^{\top}H^t \overline{w}\right)$ by $-\left(Lu^t+2H^t \overline{w}^t\right)$, where $u^t=\overline{w}^t+\beta_t(\overline{w}^t-\overline{w}^{t-1})$ and $0\leq \beta_t<1$.
Note that different from pDCAe, $H^t$ is updated during each iteration. Therefore, we call our algorithm Modified pDCAe (MpDCAe).
Algorithm \ref{AlgorithmNSVM-lin} presents the solving procedure of \eqref{NSSVMHinge7}.

\begin{algorithm}\label{PTSVMforLinearmulti}
\caption{~MpDCAe (Linear NSVM solver).}
\begin{tabular}{ll}
\textbf{Input:} Randomize $\overline{w}^0\in\mathbb{R}^{(n+1)K}$, $\{\beta_t\}\subset(0,1)$ with\\$\text{sup~}\beta_t<1$, $t=1,\,2,\ldots$, parameters  $C_1,C_2>0$, $L>0$.\\
Set $\overline{w}^{1}=\overline{w}^0$. \\
\textbf{Process:}\\
\textbf{1. Training:}\\
Compute $G$;\\
{\bf for} $t=1,\,2,\ldots$\\
\quad Compute $H^t$;\\
\quad Take $\xi^t=2H^t \overline{w}^t$;\\
\quad $u^t=\overline{w}^t+\beta_t(\overline{w}^t-\overline{w}^{t-1})$;\\
\quad $\overline{w}^{t+1}=\arg\underset{\overline{w}}\min~f_t(\overline{w})=\overline{w}^{\top}G\overline{w}+\frac{L}{2}\|\overline{w}-\overline{u}^t\|^2-\overline{w}^{\top}\xi^t$.\\
{\bf end for}\\
Obtain $\overline{w}=\begin{pmatrix}\overline{w}_1;\cdots; \overline{w}_K\end{pmatrix}^{\top}$.\\
\textbf{2. Prediction:}\\
The label of $x$ is given by\\
\quad\quad \quad\quad\quad $\textrm{label} (x) =\underset{y\in\{1,2,\ldots,K\}}{\arg\min} \mathcal{D}(g_y(x))=\frac{|g_y(x)|}{\|\overline{w}_{y}\|}.$\\
\end{tabular}
\label{AlgorithmNSVM-lin}
\end{algorithm}

In Algorithm \ref{AlgorithmNSVM-lin}, the solution $\overline{w}^{t+1}$ of the last step is easily computed as
\begin{equation}
\begin{split}
\overline{w}^{t+1}=&\arg\underset{u}\min~-u^{\top}\xi^t+\frac{L}{2}\|u-u^t\|^2+u^{\top}Gu\\
=&\arg\underset{u}\min ~u^{\top}(\frac{L}{2}\textbf{I}+G)u - (Lu^t+\xi^t)u\\
=\,&(\frac{L}{2}\textbf{I}+G)^{-1}(Lu^t+\xi^t).
\end{split}
\end{equation}
In addition, for the choice of $\beta_t$, we may follow the way in fast iterative shrinkage-thresholding algorithm (FISTA) \cite{BeckTeboulle09} and modify it to our situation. In details, one starts with $\theta_{0}=\theta_1=1$, then recursively defines for $t\geq 1$ that
\begin{equation}\label{beta}
\begin{split}
\beta_t=\min\left\{\frac{\theta_{t-1}-1}{\theta_t},\sqrt{\frac{2\lambda_{\min}}{2\lambda_{\min}+L}}\right\}
\end{split}
\end{equation}
with $\theta_{t+1}=\frac{1+\sqrt{1+4\theta_t^2}}{2}$, where $\lambda_{\min}>0$ is the smallest eigenvalue of $G$,
and resets $\theta_{t-1}=\theta_t=1$ every $\widetilde{T}$ iterations for some fixed positive integer $\widetilde{T}$.

We now give some properties about Algorithm \ref{AlgorithmNSVM-lin}.\\
\textbf{Property 2:}
(i) $\frac{L}{2}\|\overline{w}\|^2+\overline{w}^{\top}H^t \overline{w}$ is strongly convex for each $t$;

(ii) $(\overline{w}^{t+1})^{\top}H^{t+1}\overline{w}^{t+1}\leq(\overline{w}^{t+1})^{\top}H^{t}\overline{w}^{t+1}$
holds for each $t$;

(iii) $f_t(\overline{w})$ is strongly convex for each $t$.

\noindent\textbf{Proof:}
(i) By the definition of $H^t$, $H^t$ is positive semidefinite. Combining $L>0$, we know $\frac{L}{2}\|\overline{w}\|^2+\overline{w}^{\top}H^t \overline{w}$ is strongly convex.

(ii) In the $(t+1)$-th iteration, for each $i\in\{1,2,\ldots,m\}$, we note that $(\overline{w}_{j_i}^{t+1})^{\top}z_iz_i^{\top}\overline{w}_{j_i}^{t+1}
=\underset{j=1,2,\ldots,K}{\min} (\overline{w}_j^{t+1})^{\top}z_iz_i^{\top}\overline{w}_j^{t+1}$.
Therefore, from the construction of $H^{t+1}$, it is easy to see that $H^{t+1}$ satisfies $(\overline{w}^{t+1})^{\top}H^{t+1}\overline{w}^{t+1}
=\underset{l=1,2,\ldots}{\min} (\overline{w}^{t+1})^{\top}H^{l}\overline{w}^{t+1}$.

(iii) is easily obtained by definition.
\hfill$\square$
\subsection{Theoretical analysis of Algorithm 1}
Before giving the convergency result of Algorithm \ref{AlgorithmNSVM-lin}, we first give the following two assumptions.

(i) First assume the sequence $\{{\overline{w}}^{t}\}$ generated by Algorithm \ref{AlgorithmNSVM-lin} is bounded. Then under this assumption, it is easily known that there exists some $\overline{L}>0$ such that when $L>\overline{L}$, $\left|\left|\frac{L}{2\sqrt{\lambda_{\min}}}(u^t-\overline{w}^t)-\sqrt{\lambda_{\min}}(\overline{w}^{t+1}-\overline{w}^{t})\right|\right|^2
>(\overline{w}^{t})^{\top}(H^{t-1}-H^t) \overline{w}^{t}$ holds for all $t$, since there exists only limited choices of $H^t$;

(ii) Secondly, we assume $h_{t-1}({\overline{w}}^{t})=(\overline{w}^{t})^{\top} G\overline{w}^{t}-(\overline{w}^{t})^{\top}H^{t-1} \overline{w}^{t}$ is lower bounded. Note that there exists some $\overline{C}_1>0$ such that when $C_1>\overline{C}_1$, $h_{t-1}({\overline{w}}^{t})$ is convex. Since there are limited number of matrices $H^t$, this assumption will hold for $C_1>\overline{C}_1$.

Now we give a theoretical result of Algorithm \ref{AlgorithmNSVM-lin}.
\begin{theorem}\label{thm}
The sequence $\{{\overline{w}}^{t}\}$ generated by Algorithm \ref{AlgorithmNSVM-lin} satisfies $\underset{t\rightarrow \infty}{\lim}\|{\overline{w}}^{t+1}-{\overline{w}}^{t}\|=0$ under the above assumptions.
\end{theorem}
\textbf{Proof:}
The proof is given in the Appendix.\hfill $\square$

Note that in assumption (ii), instead of assuming the optimization problem is convex, we only assume that $h_{t-1}({\overline{w}}^{t})$ is lower bounded. In real computation, it is satisfied for most situations, and we do not necessarily require $C_1>\overline{C}_1$.

\subsection{Nonlinear max-min distance-based NSVM model}

Suppose $\phi(\cdot):\mathbb{R}^n\longrightarrow\mathcal{F}$ is a nonlinear map from the input space $\mathbb{R}^n$ to a high-dimensional reproducing kernel Hilbert space $\mathcal{F}$. Let $\mathcal{K}$ be the kernel function defined by $\mathcal{K}(x_p,x_q)=\langle\phi(x_p),\phi(x_q)\rangle$.
Then our nonlinear max-min distance-based NSVM formulates as
\begin{align}\label{NSSVMHinge-n1}
\begin{split}
&\underset{v_1,d_1,\ldots,v_K,d_K}{\min}~~\frac{1}{2}  \displaystyle \sum_{y=1}^{K} \|v_{y}\|^2+C_1 \sum_{i=1}^{m} \|\langle v_{y_i}, \phi(x_i) \rangle+d_{y_i} \|^2\\
&~~~~+C_2 \sum_{i=1}^{m} \underset{j\not=y_i}{\max}(\|\langle v_{y_i}, \phi(x_i) \rangle+d_{y_i} \|^2
-\|\langle v_{j}, \phi(x_i) \rangle+d_{j}\|^2,0).\\
\end{split}
\end{align}

According to the representation theory \cite{JohnNello04}, we can write
$\begin{pmatrix}v_y\\b_y\end{pmatrix}=\begin{pmatrix}\phi(X)\\e^{\top}\end{pmatrix}\alpha_y$ for some $\alpha_y\in\mathbb{R}^{m\times1}$, $y=1,2,\ldots,K$, where $\phi(X) = \left(\phi(x_1),\phi(x_2),\ldots,\phi(x_m)\right)$, and $e\in\mathbb{R}^{m\times 1}$ is the vector of ones.
Let $z_{i}=\begin{pmatrix}\phi(x_{i})\\1\end{pmatrix}$,
$\overline{v}_y=\begin{pmatrix}v_y\\d_y\end{pmatrix}$, $S=\begin{pmatrix}\phi(X)\\e^{\top}\end{pmatrix}$. Then $\overline{v}_y=S\alpha_y$, and $S^{\top}z_i=\mathcal{K}(x_i,X)^{\top}+e\in\mathbb{R}^{m\times 1}$.
As in the linear case, denote $B_{ij}=\|\langle \overline{v}_{y_i}, z_i \rangle\|^2-\|\langle \overline{v}_{j}, z_i\rangle\|^2$ and $B_i=(B_{i1},\ldots,B_{iK})^{\top}\in\mathbb{R}^K$. Again, $\underset{j=1,2,\ldots,K}{\max~}\{B_{ij}\}\geq0$, and we suppose it achieves the maximum value at $j=j_i$, $j_i\in\{1,2,\ldots,K\}$.
Define
\begin{align*}
\begin{split}
\widetilde{B}_i&=(\|\langle \overline{v}_{1},z_i \rangle\|^2,\ldots,\|\langle \overline{v}_{K}, z_i \rangle\|^2)^{\top}\\
=&(({\alpha}_{1})^{\top}S^{\top}z_iz_i^{\top}S{\alpha}_{1}
,\ldots,({\alpha}_K)^{\top}S^{\top}z_iz_i^{\top}S{\alpha}_K)^{\top}\in\mathbb{R}^K
\end{split}
\end{align*}
and
$F_i=(0,\ldots,0,1,0,\ldots,0)^{\top}\in\mathbb{R}^{K}$, where the $j_i$-th component $F_{ij_i}$ of $F_i$ is 1, and the rest components of $F_i$ are 0.
Then $F_{i}^{\top}B_i=-F_{i}^{\top}\widetilde{B}_i+\|\langle \overline{v}_{y_i}, z_i \rangle\|^2$.
Consequently, problem \eqref{NSSVMHinge-n1} is equivalent to
\begin{align}\label{NSSVMHinge-n2}
\begin{split}
&\underset{\alpha_1,\cdots,\alpha_K}{\min}~~\frac{1}{2}  \displaystyle \sum_{y=1}^{K}
{\alpha}_y^{\top}\mathcal{K}(X,X){\alpha}_y-C_2 \sum_{i=1}^{m} F_{i}^{\top}\widetilde{B}_i\\ &+(C_1+C_2)\sum_{i=1}^{m}{\alpha}_{y_i}^{\top}(\mathcal{K}(x_i,X)^{\top}+e)(\mathcal{K}(x_i,X)^{\top}+e)^{\top}{\alpha}_{y_i}.
\end{split}
\end{align}
Define $G_y=\frac{1}{2}\mathcal{K}(X,X)+(C_1+C_2)\sum_{y_i=y}(\mathcal{K}(x_i,X)^{\top}+e)(\mathcal{K}(x_i,X)^{\top}+e)^{\top}$,
$G=\text{diag}(G_y)$, and $\alpha=\begin{pmatrix}\alpha_1;\cdots; \alpha_K\end{pmatrix}^{\top}$.
Further define $\overline{S}=\text{diag}(S)$, and $M_{j_i}=\begin{pmatrix}\textbf{0},\cdots,\textbf{0},\textbf{I},\textbf{0},\cdots,\textbf{0}\end{pmatrix}$ with the $j_i$-th block of $M_{j_i}$ being $\textbf{I}$, and $\textbf{0}$ elsewhere.
Write $H_i=\overline{S}^{\top}M_{j_i}^{\top}z_iz_i^{\top}M_{j_i}\overline{S}
=\begin{pmatrix}\textbf{0}&\cdots&\textbf{0}\\
\vdots&S^{\top}z_i^{\top}z_iS&\vdots\\
\textbf{0}&\cdots&\textbf{0}\end{pmatrix}\in\mathbb{R}^{mK\times mK}$ as the block matrix whose $(j_i,j_i)$-th subblock is the $m\times m$ matrix $S^{\top}z_i^{\top}z_iS=(\mathcal{K}(x_i,X)^{\top}+e)(\mathcal{K}(x_i,X)^{\top}+e)^{\top}$, and $\textbf{0}$ elsewhere.
Let $H=C_2\sum_{i=1}^mH_i$.

Then \eqref{NSSVMHinge-n2} can be rewritten as
\begin{align}\label{NSSVMHinge-n3}
\begin{split}
\underset{\alpha}{\min}~\left(\frac{L}{2}\|\alpha\|^2+\alpha^{\top} G\alpha\right)-\left(\frac{L}{2}\|\alpha\|^2+\alpha^{\top}H \alpha\right),
\end{split}
\end{align}
where $L>0$.

Clearly, our nonlinear optimization problem \eqref{NSSVMHinge-n3} of NSVM has the same formulation as its linear version \eqref{NSSVMHinge7}, and can be solved analogously as \eqref{NSSVMHinge7}.
Also, we see that the above nonlinear NSVM only involves the kernel operation, and therefore the predefined kernel can be applied directly. After obtaining optimal $\alpha=\begin{pmatrix}\alpha_1;\cdots; \alpha_K\end{pmatrix}^{\top}$, an unseen sample $x$ is assigned according to the following rule
\begin{align}\label{assign-n}
\begin{split}\text{Class}~x = \underset{y=1,\ldots,K}{\arg\min}~~ \frac{|\alpha_y^{\top}(\mathcal{K}(x,X)^{\top}+e)|}{\alpha_y^{\top}[\mathcal{K}(X,X)+ee^{\top}]\alpha_y}.
\end{split}
\end{align}

\section{Experimental comparisons}\label{experiments}

In this section, we compare our NSVM with six state-of-the-art NSVMs, including GEPSVM \cite{GEPSVM}, TBSVM \cite{TBSVM}, LSTSVM \cite{LSTSVM}, NPSVM \cite{NPSVM}, PCC \cite{PCC}, and NHSVM \cite{NHSVM}.
Experiments are conducted on an artificial data set and some benchmark UCI data sets \cite{UCI98}. For multiclass data, the binary classification methods are carried out by using the ``1-against-all" technique, and all the methods are implemented in MATLAB 2017a environment on a PC with Intel i7 processor (3.60 GHz) with 32 GB RAM. For nonlinear SVMs, Radial Basis Function (RBF) kernel is used. The parameters for all the methods, including the regularization trade-off parameters $C_1$ and $C_2$ and kernel parameter $\sigma$, are selected from the set $\{2^{-10},\ldots,2^{10}\}$. For parameters selection, the standard 10-fold cross-validation (10-CV) technique is employed. The classification accuracy is used as the metric for comparison.

\subsection{An artificial data set}

\begin{figure*}
\begin{center}{
\resizebox*{4.9cm}{!}{
\subfigure[Original data]{\includegraphics[width=0.22\textheight]{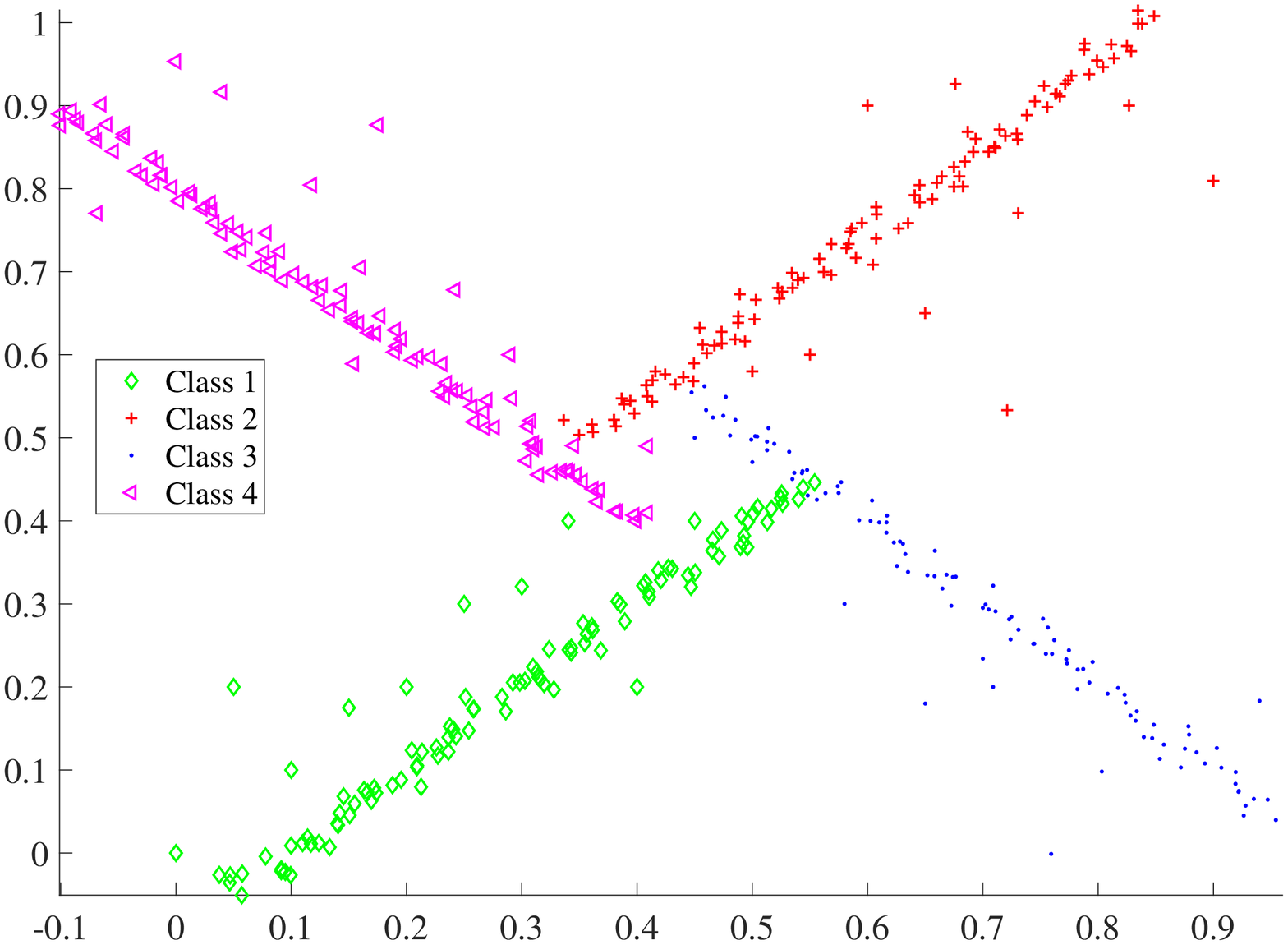}}}
\resizebox*{4.9cm}{!}{
\subfigure[GEPSVM]{\includegraphics[width=0.22\textheight]{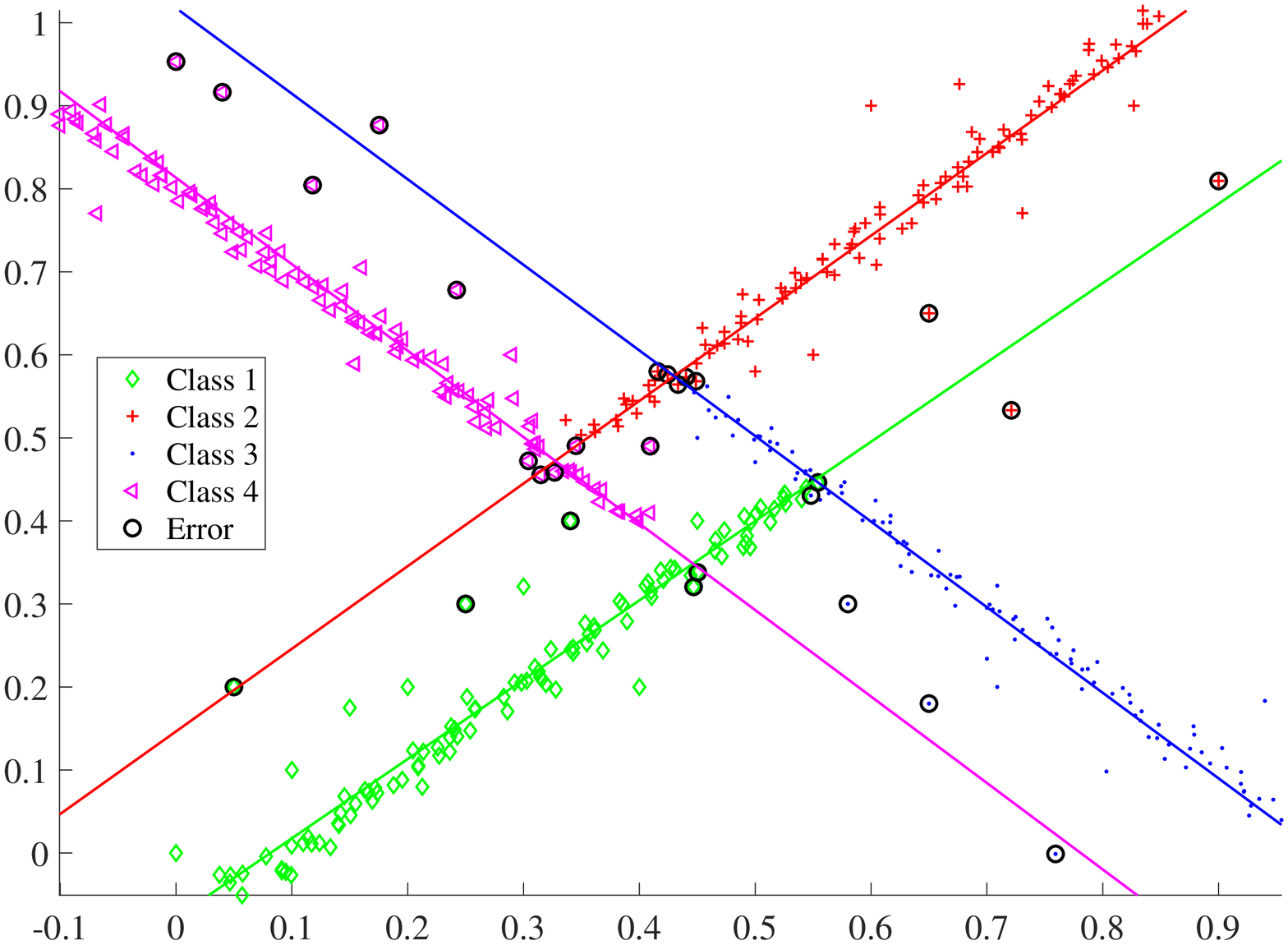}}}
\resizebox*{4.9cm}{!}{
\subfigure[TBSVM]{\includegraphics[width=0.22\textheight]{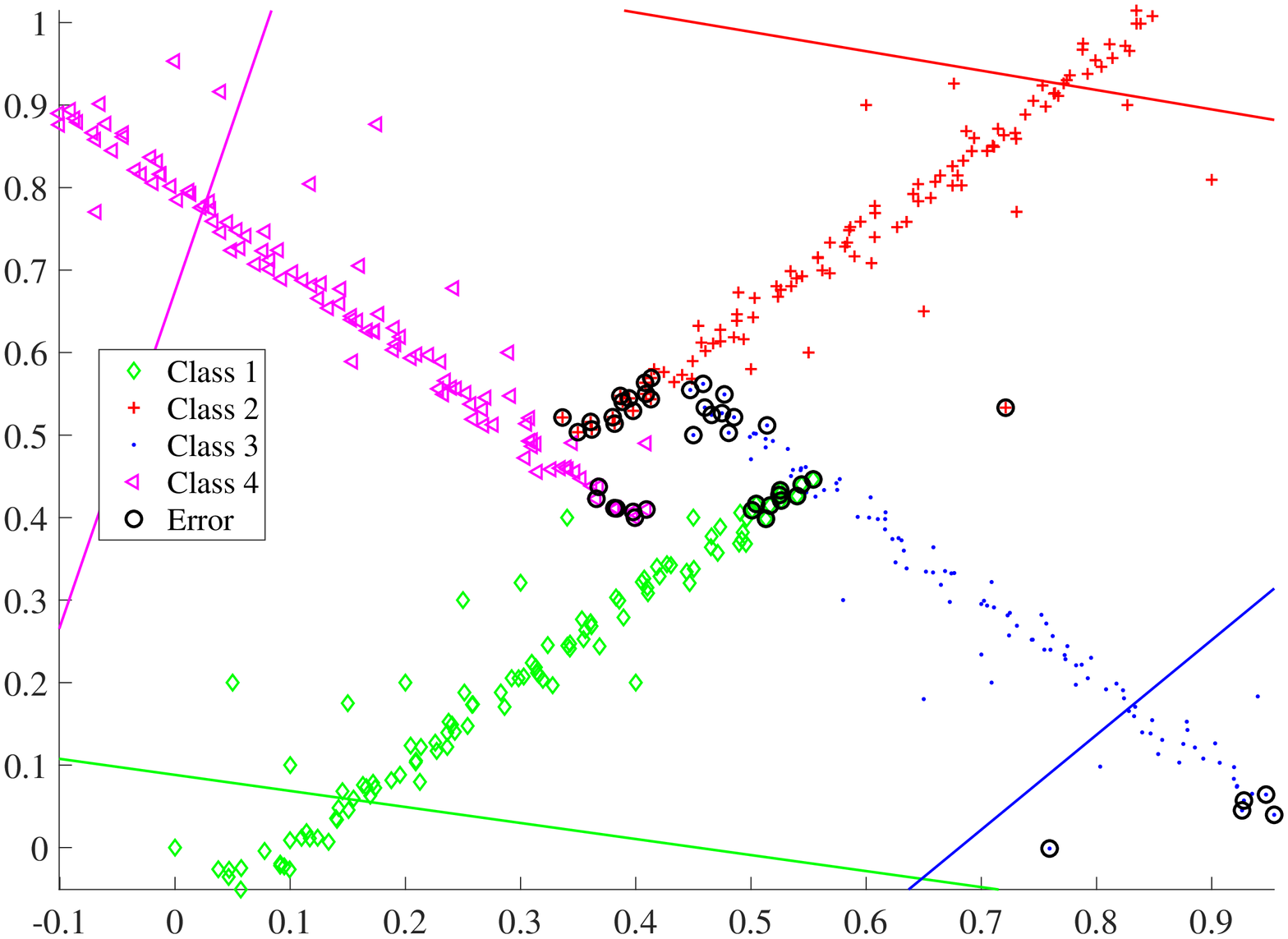}}}
\resizebox*{4.9cm}{!}{
\subfigure[LSTSVM]{\includegraphics[width=0.22\textheight]{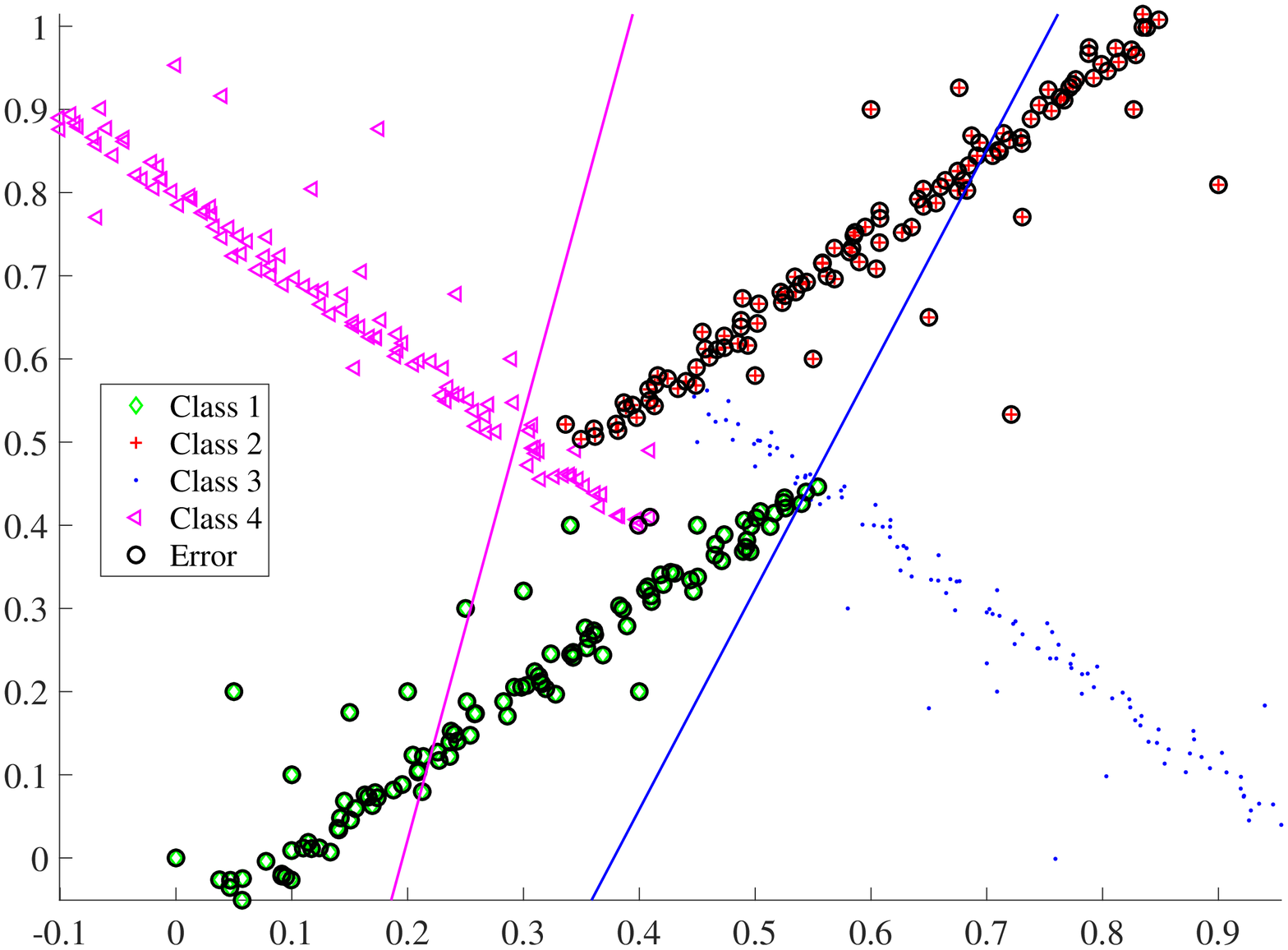}}}
\resizebox*{4.9cm}{!}{
\subfigure[ NPSVM ]{\includegraphics[width=0.22\textheight]{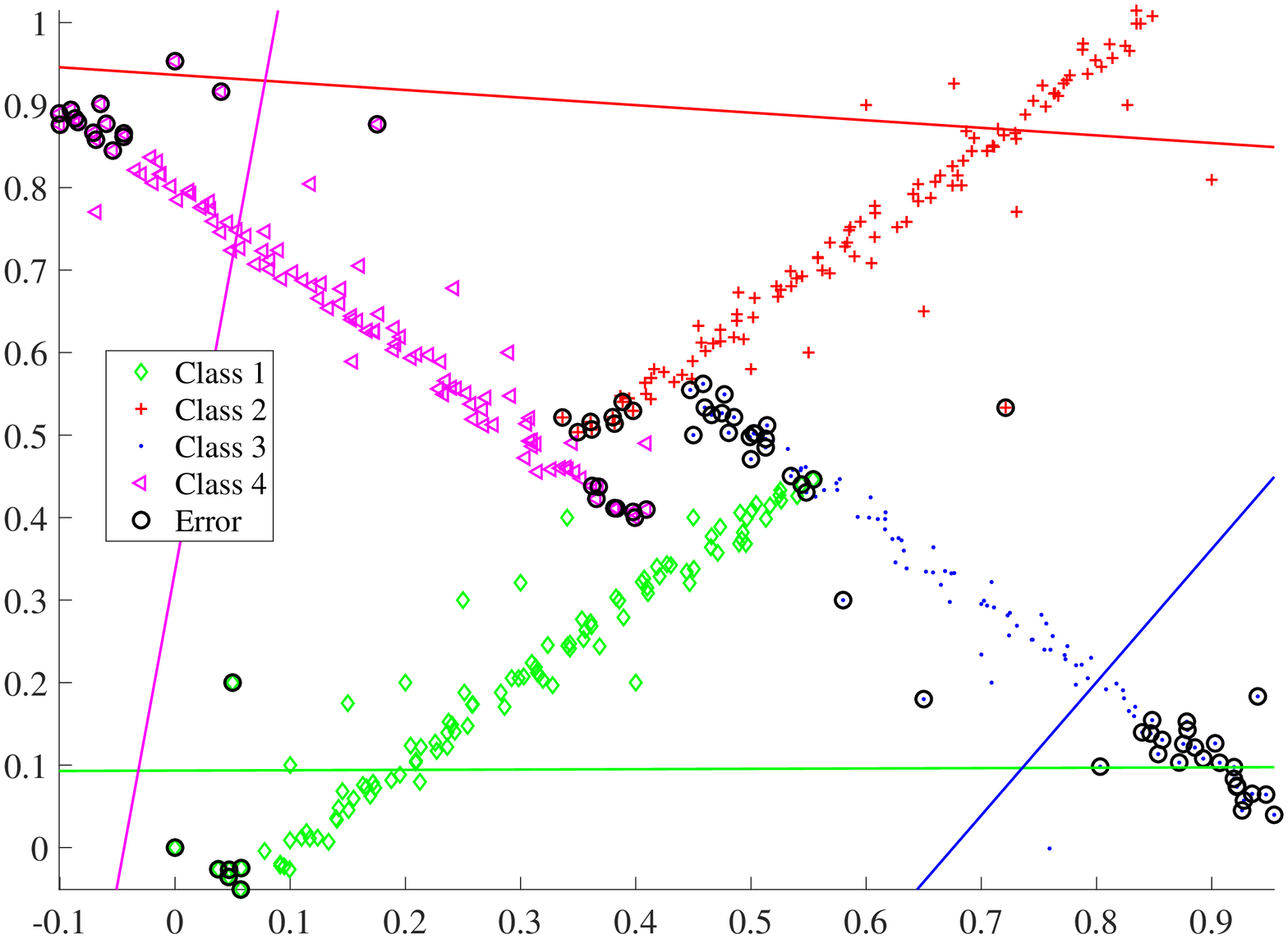}}}
\resizebox*{4.9cm}{!}{
\subfigure[PCC ]{\includegraphics[width=0.22\textheight]{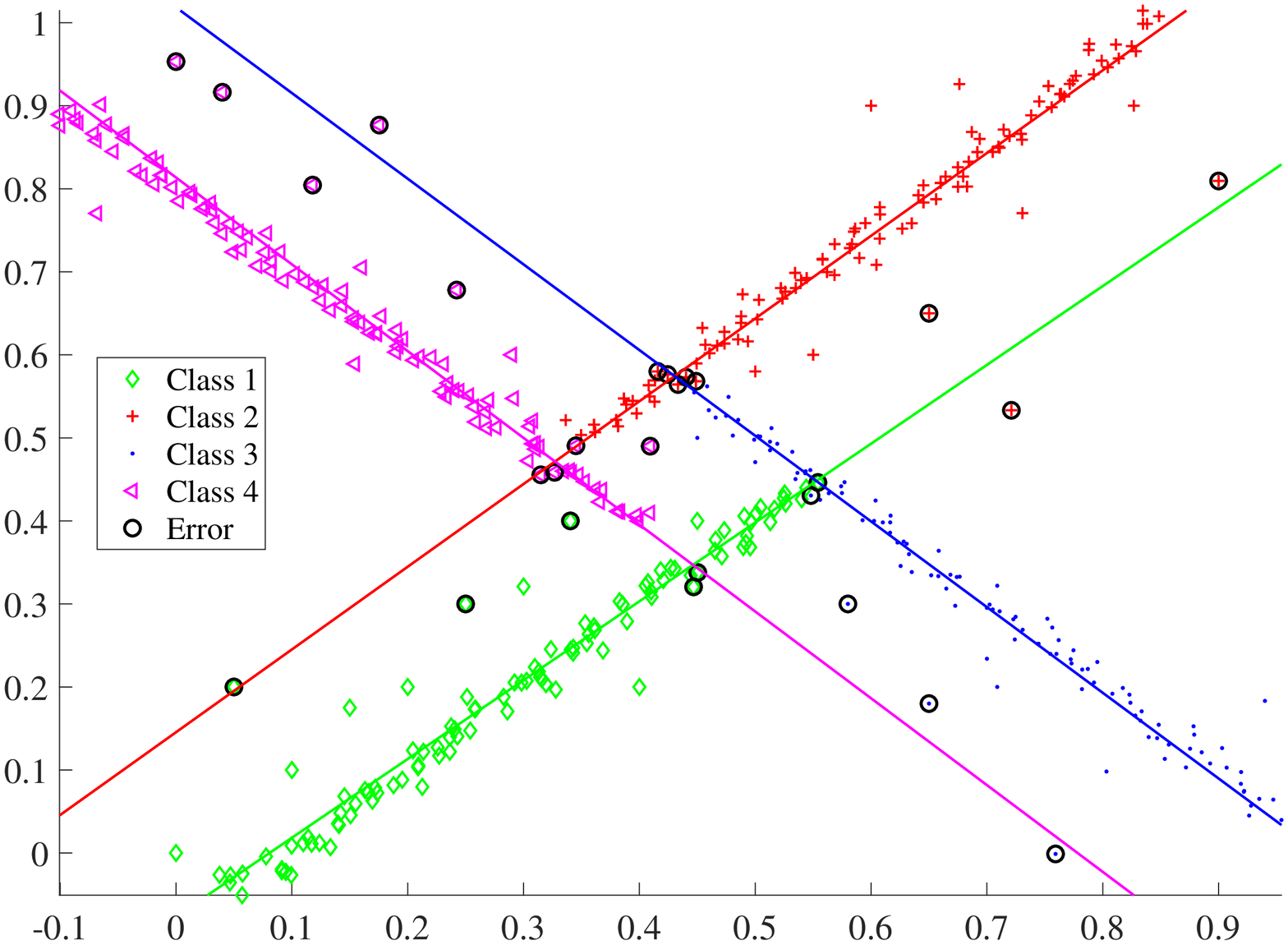}}}
\resizebox*{4.9cm}{!}{
\subfigure[NHSVM]{\includegraphics[width=0.22\textheight]{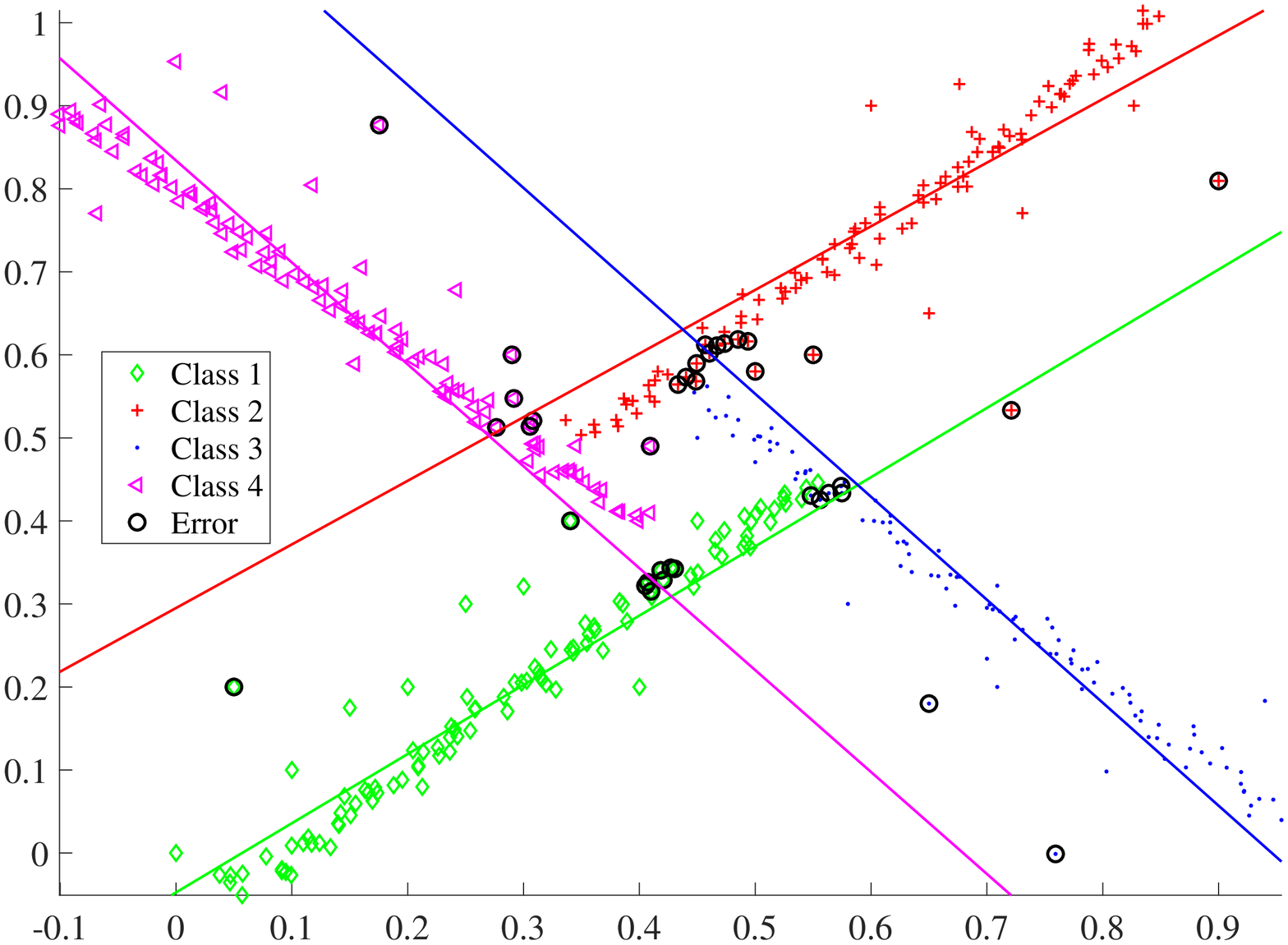}}}
\resizebox*{4.9cm}{!}{
\subfigure[NSVM]{\includegraphics[width=0.22\textheight]{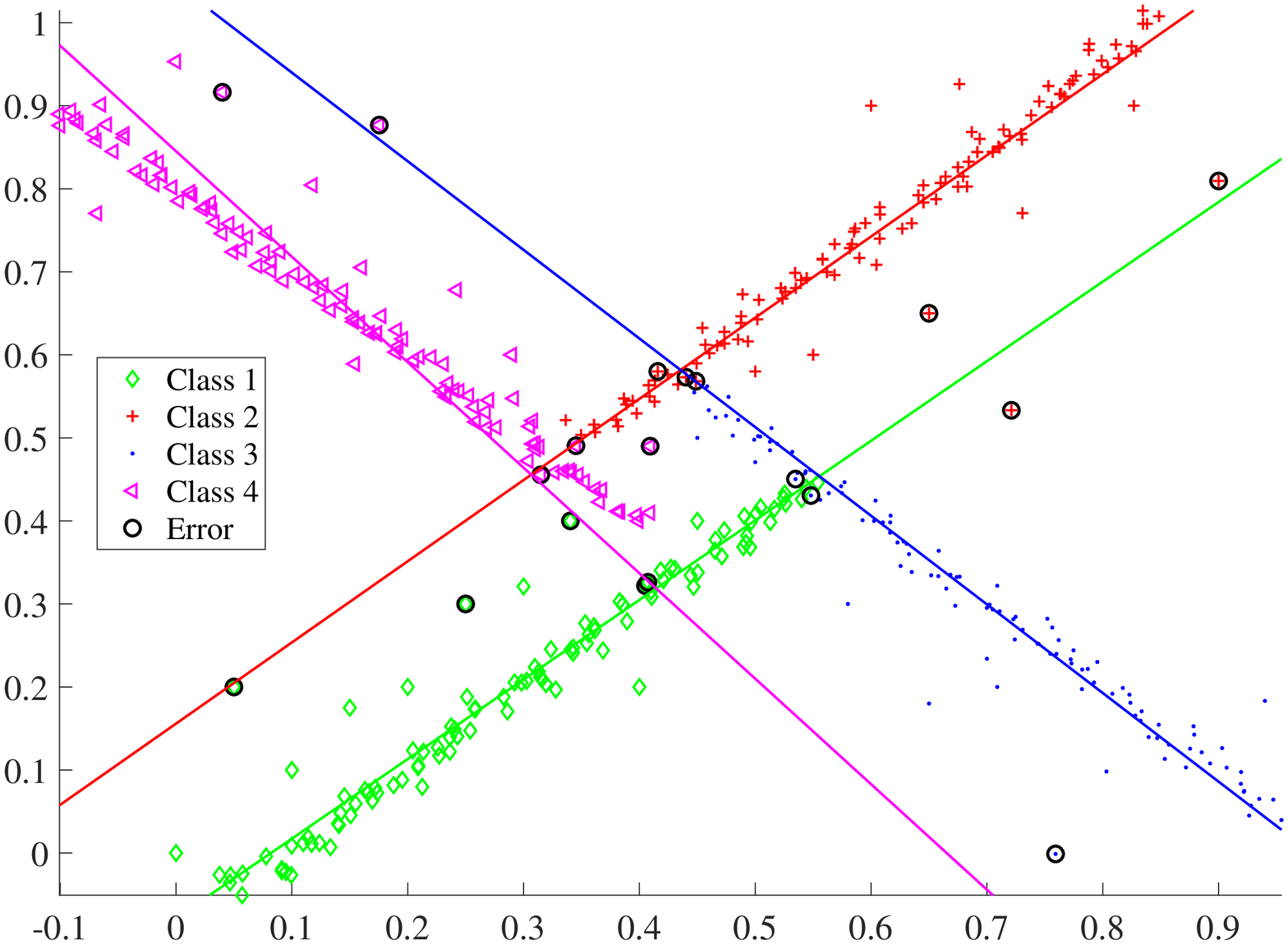}}}
\caption{Results of linear classifiers on ``Cross Planes" data set. (b)-(e) belong to the \textit{Single Models}, and (f)-(h) belong to the \textit{Union Models}.}
\label{figarti}}
\end{center}
\end{figure*}

We first apply all the methods to a two-dimensional artificial ``Cross Planes" data set, which was usually used in NSVMs to indicate their classification abilities \cite{GEPSVM}. The data set contains four classes, where each class is generated from a line with random perturbation and some outliers, as shown in Fig.\ref{figarti} (a). We apply each method on this data, and the obtained separating lines for all the methods are depicted in Fig.\ref{figarti} (b)-(h). The samples that are classified wrongly are circled in black.
From the figure, we see that on this data set, NSVMs belonging to \textit{The framework of Union Models} perform better than the ones belonging to \textit{The framework of Single Models} except GEPSVM. For TBSVM, LSTSVM and NPSVM, they have poor classification ability on samples in the central diamond area. In contrast, NSVMs belonging to \textit{The framework of Union Models} can predict the labels of most of these samples correctly.
To see the results clearer, the corresponding classification accuracies are listed in Table \ref{Tablearti2}. The results support the above argument, and also show the superiority of our proposed NSVM.
\begin{table}
\begin{center}
\centering \caption{Classification results on an artificial data set.}
\begin{tabular}{c|cccccccccc}
\toprule
Classifier ~&~ GEPSVM ~& TBSVM ~& LSTSVM ~& NPSVM ~& PCC ~& NHSVM ~& NSVM \\
\hline
AC  ~&~ 93.64  ~&  89.32    ~&  49.55   ~&  80.68   ~&  93.86  ~& 91.59  ~&  \textbf{ 95.68}  \\
\bottomrule
\end{tabular} \label{Tablearti2}
\end{center}
\end{table}
\subsection{Benchmark UCI data sets}
\subsubsection{Performance analysis}
In this subsection, we evaluate our NSVM with other methods on some UCI data sets, the information of which is listed in Table \ref{UCIinfo}. The data sets are from different areas, with different scales of samples and different density ratio. To compare the overall performance of each method, we compute the 10-CV accuracy for each data set, as well as the $p$-value in 5\% significance level. Mean accuracy (AC) and standard accuracy (Std) over 10-CV accuracies are considered. The $p$-value is calculated by performing a paired $t$-test by comparing our best to the other methods under the assumption of the null hypothesis that there is no difference between the test set accuracy distributions.

We first compare the behavior of our linear NSVM with other linear methods. The classification results as well $p$-values are presented in Table \ref{TableUCIlinear}. From the table, we have the following observations: (i) Our NSVM owns the highest accuracy on more data sets than other methods. (ii) By observing $p$-values, it can be seen that the behavior of our NSVM is statistically different than those of the other methods on most data sets. (iii) By comparing $t$-test values, we also compute a comprehensive metric Win-Tie-Loss (W-T-L) on accuracy to characterize the relative performance, which denotes the number of metrics that our NSVM is significantly superior/equal/inferior to the compared classifiers. The corresponding W-T-L values that are listed at the bottom of the table confirm  the conclusion that NSVM performs well on most of the data sets, which reveals the feasibility of our approach.
\begin{table}[htbp]
\begin{center}
\caption{Details of benchmark UCI data sets}
\resizebox{4.0in}{!}
{
\begin{tabular}{l|ccccccccccc}
\toprule
Data set ~& No. Sample ~& No.Feature ~& No.Class ~& Density Ratio \\
\hline
Iris  & 150 & 4 & 3 & 100\%\\
Wine  & 178 & 13 & 3 & 100\%\\
DNA  & 2000 & 180 & 3 & 25.34\%\\
Zoo  & 101 & 16 & 7 & 45.67\%\\
Vowel  & 528 & 10 & 11 & 99.94\%\\
Pathbased  & 300 & 2 & 3 & 100\%\\
Haberman  & 306 & 3 & 2 & 85.19\%\\
Heartc  & 303 & 14 & 2 & 72.89\%\\
Spanbas  & 4601 & 57 & 2 & 22.59\%\\
Dermatology  & 366 & 34 & 6 & 40.37\%\\
Seeds  & 210 & 7 & 3 & 100\%\\
WPBC  & 198 & 34 & 2 & 98.71\%\\
Hepatitis  & 155 & 19 & 2 & 99.97\%\\
\bottomrule
\end{tabular}
}
\label{UCIinfo}
\end{center}
\end{table}

We then investigate the classification ability of our nonlinear NSVM and other nonlinear classifiers, and the corresponding results are described in Table \ref{TableUCInonlinear}. From the table, we have the following observations: (i) On most of the data sets, the nonlinear classifiers have better classification ability. For example, on the ``Pathbased" data set, almost all the methods enhanced their accuracies for 30\%. (ii) Our NSVM possesses the highest classification accuracy on 6 out of 9 data sets. (iii) $p$-values show that the behavior of our nonlinear NSVM is statistically different than those of the other nonlinear methods on most data sets. (iv) As the linear case, the W-T-L values again show that our NSVM outperforms the other methods.
\begin{table*}[htbp]
\begin{center}
\caption{Linear classification results on benchmark UCI data sets}
\resizebox{5.5in}{!}
{
\begin{tabular}{l|ccccccccccc}
\toprule
& GEPSVM ~& TBSVM & LSTSVM & NPSVM & PCC & NHSVM & NSVM\\
{\multirow{1}{*}{Data set }} & AC$\pm$Std & AC$\pm$Std & AC$\pm$Std & AC$\pm$Std & AC$\pm$Std & AC$\pm$Std & AC$\pm$Std\\
& $p$& $p$& $p$& $p$& $p$ & $p$& $p$\\
\hline
Iris   &~96.94$\pm$0.63 ~& 93.77$\pm$0.83 ~& 66.90$\pm$0.49 ~&95.83$\pm$0.47 ~& 96.53$\pm$0.47 ~& 94.68$\pm$0.58 ~& \textbf{98.34$\pm$0.35} \\
  ~& 0.0004 ~& 0.0000 ~& 0.0000 ~& 0.0000 ~& 0.0000 ~& 0.0000  ~&--  \\
Wine   &~90.76$\pm$1.04 ~& 98.10$\pm$0.44 ~& 96.89$\pm$0.52 ~& 98.04$\pm$0.92 ~& 92.43$\pm$1.19 ~& \textbf{98.27$\pm$0.59} ~& 95.24$\pm$0.57 \\
  ~& 0.0000 ~& 0.0000 ~& 0.0002 ~& 0.0001 ~& 0.0001 ~& 0.0000  ~&--   \\
DNA   &~92.02$\pm$0.14 ~& \textbf{94.67$\pm$0.27} ~& 94.22$\pm$0.15 ~& 92.19$\pm$0.29 ~& 93.03$\pm$0.19 ~& 94.13$\pm$0.14 ~& 92.09$\pm$0.20 \\
  ~& 0.3104 ~& 0.0000 ~& 0.0000 ~& 0.4667 ~& 0.0000 ~& 0.0000  ~&--  \\
Zoo  &~72.14$\pm$8.15 ~& 93.65$\pm$1.27 ~& 87.44$\pm$1.02 ~& 94.47$\pm$1.28 ~& 79.68$\pm$1.89 ~& 91.99$\pm$1.11 ~& \textbf{95.60$\pm$0.36} \\
  ~& 0.0000 ~& 0.0021 ~& 0.0000 ~& 0.0081 ~& 0.0000 ~& 0.0000 ~&--   \\
Vowel   &~52.28$\pm$1.57 ~& 51.77$\pm$1.19 ~& 31.33$\pm$0.56 ~& 33.34$\pm$0.69 ~& 50.02$\pm$1.53 ~& 49.81$\pm$1.22 ~& \textbf{52.50$\pm$0.65} \\
  ~& 0.7011 ~& 0.0422 ~& 0.0000 ~& 0.0000 ~& 0.0008 ~& 0.0004  ~&--  \\
Pathbased  &~63.47$\pm$0.61 ~& 63.82$\pm$0.24 ~& 50.42$\pm$0.78 ~& 69.50$\pm$3.42 ~& 59.71$\pm$1.47 ~& \textbf{74.38$\pm$1.27} ~& 69.72$\pm$0.93 \\
  ~& 0.0000 ~& 0.0000 ~& 0.0000 ~& 0.8422 ~& 0.0000 ~& 0.0000  ~&--  \\
Haberman  &~74.68$\pm$1.04 ~& \textbf{75.41$\pm$0.87} ~& 69.70$\pm$1.09 ~& 73.82$\pm$0.37 ~& 73.82$\pm$1.41 ~& 73.03$\pm$0.36 ~& 74.65$\pm$0.71 \\
  ~& 0.9392 ~& 0.0652 ~& 0.0000 ~& 0.0019 ~& 0.0939 ~& 0.0005 ~&--    \\
Heartc  &~89.64$\pm$0.42 ~& 98.60$\pm$0.36 ~& 50.84$\pm$1.03 ~& 99.51$\pm$0.19 ~& 97.25$\pm$0.53 ~& 99.02$\pm$0.37 ~& \textbf{99.53$\pm$0.26} \\
  ~& 0.0000 ~& 0.0001 ~& 0.0000 ~& 0.8275 ~& 0.0000 ~& 0.0022   ~&--   \\
Spanbas  &~84.28$\pm$1.71 ~& 92.51$\pm$0.14 ~& 79.71$\pm$0.62 ~& 92.70$\pm$0.10 ~& 88.68$\pm$0.14 ~& \textbf{92.88$\pm$0.08} ~& 86.29$\pm$0.13 \\
  ~& 0.0060 ~& 0.0000 ~& 0.0000 ~& 0.0000 ~& 0.0000 ~& 0.0000  ~&--  \\
\hline
W-T-L (AC) & 6-3-0 & 4-2-3  & 7-0-2 & 4-3-2 & 6-1-2 & 5-0-4 \\
\bottomrule
\end{tabular}
}
\label{TableUCIlinear}
\end{center}
\end{table*}
\begin{table*}[htbp]
\begin{center}
\caption{Nonlinear classification results on benchmark UCI data sets}
\resizebox{5.5in}{!}
{
\begin{tabular}{l|ccccccccccc}
\toprule
& GEPSVM ~& TBSVM & LSTSVM & NPSVM & PCC & NHSVM & NSVM\\
{\multirow{1}{*}{Data set }} & AC$\pm$Std & AC$\pm$Std & AC$\pm$Std & AC$\pm$Std & AC$\pm$Std & AC$\pm$Std & AC$\pm$Std\\
& $p$& $p$& $p$& $p$& $p$ & $p$& $p$\\
\hline
Iris   &~91.54$\pm$3.96 ~& 96.43$\pm$0.51 ~& 96.65$\pm$0.85 ~& 96.48$\pm$1.23 ~& 96.27$\pm$0.90 ~& 96.22$\pm$0.79 ~& \textbf{97.31$\pm$0.91} \\
  ~&0.0017~&0.0089 ~&0.1891~&0.1578~&0.0428~&0.0087 ~&-- \\
Dermatology  &~91.71$\pm$0.53 ~& \textbf{98.16$\pm$0.25} ~& 96.96$\pm$0.16 ~& 97.75$\pm$0.25 ~& 90.32$\pm$0.52 ~& 97.46$\pm$0.24 ~& 96.57$\pm$0.22\\
  ~&0.0000~&0.0000~&0.0011~&0.0000~&0.0000~&0.0000 ~&-- \\
Seeds  &~91.37$\pm$1.51 ~& 93.97$\pm$0.90 ~& 93.49$\pm$0.61 ~& \textbf{ 96.35$\pm$0.65} ~& 91.56$\pm$0.55 ~& 93.38$\pm$0.69 ~& 93.02$\pm$0.82\\
  ~&0.0012~&0.0088~&0.2450~&0.0000~&0.0003~&0.3399  ~&--  \\
Wine   &~94.85$\pm$1.02 ~& \textbf{98.57$\pm$0.72} ~& 98.08$\pm$0.27 ~& 97.93$\pm$0.79 ~& 94.59$\pm$0.60 ~& 97.10$\pm$0.73 ~& 97.40$\pm$0.35\\
 ~&0.0001~&0.0010~&0.0011~&0.0525~&0.0000~&0.3659  ~&--  \\
Pathbased  &~95.22$\pm$3.93 ~& 98.54$\pm$0.47 ~& 98.48$\pm$0.39 ~& 99.19$\pm$0.32 ~& 98.29$\pm$0.34 ~& 98.94$\pm$0.23 ~& \textbf{99.36$\pm$0.14}\\
  ~&0.0095~&0.0005~&0.0002~&0.2300~&0.0000~&0.0005 ~&--  \\

WPBC  &~76.93$\pm$1.10 ~& 79.77$\pm$1.49 ~& 70.61$\pm$3.21 ~& 79.06$\pm$1.53 ~& 77.58$\pm$0.95 ~& 80.19$\pm$1.16 ~& \textbf{81.49$\pm$1.14}\\
   ~&0.0000~&0.0046~&0.0000~&0.0009~&0.0000~&0.0517 ~&--  \\

Haberman  &~73.59$\pm$0.99 ~& 75.50$\pm$0.79 ~& 75.48$\pm$1.25 ~& 73.38$\pm$1.11 ~& 71.32$\pm$1.56 ~& 74.99$\pm$0.93 ~& \textbf{76.52$\pm$0.72}\\
  ~&0.0001~&0.0144~&0.0915~&0.0000~&0.0000~&0.0028 ~&--   \\
Heartc  &~81.78$\pm$4.04 ~& 99.29$\pm$0.25 ~& 93.07$\pm$0.69 ~& 99.75$\pm$0.23 ~& 93.10$\pm$0.78 ~& 99.87$\pm$0.16 ~& \textbf{99.91$\pm$0.14}\\
  ~&0.0000~&0.0001~&0.0000~&0.0586~&0.0000~&0.5044  ~&-- \\
Hepatitis  &~80.45$\pm$1.92 ~& 81.45$\pm$1.45 ~& 79.88$\pm$3.68 ~& 83.59$\pm$0.98 ~& 80.39$\pm$1.18 ~& 82.07$\pm$1.81 ~& \textbf{84.94$\pm$0.82}\\
  ~&0.0002~&0.0004~&0.0041~&0.0085~&0.0000~&0.0003   ~&-- \\
\hline
W-T-L (AC) & 9-0-0 & 6-0-3  & 4-3-2 & 3-4-2 & 8-1-0 & 4-4-1 \\
\bottomrule
\end{tabular}
}
\label{TableUCInonlinear}
\end{center}
\end{table*}

\subsubsection{Computation complexity analysis}
We give the computation complexity of each method. Remember that $m$ is the number of samples, $n$ is the feature dimension, and $K$ is the number of classes. We only consider the linear case. For GEPSVM, TBSVM, LSTSVM and NPSVM, they belong to the \textit{The framework of Single Models}, and hence they need to solve $K$ subproblems. For convenience, to compute their computation complexity, we here suppose each class contains $\frac{m}{K}$ samples.
GEPSVM solves eigenvalue problems, and its computation complexity is $O(Kn^3)$. TBSVM solves quadratic programm problems, and its computation complexity is $O(Kn^3)$. LSTSVM copes with linear systems of equations, and its computation complexity is $O(Kn^3)$. NPSVM can be solved efficiently by the SMO-type technique, and its time complexity is $O(K\cdot1.5T_0n)$, where $T_0$ is the iteration number in solving NPSVM. PCC and NHSVM belong to the \textit{The framework of Union Models}, and their computation complexity is $O(n^3)$.
For our NSVM, the main cost lies in the computation of $\overline{w}^{t+1}$ in Algorithm 1. Assume the iteration number of Algorithm 1 is $T$. Then the computation complexity of NSVM is $O(Tn^3)$.

\begin{figure*}
\begin{center}{
\resizebox*{4.9cm}{!}{
\subfigure[Iris]{\includegraphics[width=0.22\textheight]{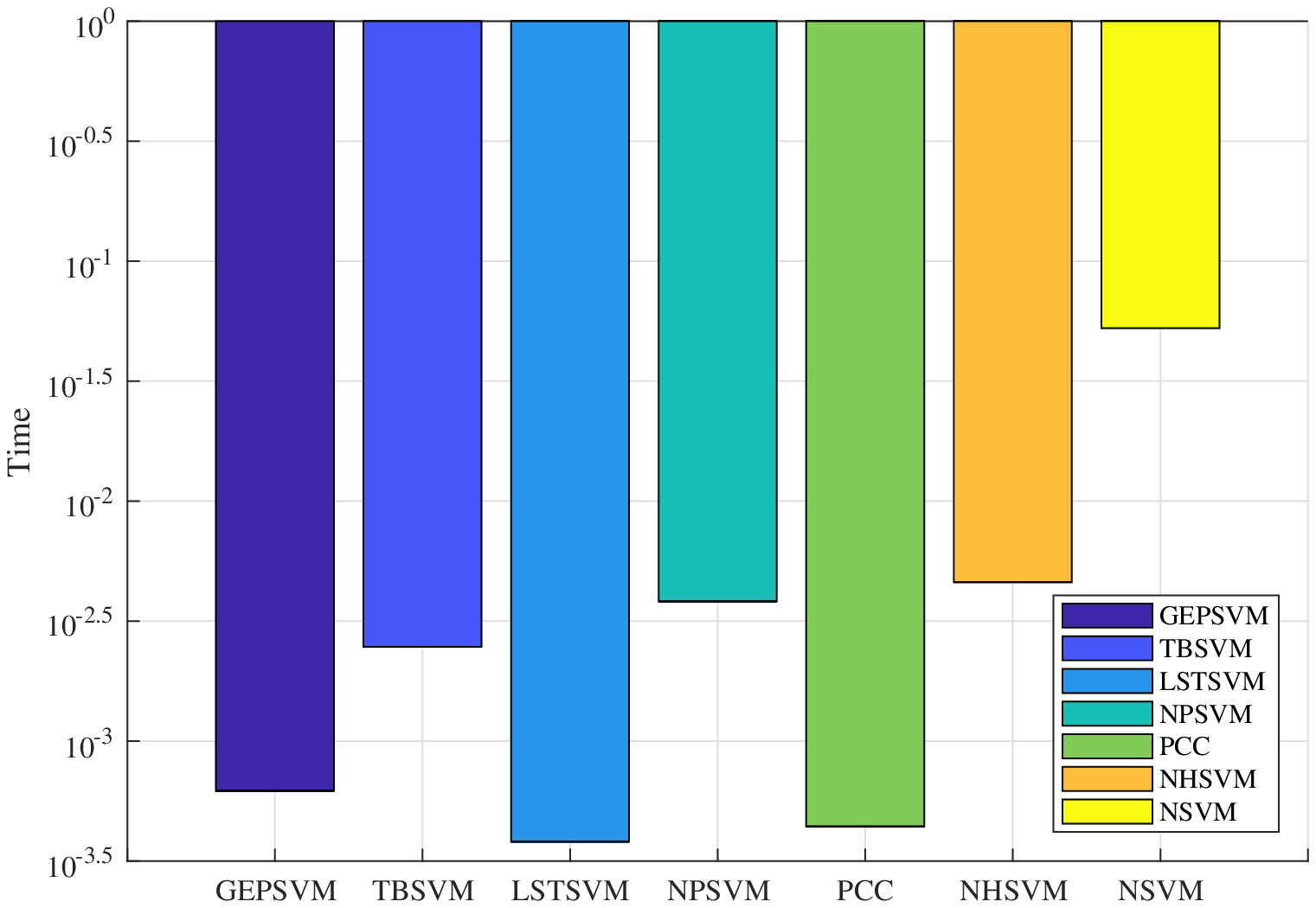}}}
\resizebox*{4.9cm}{!}{
\subfigure[Wine]{\includegraphics[width=0.22\textheight]{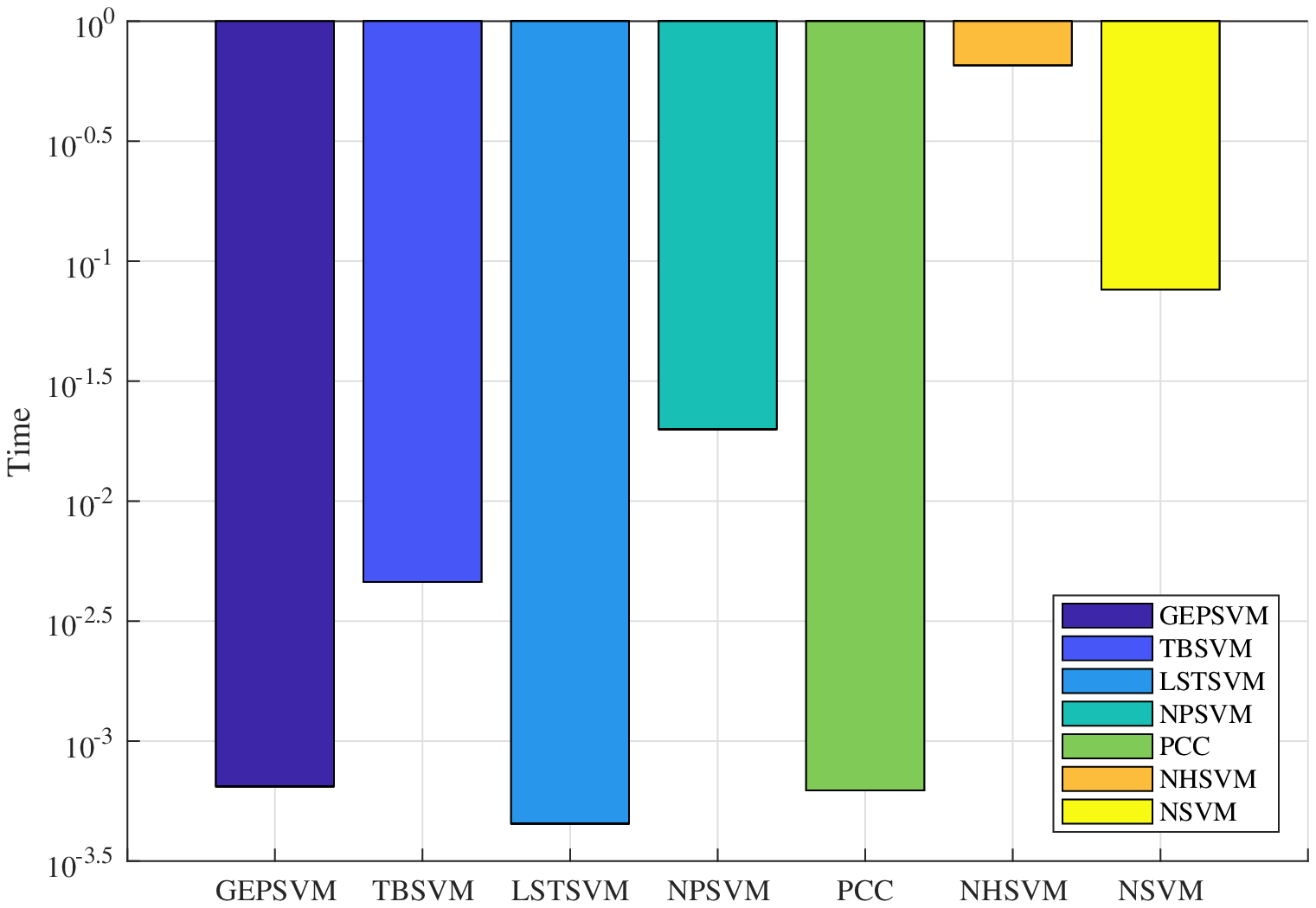}}}
\resizebox*{4.9cm}{!}{
\subfigure[DNA]{\includegraphics[width=0.22\textheight]{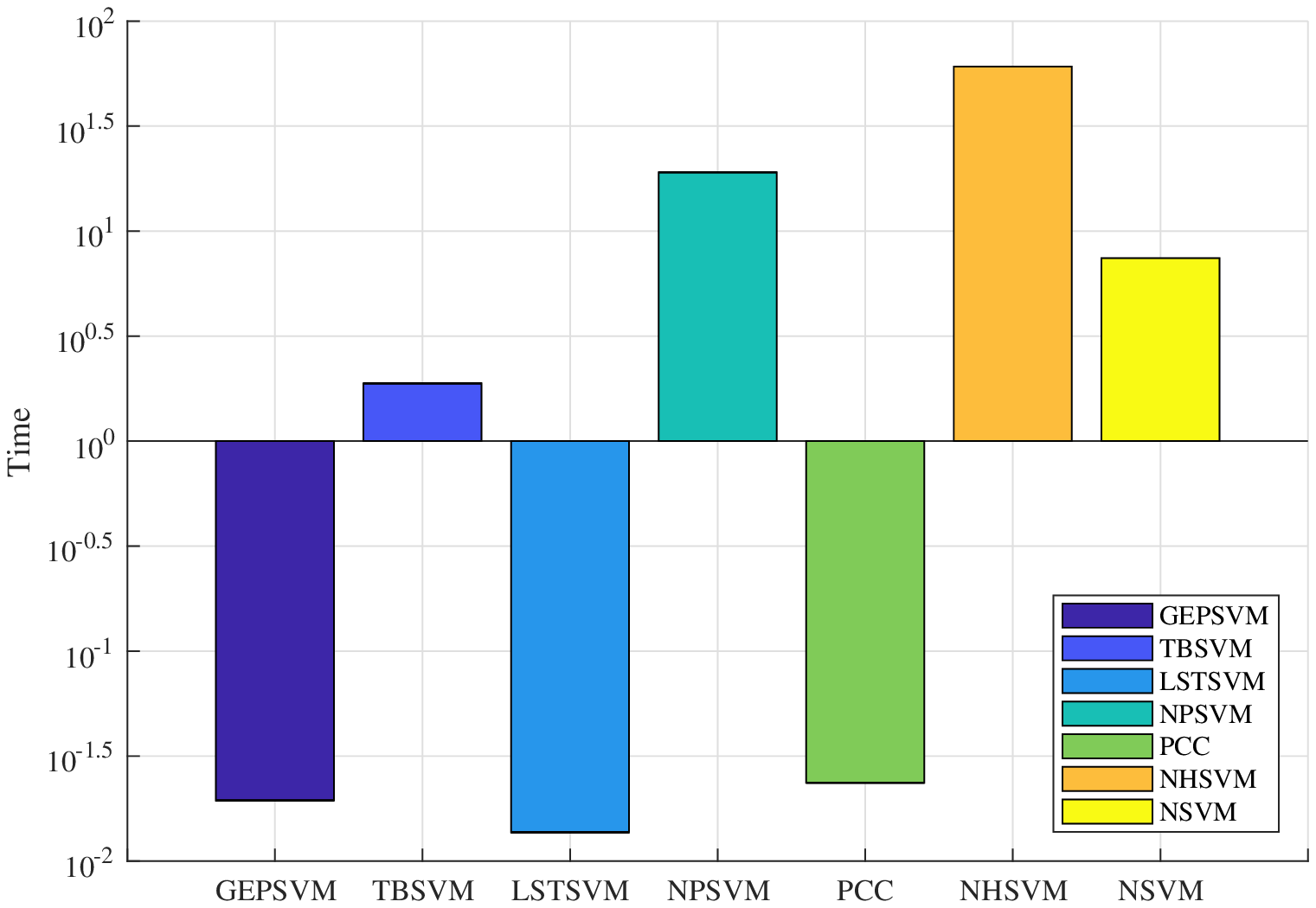}}}
\resizebox*{4.9cm}{!}{
\subfigure[Zoo]{\includegraphics[width=0.22\textheight]{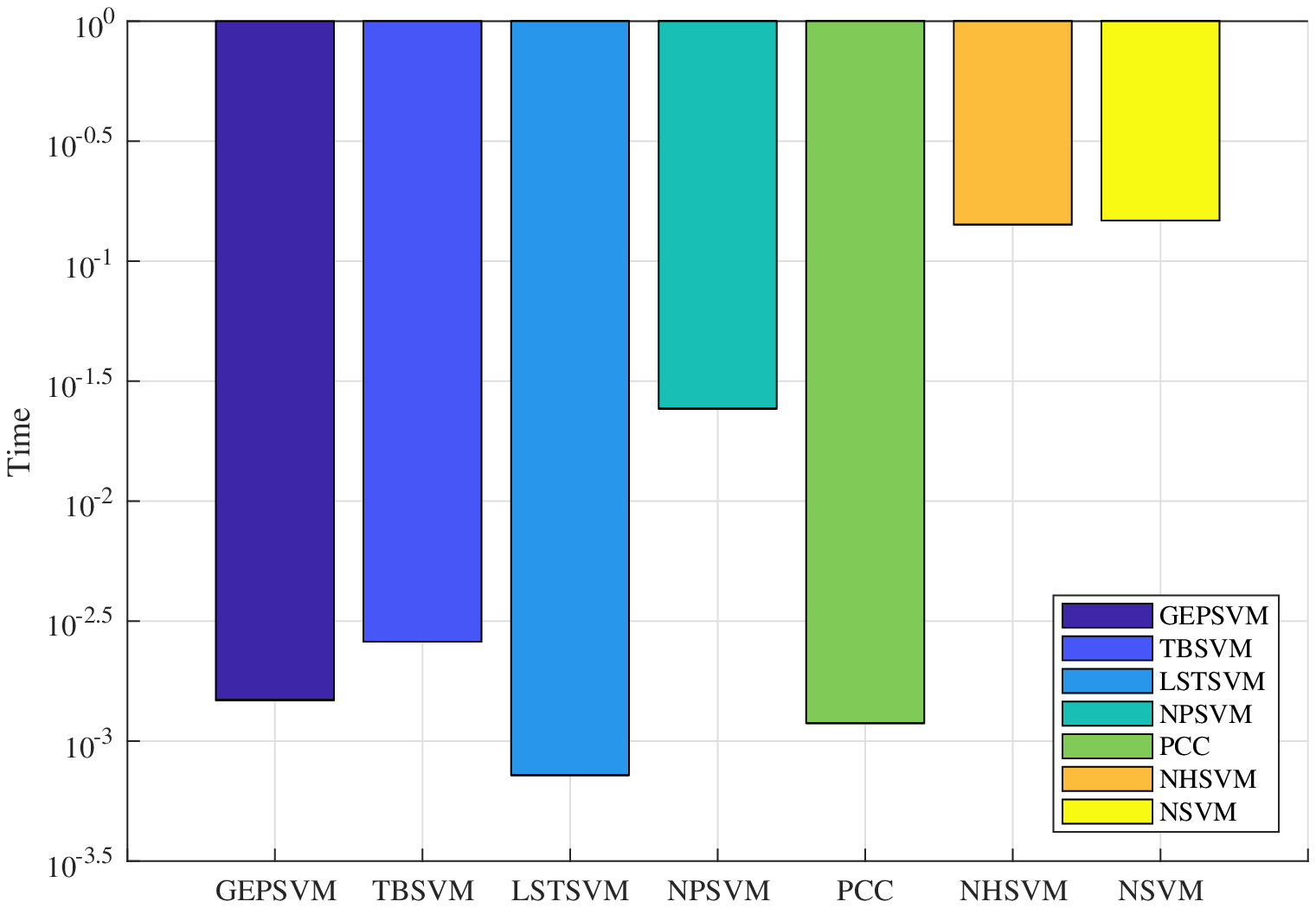}}}
\resizebox*{4.9cm}{!}{
\subfigure[Vowel]{\includegraphics[width=0.22\textheight]{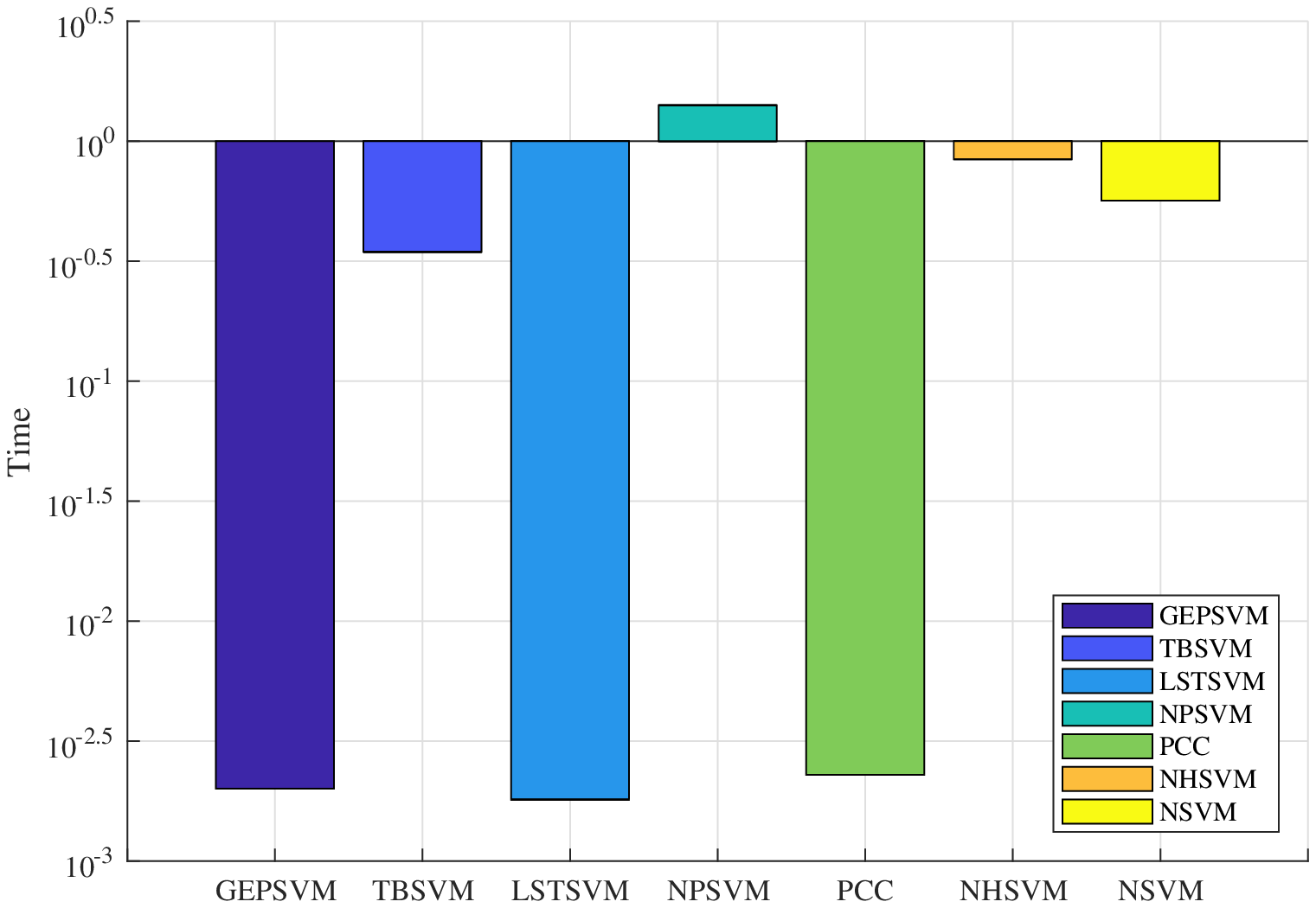}}}
\resizebox*{4.9cm}{!}{
\subfigure[Pathbased]{\includegraphics[width=0.22\textheight]{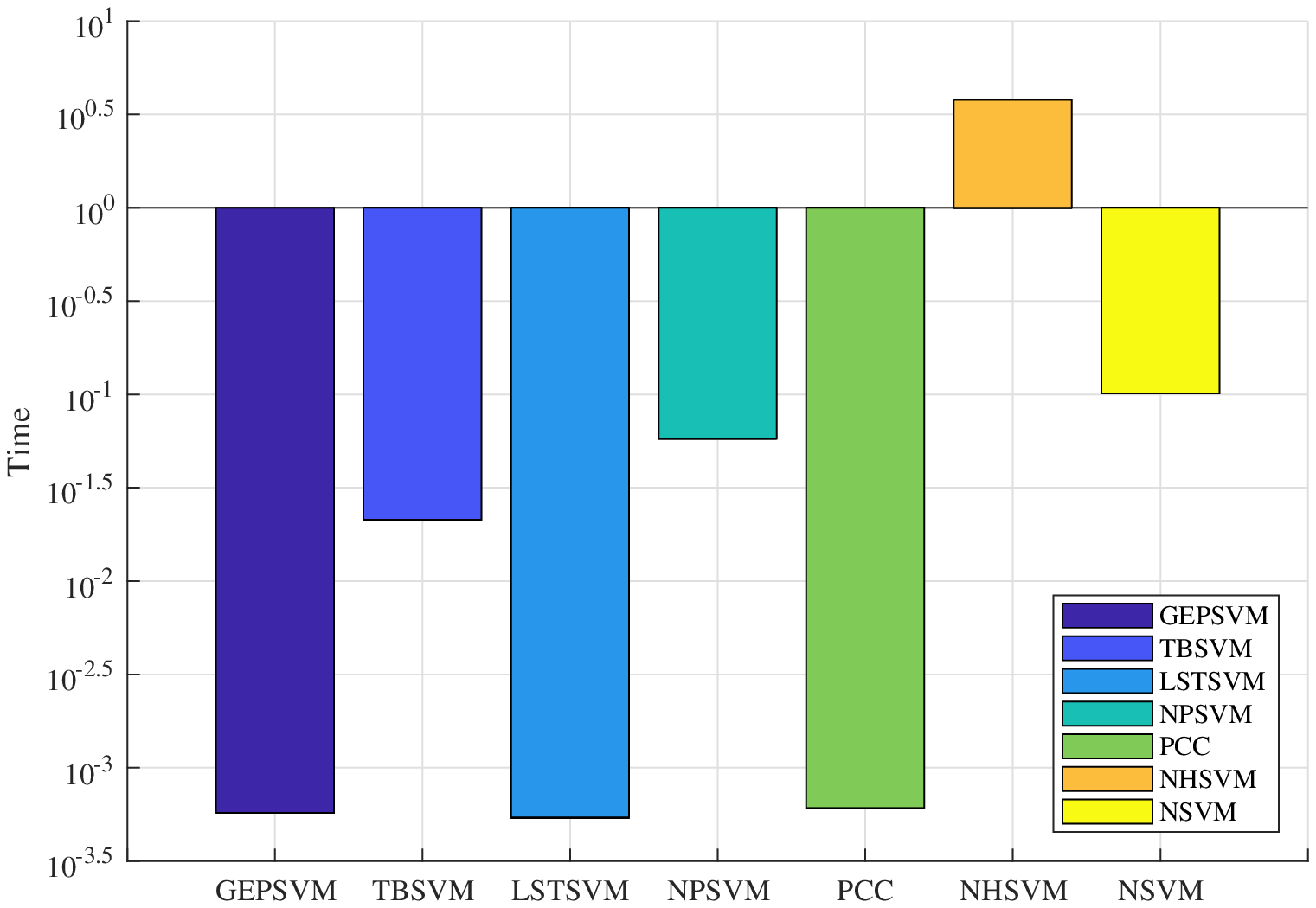}}}
\resizebox*{4.9cm}{!}{
\subfigure[Haberman]{\includegraphics[width=0.22\textheight]{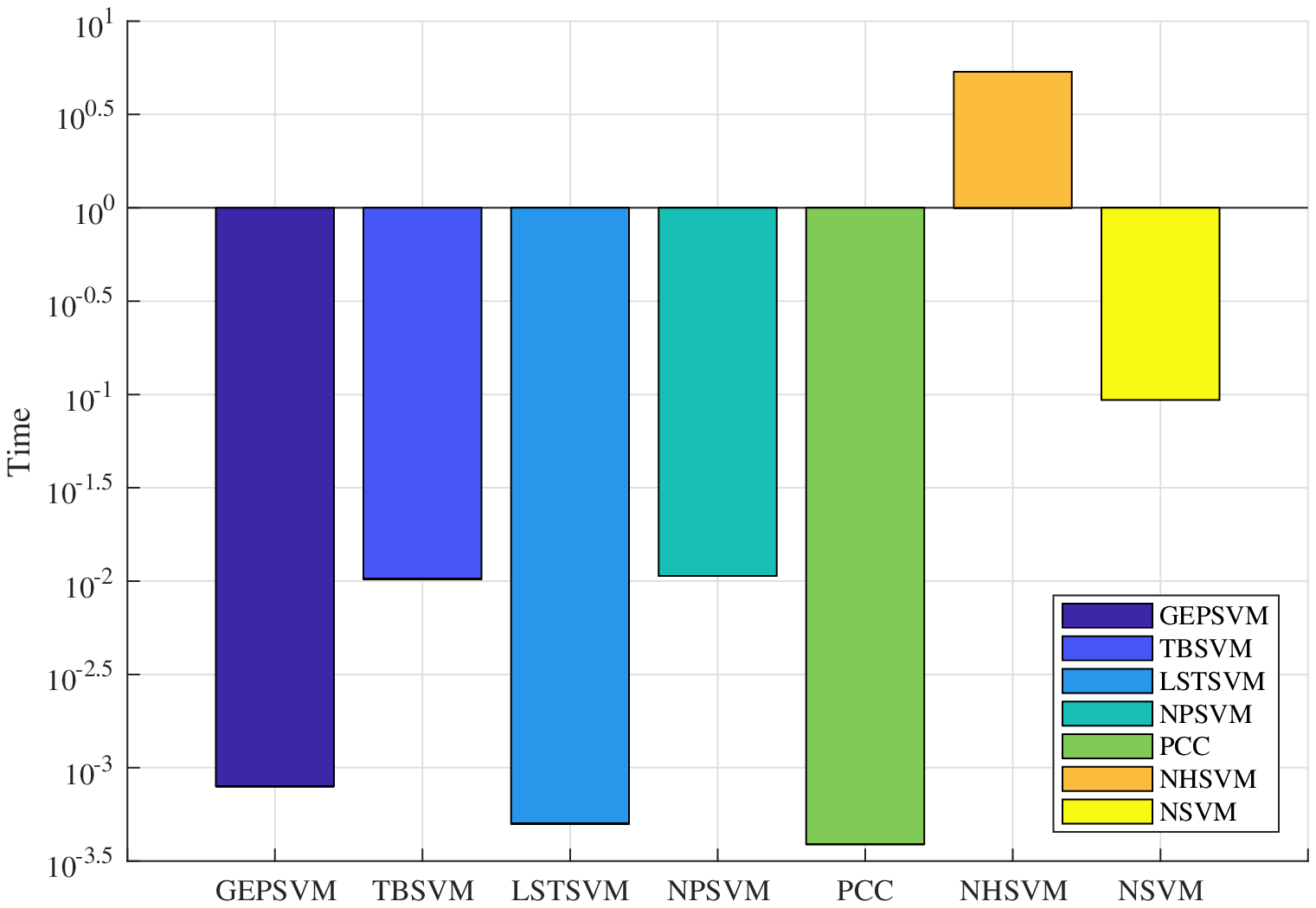}}}
\resizebox*{4.9cm}{!}{
\subfigure[Heartc]{\includegraphics[width=0.22\textheight]{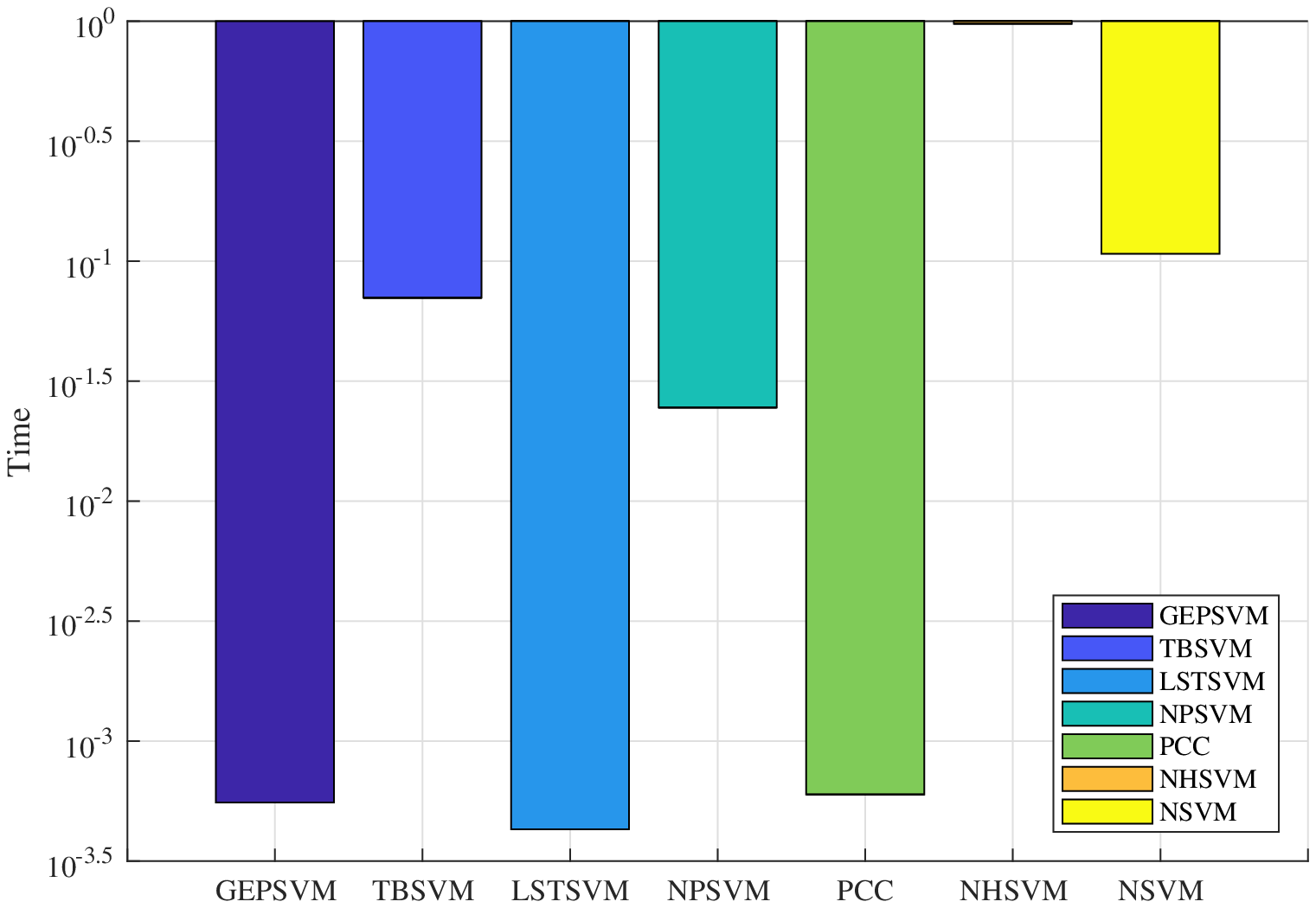}}}
\resizebox*{4.9cm}{!}{
\subfigure[Spanbas]{\includegraphics[width=0.22\textheight]{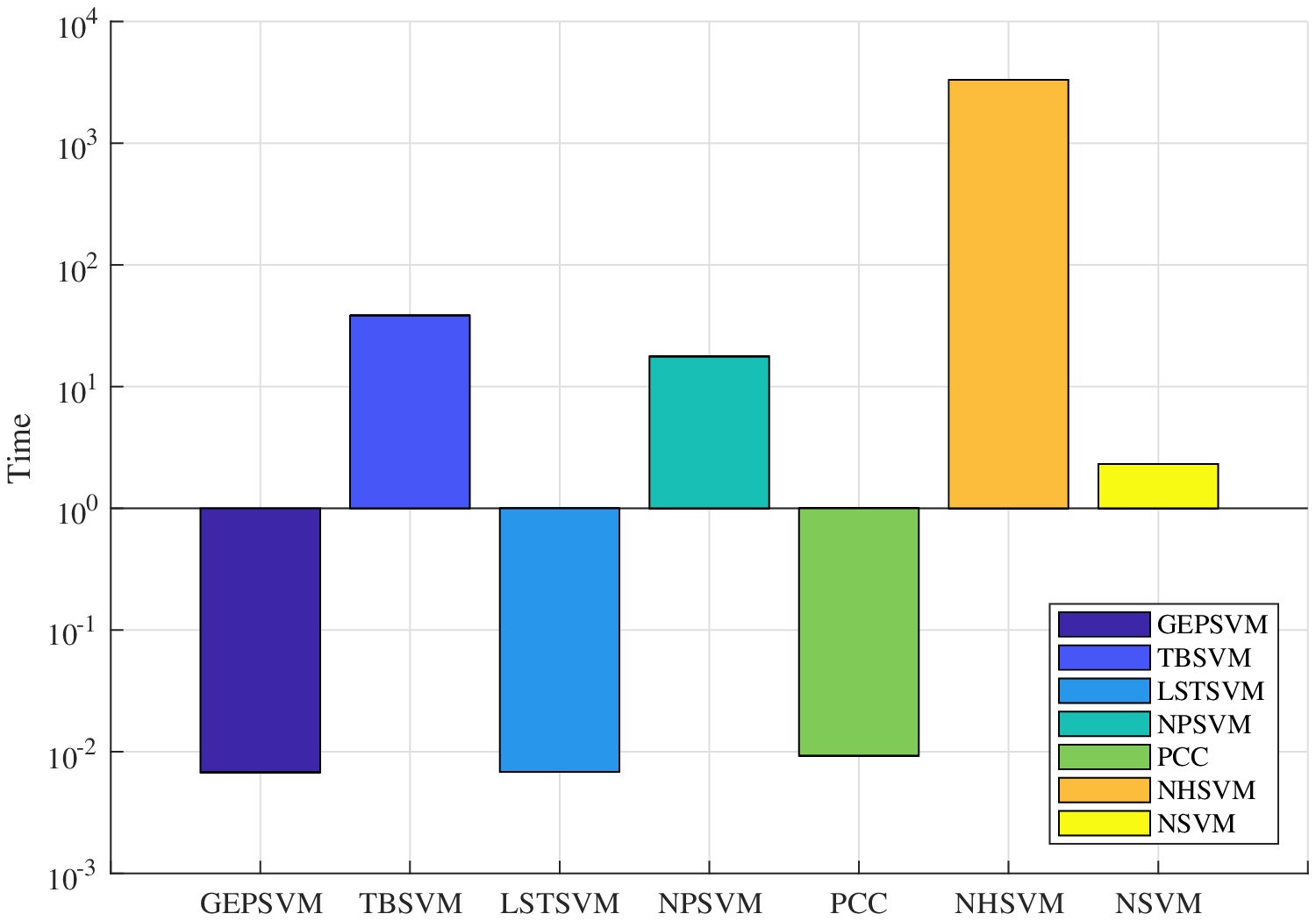}}}
\caption{Comparison of computation time of linear methods on UCI data sets.} \label{figUCItime}}
\end{center}
\end{figure*}

\begin{figure*}
\begin{center}{
\resizebox*{4.9cm}{!}{
\subfigure[Iris]{\includegraphics[width=0.22\textheight]{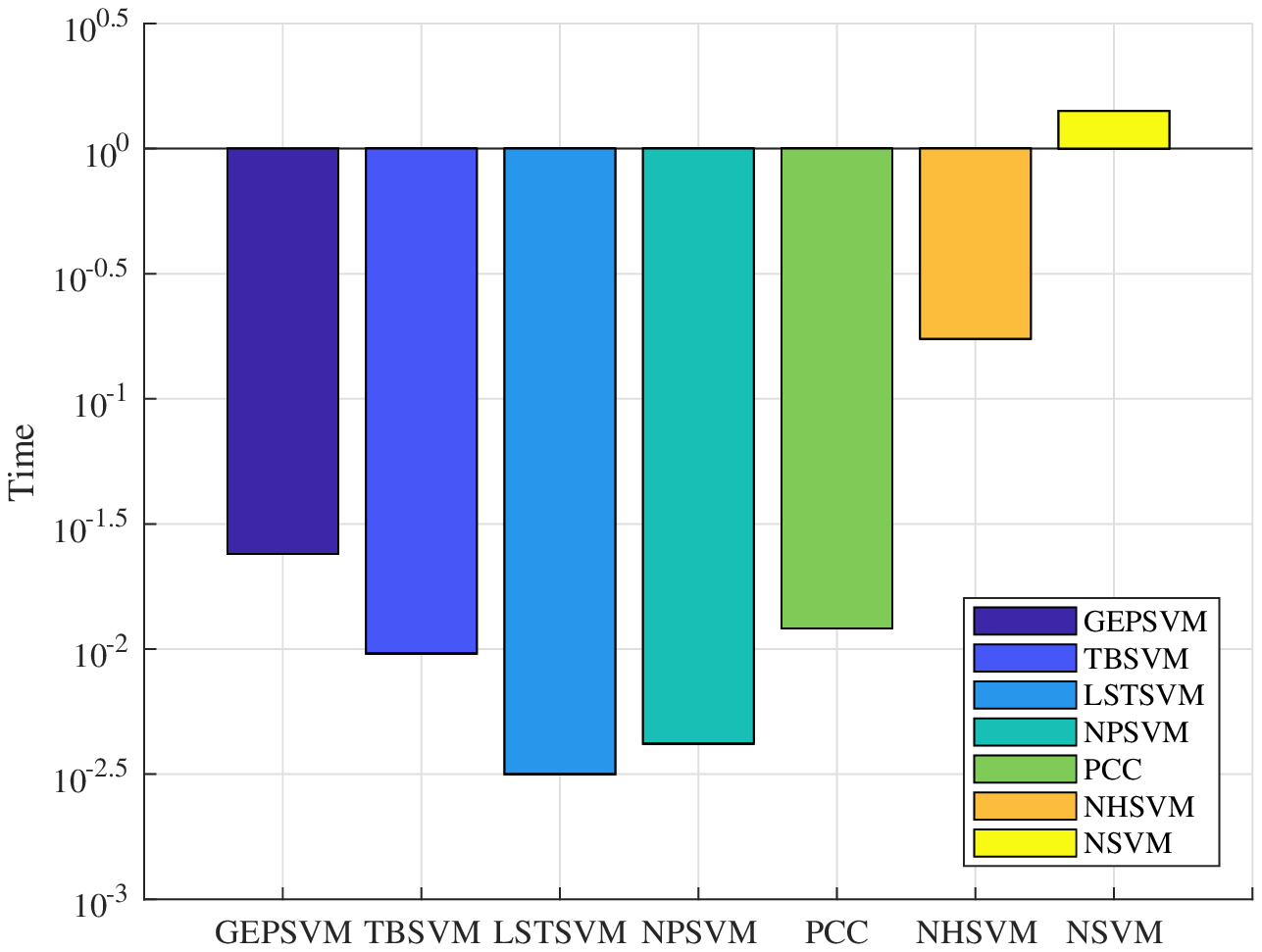}}}
\resizebox*{4.9cm}{!}{
\subfigure[Dermatology]{\includegraphics[width=0.22\textheight]{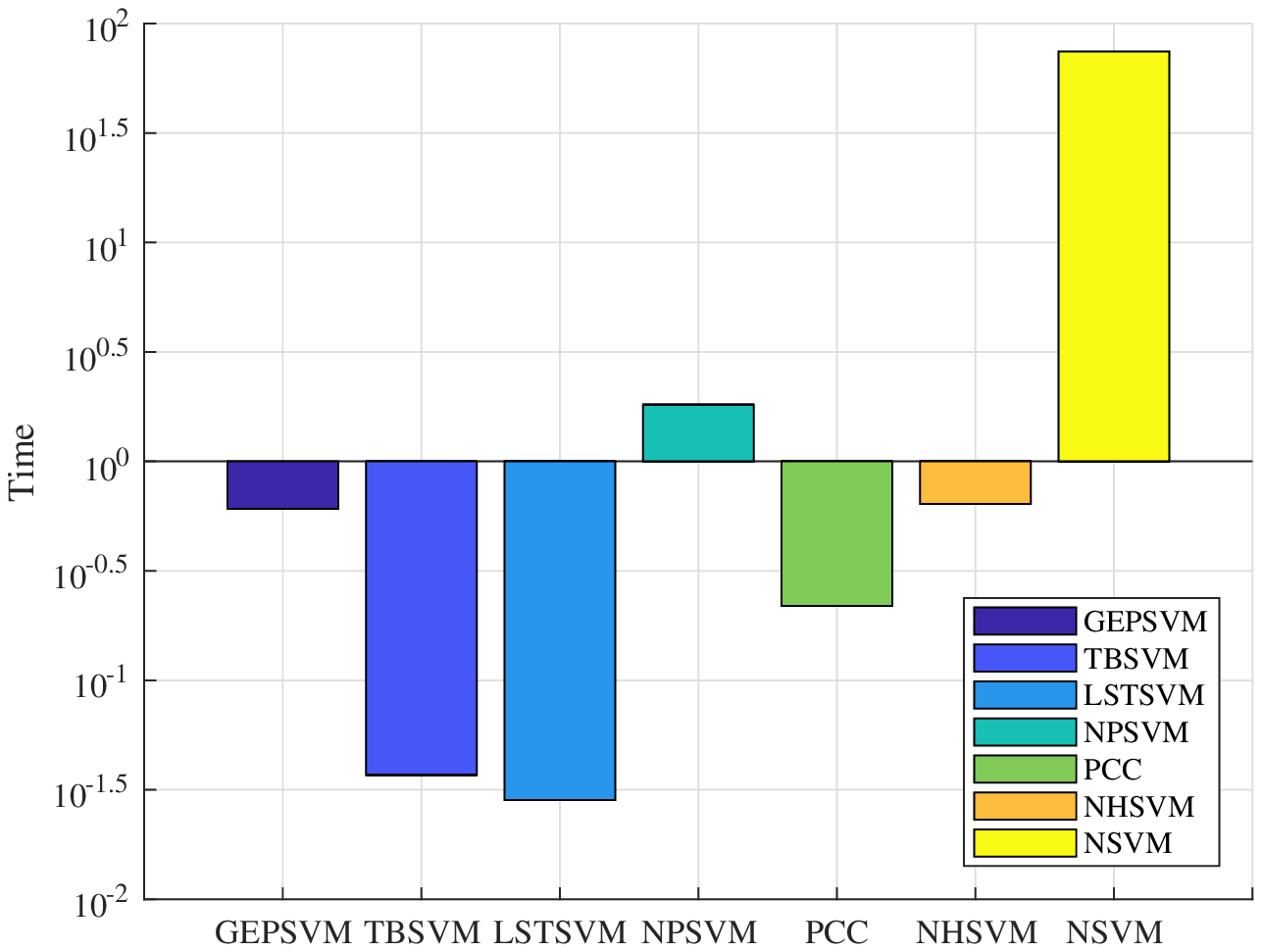}}}
\resizebox*{4.9cm}{!}{
\subfigure[Seeds]{\includegraphics[width=0.22\textheight]{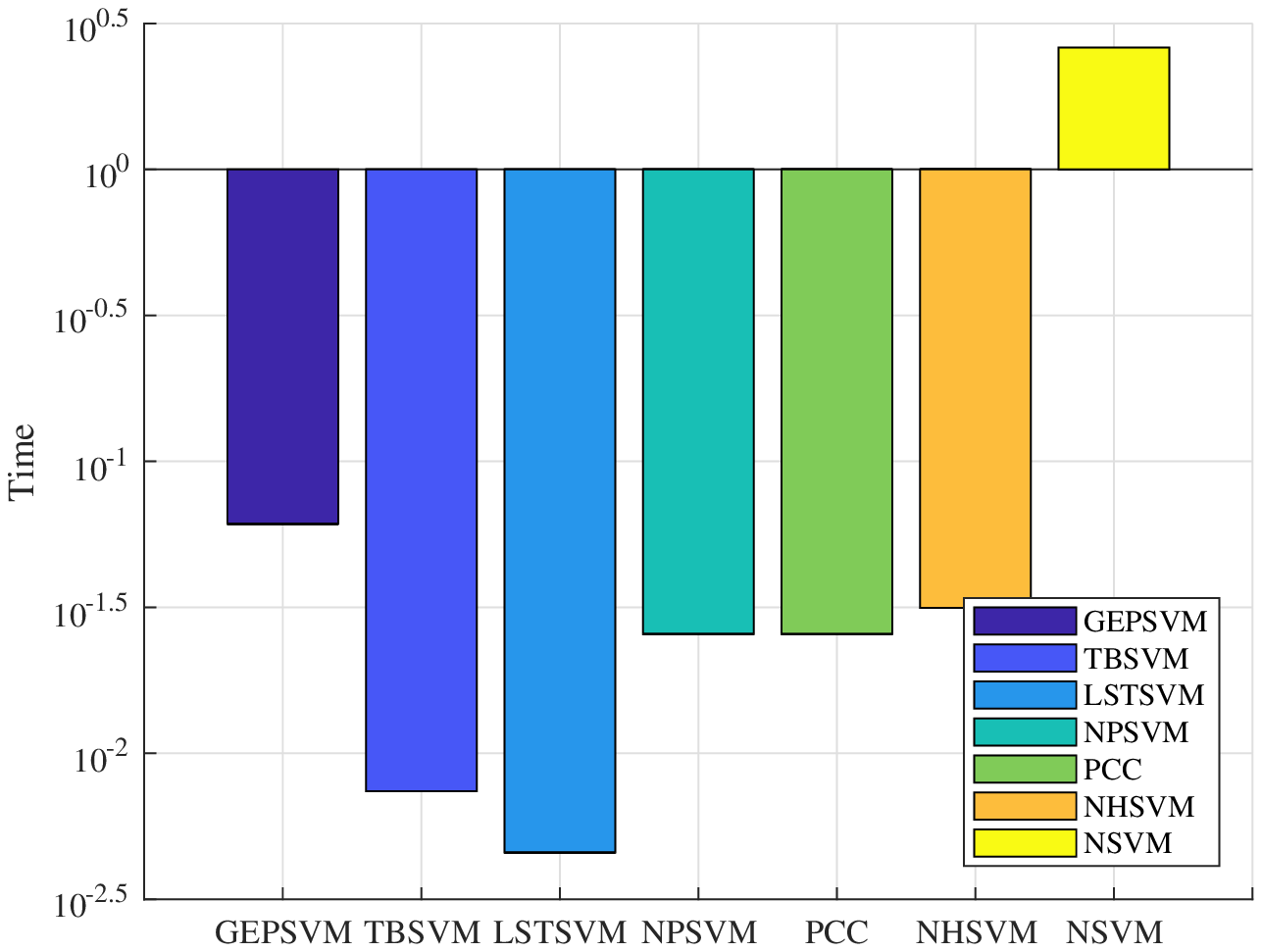}}}
\resizebox*{4.9cm}{!}{
\subfigure[Wine]{\includegraphics[width=0.22\textheight]{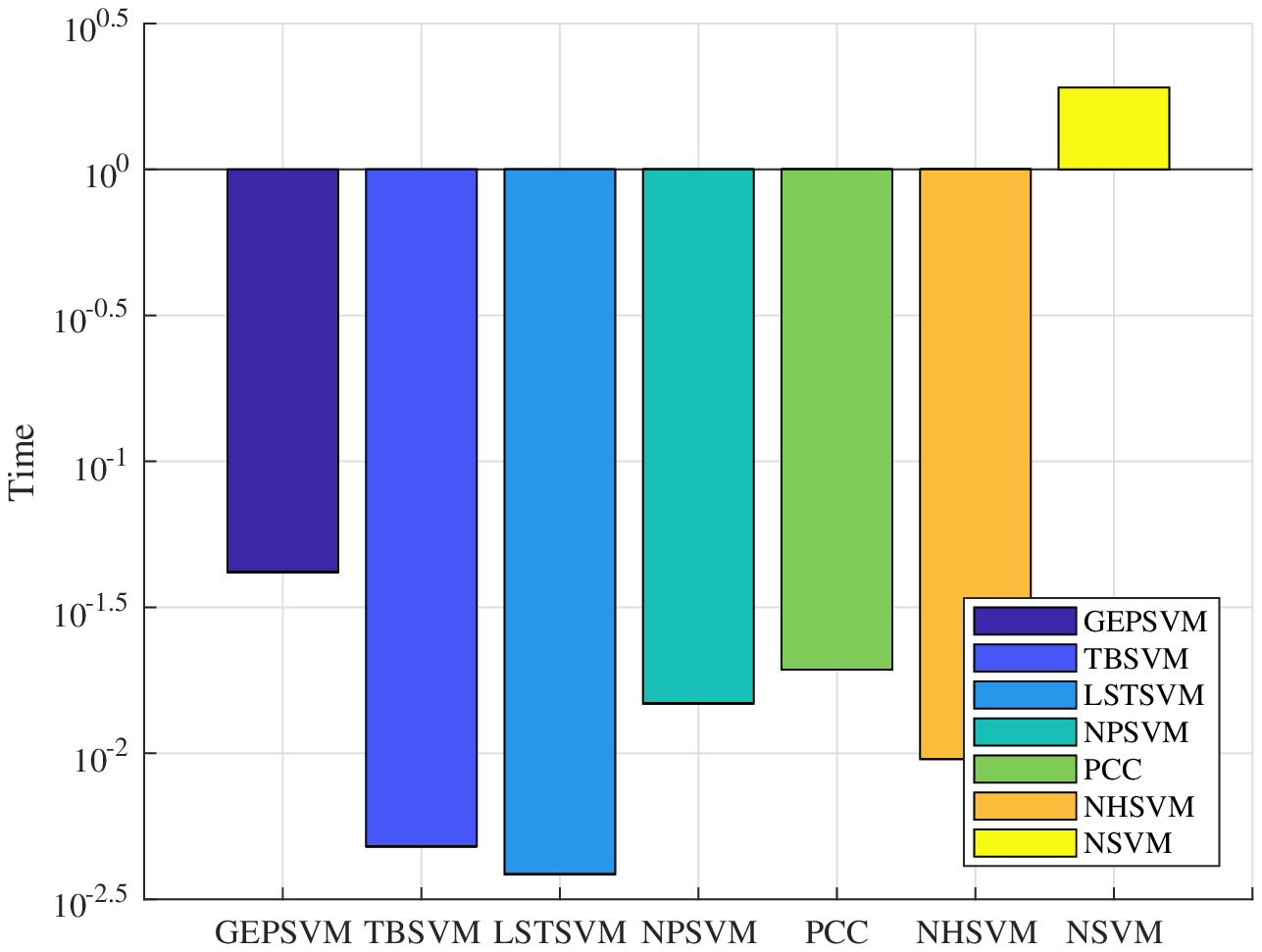}}}
\resizebox*{4.9cm}{!}{
\subfigure[Pathbased]{\includegraphics[width=0.22\textheight]{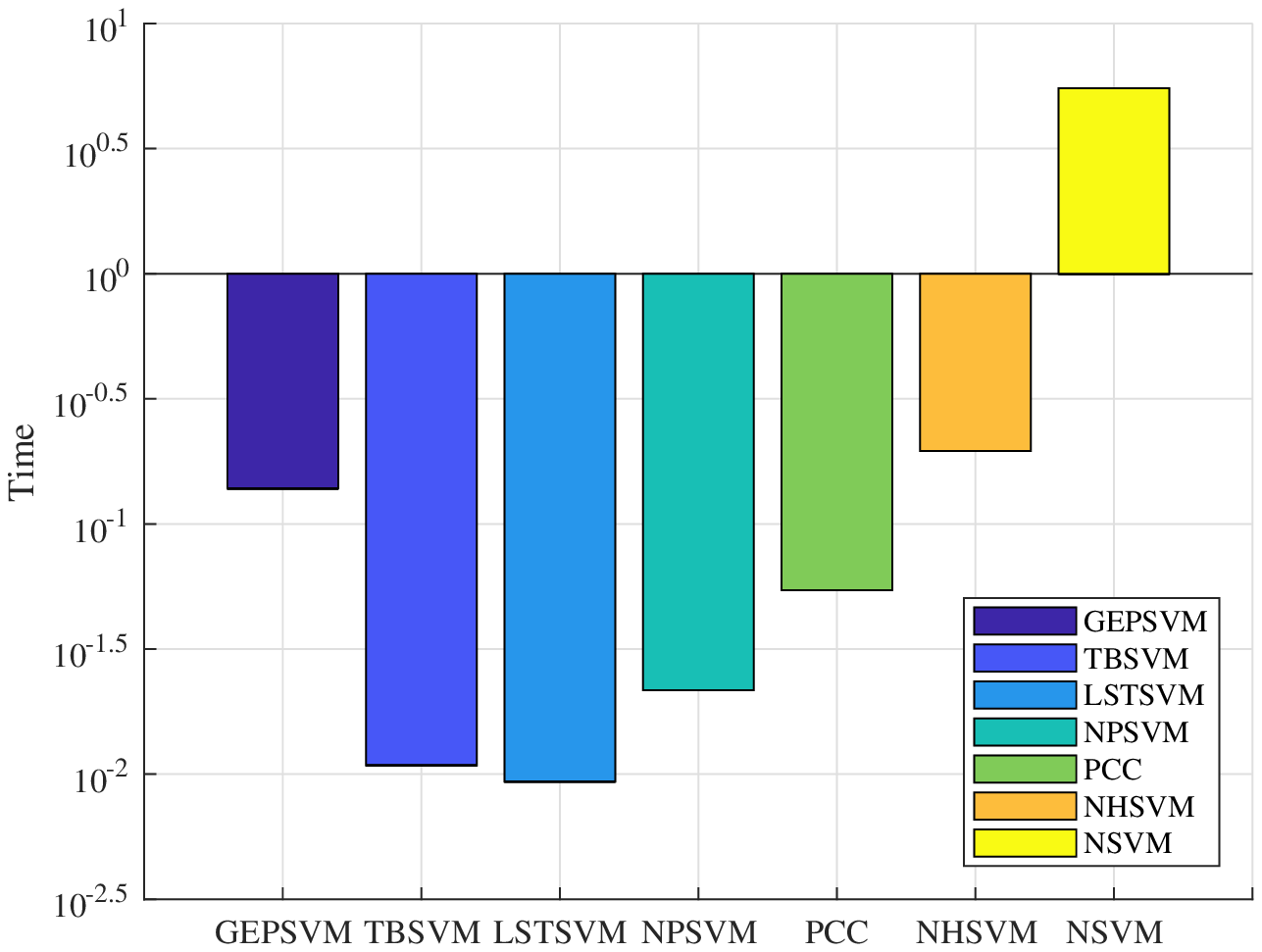}}}
\resizebox*{4.9cm}{!}{
\subfigure[WPBC]{\includegraphics[width=0.22\textheight]{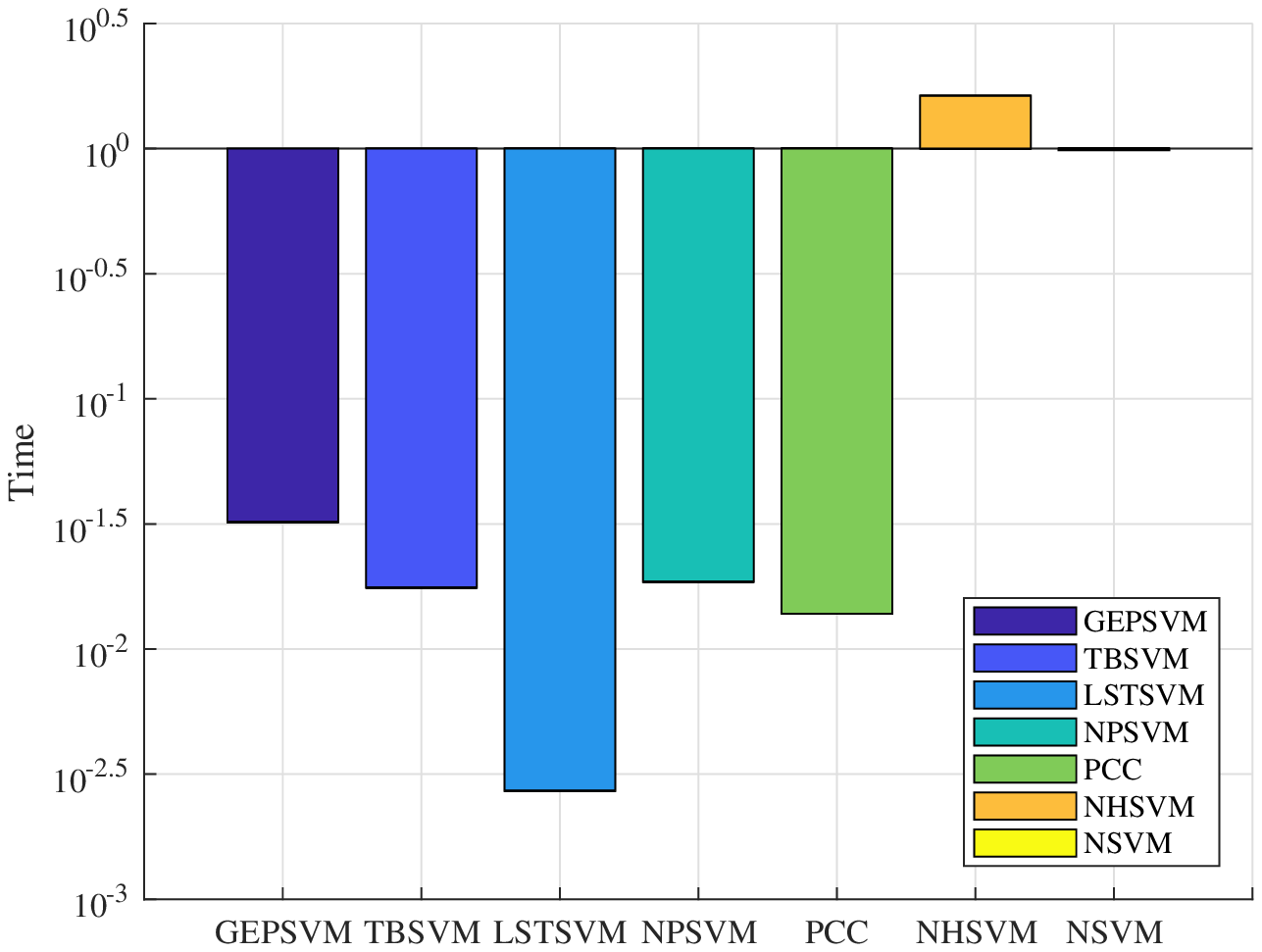}}}
\resizebox*{4.9cm}{!}{
\subfigure[Haberman]{\includegraphics[width=0.22\textheight]{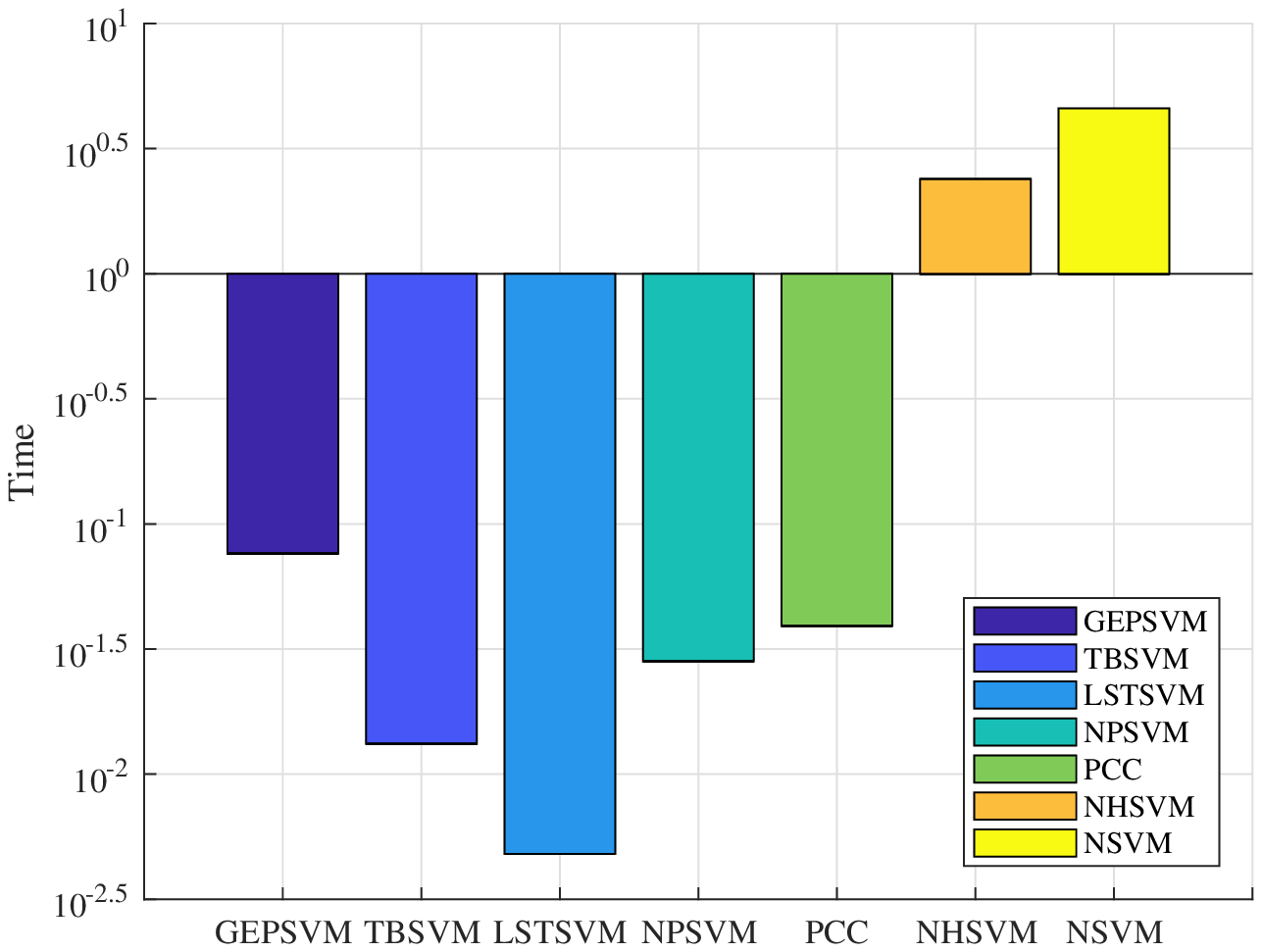}}}
\resizebox*{4.9cm}{!}{
\subfigure[Heartc]{\includegraphics[width=0.22\textheight]{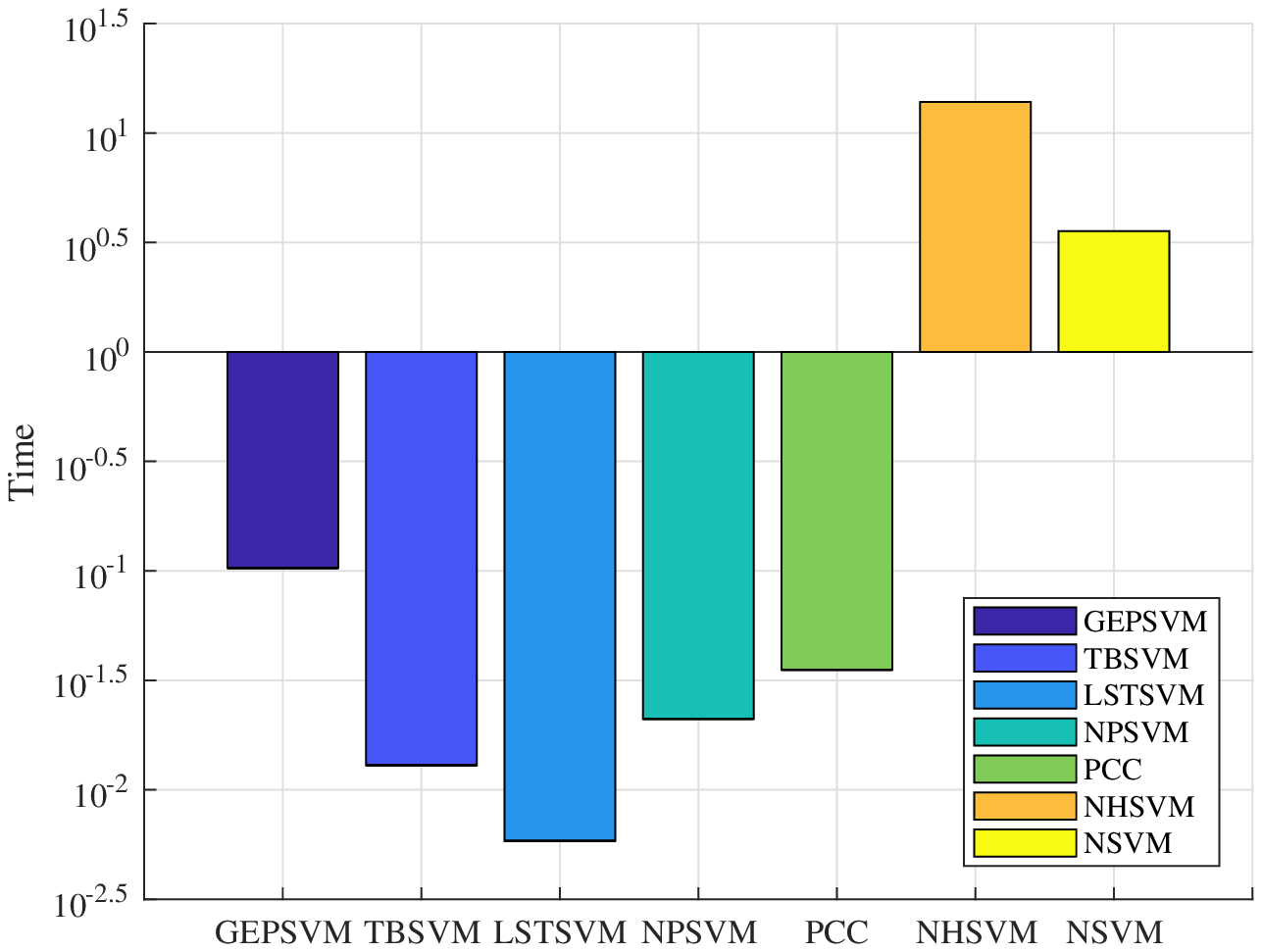}}}
\resizebox*{4.9cm}{!}{
\subfigure[Hepatitis]{\includegraphics[width=0.22\textheight]{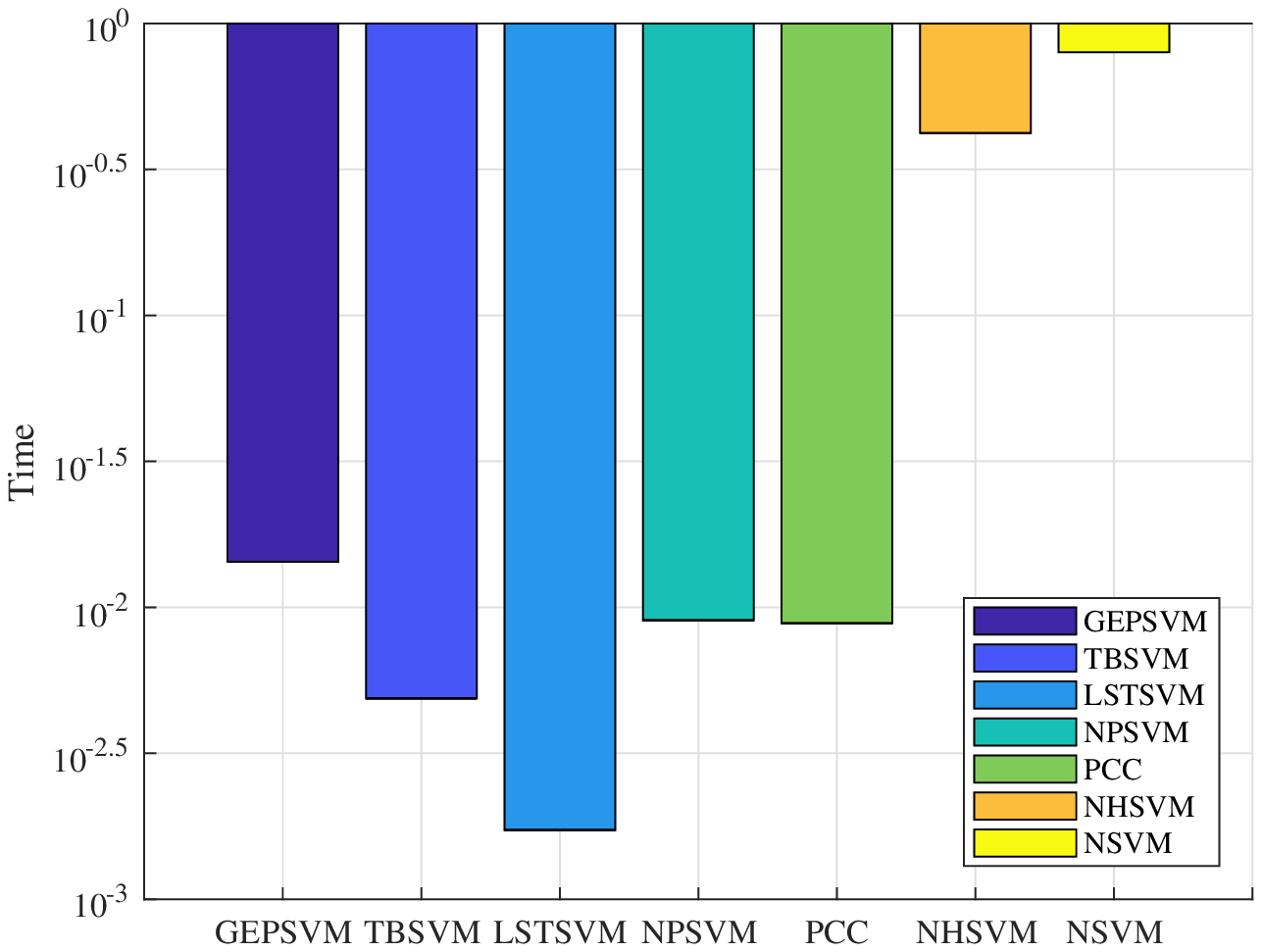}}}
\caption{Comparison of computation time of nonlinear methods on UCI data sets.} \label{figUCItimenon}}
\end{center}
\end{figure*}

Now we give the computation time of each method on the UCI data set for both linear and nonlinear cases. Here the iteration number is set to $20$ for NSVM. 
Fig.\ref{figUCItime} demonstrates the computation time for linear classifiers. From the figure, we see GEPSVM, LSTSVM and PCC have the fastest computation speed and comparable to each other. TBSVM and NPSVM run slower than GEPSVM, LSTSVM and PCC, and NHSVM and NSVM have the lowest computation speed on most of the data sets. However, on relatively large scale data, for example, ``DNA" and ``Spanbas" data, NPSVM and NHSVM cost more computation time than our NSVM.
For the nonlinear results in Fig.\ref{figUCItimenon},
we see all the methods run longer than their linear counterparts. Compared to the linear case, the computation time of GEPSVM, PCC and our NSVM grow relatively rapidly comparing to TBSVM, LSTWSVM and NPSVM.

\section{Conclusions}\label{conclusion}
Nonparallel support vector machines for classification problem is considered in this paper, and two general models for two types of NSVMs are proposed. The first type model constructs each hyperplanes separately. It solves a series of small optimization problems, but it is hard to measure the loss of each sample. The second type model constructs all hyperplanes simultaneously, and it solves one big optimization problem with the ascertained loss of each sample. Based on the second type model, we construct a max-min distance-based nonparallel support vector machine (NSVM). Experimental results on benchmark data sets show the advantages of our NSVM. Extending NSVMs to other machine learning problems with this two types of frameworks is interesting. In addition, designing efficient solving algorithms for the optimization problem of the second type NSVMs is also worth considering. The corresponding Matlab code of the paper can be downloaded from http://www.optimal-group.org/Resources/Code/NSVM.html.

\section*{Appendix}\label{Appendix}
\textbf{Proof of Theorem \ref{thm}:}
We first prove that the sequence $\{h_{t-1}({\overline{w}}^{t})+\frac{L}{2}\|{\overline{w}}^{t}-{\overline{w}}^{t-1}\|^2\}$
is nonascending.
Since $f_t(\overline{w})$ is strongly convex, and ${\overline{w}}^{t+1}$ is the global minimizer of $f_t(\overline{w})$, we have
\begin{align}\label{ftIneq}
\begin{split}
f_t(\overline{w}^{t+1})&\leq f_t(\overline{w}^{t})-(\overline{w}^{t+1}-\overline{w}^{t})^{\top}(G+\frac{L}{2}I)(\overline{w}^{t+1}-\overline{w}^{t})\\
&\leq f_t(\overline{w}^{t})-(\lambda_{\min}+\frac{L}{2})\|\overline{w}^{t+1}-\overline{w}^{t}\|^2.
\end{split}
\end{align}
Then
\begin{align*}
\begin{split}
&h_t(\overline{w}^{t+1})\\
=&\left(\frac{L}{2}\|\overline{w}^{t+1}\|^2+(\overline{w}^{t+1})^{\top} G\overline{w}^{t+1}\right)\\
&-\left(\frac{L}{2}\|\overline{w}^{t+1}\|^2+(\overline{w}^{t+1})^{\top}H^{t} \overline{w}^{t+1}\right)\\
=&\frac{L}{2}\|\overline{u}^{t}\|^2+\langle Lu^t,\overline{w}^{t+1}-u^t\rangle+\frac{L}{2}\|\overline{w}^{t+1}-u^t\|^2\\
&+(\overline{w}^{t+1})^{\top} G\overline{w}^{t+1}-\left(\frac{L}{2}\|\overline{w}^{t+1}\|^2+(\overline{w}^{t+1})^{\top}H^{t} \overline{w}^{t+1}\right)\\
\leq&\frac{L}{2}\|\overline{u}^{t}\|^2+\langle Lu^t,\overline{w}^{t+1}-u^t\rangle+\frac{L}{2}\|\overline{w}^{t+1}-u^t\|^2+(\overline{w}^{t+1})^{\top} G\overline{w}^{t+1}\\
&-\left(\frac{L}{2}\|\overline{w}^{t}\|^2+(\overline{w}^{t})^{\top}H^{t} \overline{w}^{t}\right)-\langle L\overline{w}^{t}+2H^t\overline{w}^{t},\overline{w}^{t+1}-\overline{w}^{t}\rangle\\
\leq&\frac{L}{2}\|\overline{u}^{t}\|^2+\langle Lu^t,\overline{w}^{t+1}-u^t\rangle+\frac{L}{2}\|\overline{w}^{t}-u^t\|^2+(\overline{w}^{t})^{\top} G\overline{w}^{t}\\
&-\left(\frac{L}{2}\|\overline{w}^{t}\|^2+(\overline{w}^{t})^{\top}H^{t} \overline{w}^{t}\right)-\langle L\overline{w}^{t},\overline{w}^{t+1}-\overline{w}^{t}\rangle\\
&-(\lambda_{\min}+\frac{L}{2})\|\overline{w}^{t+1}-\overline{w}^{t}\|^2\\
=&\big[\frac{L}{2}\|\overline{u}^{t}\|^2+
\langle Lu^t,\overline{w}^{t}-u^t\rangle-\langle Lu^t,\overline{w}^t\rangle
+\langle Lu^t,\overline{w}^{t+1}\rangle\\
&-\langle L\overline{w}^{t},\overline{w}^{t+1}-\overline{w}^{t}\rangle-\lambda_{\min}\|\overline{w}^{t+1}-\overline{w}^{t}\|^2\big]\\
&+\big[(\overline{w}^{t})^{\top} G\overline{w}^{t}-\left(\frac{L}{2}\|\overline{w}^{t}\|^2+(\overline{w}^{t})^{\top}H^{t} \overline{w}^{t}\right)\\
&+\frac{L}{2}\|\overline{w}^{t}-u^t\|^2-\frac{L}{2}\|\overline{w}^{t+1}-\overline{w}^{t}\|^2\big]\\
\end{split}
\end{align*}
\begin{align}\label{hIneq}
\begin{split}
\leq&\big[-\langle Lu^t,\overline{w}^t\rangle
+\langle Lu^t,\overline{w}^{t+1}\rangle-\langle L\overline{w}^{t},\overline{w}^{t+1}-\overline{w}^{t}\rangle\\
&-\lambda_{\min}\|\overline{w}^{t+1}-\overline{w}^{t}\|^2\big]+\big[\frac{L}{2}\|\overline{w}^{t}\|^2+(\overline{w}^{t})^{\top} G\overline{w}^{t}\\
&-\left(\frac{L}{2}\|\overline{w}^{t}\|^2+(\overline{w}^{t})^{\top}H^{t} \overline{w}^{t}\right)+\frac{L}{2}\|\overline{w}^{t}-u^t\|^2\\
&-\frac{L}{2}\|\overline{w}^{t+1}-\overline{w}^{t}\|^2\big]\\
=&-\left|\left|\frac{L}{2\sqrt{\lambda_{\min}}}(u^t-\overline{w}^t)-\sqrt{\lambda_{\min}}(\overline{w}^{t+1}-\overline{w}^{t})\right|\right|^2\\
&+(\overline{w}^{t})^{\top}(H^{t-1}-H^t) \overline{w}^{t}+\big[\frac{L}{2}\|\overline{w}^{t}\|^2+(\overline{w}^{t})^{\top} G\overline{w}^{t}\\
&-\left(\frac{L}{2}\|\overline{w}^{t}\|^2+(\overline{w}^{t})^{\top}H^{t-1} \overline{w}^{t}\right)+\left(\frac{L}{2}+\frac{L^2}{4\lambda_{\min}}\right)\|\overline{w}^{t}-u^t\|^2\\
&-\frac{L}{2}\|\overline{w}^{t+1}-\overline{w}^{t}\|^2\big]\\
\leq&h_{t-1}(\overline{w}^{t})+\left(\frac{L}{2}+\frac{L^2}{4\lambda_{\min}}\right)\beta_t^2\|\overline{w}^{t}-\overline{w}^{t-1}\|^2\\
&-\frac{L}{2}\|\overline{w}^{t+1}-\overline{w}^{t}\|^2.
\end{split}
\end{align}

The first inequality of \eqref{hIneq} follows from the first order approximation of $\frac{L}{2}\|\overline{w}^{t}\|^2+(\overline{w}^{t})^{\top}H \overline{w}^{t}$.
The second inequality of \eqref{hIneq} follows from \eqref{ftIneq}. The last inequality of \eqref{hIneq} derives from assumption 1.
The above inequality implies that
\begin{align}\label{htht-1}
\begin{split}
&\left[\frac{L}{2}-\left(\frac{L}{2}+\frac{L^2}{4\lambda_{\min}}\right)\beta_t^2\right]\|\overline{w}^{t}-\overline{w}^{t-1}\|^2\\
\leq&\left[h_{t-1}(\overline{w}^{t})+\frac{L}{2}\|\overline{w}^{t}-\overline{w}^{t-1}\|^2\right]-\left[h_{t}(\overline{w}^{t+1})+\frac{L}{2}\|\overline{w}^{t+1}-\overline{w}^{t}\|^2\right].
\end{split}
\end{align}
Take
$\beta_t\leq\sqrt{\frac{2\lambda_{\min}}{2\lambda_{\min}+L}}$. Then $\left[\frac{L}{2}-\left(\frac{L}{2}+\frac{L^2}{4\lambda_{\min}}\right)\beta_t^2\right]\|\overline{w}^{t}-\overline{w}^{t-1}\|^2\geq 0$ and hence the sequence
$\{h_{t-1}(\overline{w}^{t})+\frac{L}{2}\|\overline{w}^{t}-\overline{w}^{t-1}\|^2\}$ is nonascending.

Since $\overline{w}^{1}=\overline{w}^0$, by summing \eqref{htht-1} from 0 to $\infty$, we have
\begin{align}\label{sumeq}
\begin{split}
&\left[\frac{L}{2}-\left(\frac{L}{2}+\frac{L^2}{4\lambda_{\min}}\right)\right]\sum_{t=0}^{\infty}\beta_t^2\|\overline{w}^{t}-\overline{w}^{t-1}\|^2\\
\leq&h_{0}(\overline{w}^{1})
-\underset{t\rightarrow\infty}{\lim\inf}\left[h_{t}(\overline{w}^{t+1})+\frac{L}{2}\|\overline{w}^{t+1}-\overline{w}^{t}\|^2\right]\\
\leq&h_{0}(\overline{w}^{1}).
\end{split}
\end{align}
The last inequality of \eqref{sumeq} follows from assumption 2. Clearly, \eqref{sumeq} implies $\underset{t\rightarrow \infty}{\lim}\|{\overline{w}}^{t+1}-{\overline{w}}^{t}\|=0$.
\hfill$\square$

\section*{Acknowledgment}
This work is supported by the Hainan Provincial Natural Science Foundation of China (No.620QN234 and No.120RC449), National Natural Science Foundation of China (No.62066012, No.61703370, No.61866010 and No.11871183).


\begin{thebibliography}{}

\end{thebibliography}


\section*{References}
\begin{thebibliography}{99}


\bibitem{GEPSVM}
Mangasarian O L, Wild E W. Multisurface proximal support vector machine classification via generalized eigenvalues. IEEE Transactions on Pattern Analysis and Machine Intelligence, 2006, 28(1): 69-74.

\bibitem{TWSVMbook17}
Jayadeva, Khemchandani R, Chandra S. Twin support vector machines. Springer: Berlin, Germany, 2017.

\bibitem{Nitta03}
Nitta T. Solving the XOR problem and the detection of symmetry using a single complex-valued neuron. Neural Networks, 2003, 16(8): 1101-1105.

\bibitem{SVM}
Cortes C, Vapnik V. Support-vector networks. Machine Learning, 1995, 20(3): 273-297.

\bibitem{SVMbook12}
Deng N Y, Tian Y J, Zhang C H. Support vector machines: theory, algorithms, and extensions. Boca Raton: CRC Press;2012.

\bibitem{LiYuang17}
Li Y, Yuan Y. Convergence analysis of two-layer neural networks with relu activation. Advances in Neural Information Processing Systems. 2017: 597-607.

\bibitem{IGEPSVM}
Shao Y H, Deng N Y, Chen W J, et al. Improved generalized eigenvalue proximal support vector machine. IEEE Signal Processing Letters, 2013, 20(3): 213-216.

\bibitem{TWSVM}
Jayadeva, Khemchandani R, Chandra S. Twin support vector machines for pattern classification. IEEE Transactions on Pattern Analysis and Machine Intelligence, 2007, 29(5): 905-910.

\bibitem{TBSVM}
Shao Y H, Zhang C H, Wang X B, et al. Improvements on twin support vector machines. IEEE Transactions on Neural Networks, 2011, 22(6): 962-968

\bibitem{NPSVM}
Tian Y, Qi Z, Ju X, et al. Nonparallel support vector machines for pattern classification. IEEE Transactions on Cybernetics, 2014, 44(7): 1067-1079.

\bibitem{INPSVM}
Liu L, Chu M, Gong R, et al. An improved nonparallel support vector machine. IEEE Transactions on Neural Networks and Learning Systems, 2020.

\bibitem{BFHC}
Cevikalp H. Best fitting hyperplanes for classification. IEEE Transactions on Pattern Analysis and Machine Intelligence, 2017, 39(6): 1076-1088.

\bibitem{L1NPSVM}
Li C N, Shao Y H, Deng N Y. Robust L1-norm nonparallel proximal support vector machine. Optimization, 2016, 65(1):169-183.

\bibitem{L1GEPSVM}
Yan H, Ye Q, Zhang T, et al. L1-norm GEPSVM classifier based on an effective iterative algorithm for classification. Neural Processing Letters, 2017: 1-26.

\bibitem{LpNPSVM}
Sun X Q, Chen Y J, Shao Y H, Li C N, Wang C H. Robust nonparallel proximal support vector machine with Lp-norm regularization. IEEE Access, 2018, 6: 20334-20347.

\bibitem{LSTSVM}
Kumar M A, Gopal M. Least squares twin support vector machines for pattern classification. Expert Systems with Applications, 2009, 36(4): 7535-7543

\bibitem{ILSTSVM}
Xu Y, Xi W, Lv X, et al. An improved least squares twin support vector machine. Journal of Information and Computational Science, 2012, 9(4): 1063-1071.

\bibitem{PinTWSVM}
Xu Y, Yang Z, Pan X. A novel twin support-vector machine with pinball loss. IEEE Transactions on Neural Networks and Learning Systems, 2017, 28(2): 359-370.

\bibitem{RNPSVM}
Liu D, Shi Y, Tian Y. Ramp loss nonparallel support vector machine for pattern classification. Knowledge-Based Systems, 2015, 85: 224-233.

\bibitem{NHSVM}
Shao Y H, Chen W J, Deng N Y. Nonparallel hyperplane support vector machine for binary classification problems. Information Sciences, 2014, 263: 22-35.

\bibitem{NHSVMSOC}
Carrasco M, L\'{a}pez J, Maldonado S. A second-order cone programming formulation for nonparallel hyperplane support vector machine. Expert Systems with Applications, 2016, 54: 95-104.

\bibitem{LSNHSVM}
Kumar D, Thakur M. All-in-one multicategory least squares nonparallel hyperplanes support vector machine. Pattern Recognition Letters, 2018, 105: 165-174.

\bibitem{PCC}
Shao Y H, Deng N Y, Chen W J. A proximal classifier with consistency. Knowledge-Based Systems, 2013, 49: 171-178.



\bibitem{MBSVM13}
Yang Z X, Shao Y H, Zhang X S. Multiple birth support vector machine for multi-class classification. Neural Computing and Applications, 2013, 22(1): 153-161.

\bibitem{BSTWSVM13}
Shao Y H, Chen W J, Huang W B, et al. The best separating decision tree twin support vector machine for multi-class classification. Procedia Computer Science, 2013, 17: 1032-1038.

\bibitem{MNHSVM15}
Ju X, Tian Y, Liu D, et al. Nonparallel hyperplanes support vector machine for multi-class classification. Procedia Computer Science, 2015, 51: 1574-1582.

\bibitem{MLSTMSVMDAG16}
Zhang X, Ding S, Sun T. Multi-class LSTMSVM based on optimal directed acyclic graph and shuffled frog leaping algorithm. International Journal of Machine Learning and Cybernetics, 2016, 7(2): 241-251.

\bibitem{MTSVM16}
Tomar D, Agarwal S. Multi-class twin support vector machine for pattern classification. Proceedings of 3rd International Conference on Advanced Computing, Networking and Informatics. Springer, New Delhi, 2016: 97-110.

\bibitem{MIBSVM17}
Zhang X, Ding S, Xue Y. An improved multiple birth support vector machine for pattern classification. Neurocomputing, 2017, 225: 119-128.

\bibitem{WLMBSVM17}
Ding S, Zhang X, An Y, et al. Weighted linear loss multiple birth support vector machine based on information granulation for multi-class classification. Pattern Recognition, 2017, 67: 32-46.

\bibitem{MILSTWSVM18}
de Lima M D, Costa N L, Barbosa R. Improvements on least squares twin multi-class classification support vector machine. Neurocomputing, 2018, 313: 196-205.

\bibitem{RastogiSaigal18}
Rastogi R, Saigal P, Chandra S. Angle-based twin parametric-margin support vector machine for pattern classification. Knowledge-Based Systems, 2018, 139: 64-77.

\bibitem{SaigalRastogi19}
Saigal P, Chandra S, Rastogi R. Multi-category ternion support vector machine. Engineering Applications of Artificial Intelligence, 2019, 85: 229-242.

\bibitem{LiuGong20}
Liu L , Gong R , Chu M , et al. Nonparallel support vector machine with large margin distribution for pattern classification. Pattern Recognition, 2020, 106: 107374.

\bibitem{MVSVM11}
Ye Q, Zhao C, Ye N, et al. Multi-weight vector projection support vector machines. Pattern Recognition Letters, 2010, 31(13): 2006-2011.

\bibitem{PTSVM11}
Chen X, Yang J, Ye Q, et al. Recursive projection twin support vector machine via within-class variance minimization. Pattern Recognition, 2011, 44(10-11): 2643-2655.

\bibitem{RLSPTSVM12}
Shao Y H, Deng N Y, Yang Z M. Least squares recursive projection twin support vector machine for classification. Pattern Recognition, 2012, 45(6): 2299-2307.

\bibitem{RPTSVM13}
Shao Y H, Wang Z, Chen W J, et al. A regularization for the projection twin support vector machine. Knowledge-Based Systems, 2013, 37: 203-210.

\bibitem{RLSPTSVMn14}
Ding S, Hua X. Recursive least squares projection twin support vector machines for nonlinear classification. Neurocomputing, 2014, 130: 3-9.


\bibitem{MPTSVM16}
Li C N, Huang Y F, Wu H J, et al. Multiple recursive projection twin support vector machine for multi-class classification. International Journal of Machine Learning and Cybernetics, 2016, 7(5): 729-740.

\bibitem{RichhariyaTanveer20}
Richhariya B, Tanveer M, Alzheimer's Disease Neuroimaging Initiative. Least squares projection twin support vector clustering (LSPTSVC). Information Sciences, 2020, 533: 1-23.

\bibitem{Peng10}
Peng X. TSVR: an efficient twin support vector machine for regression. Neural Networks, 2010, 23(3): 365-372.


\bibitem{ShaoZhangYang13}
Shao Y H, Zhang C H, Yang Z M, et al. An $\varepsilon$-twin support vector machine for regression. Neural Computing and Applications, 2013, 23(1): 175-185.

\bibitem{BalasundaramTanveer13}
Balasundaram S, Tanveer M. On Lagrangian twin support vector regression. Neural Computing and Applications, 2013, 22(1): 257-267.

\bibitem{YangHuaShao16}
Yang Z M, Hua X Y, Shao Y H, et al. A novel parametric-insensitive nonparallel support vector machine for regression. Neurocomputing, 2016, 171: 649-663.

\bibitem{WangShiNiu17}
Wang H, Shi Y, Niu L, et al. Nonparallel support vector ordinal regression. IEEE Transactions on Cybernetics, 2017, 47(10): 3306-3317.

\bibitem{YeShaoDeng17}
Ye Y F, Shao Y H, Deng N Y, et al. Robust Lp-norm least squares support vector regression with feature selection. Applied Mathematics and Computation, 2017, 305: 32-52.

\bibitem{LiuShaoWang18}
Liu M Z, Shao Y H, Wang Z, et al. Minimum deviation distribution machine for large scale regression. Knowledge-Based Systems, 2018, 146: 167-180.

\bibitem{TangTianYang18}
Tang L, Tian Y, Yang C. Nonparallel support vector regression model and its SMO-type solver. Neural Networks, 2018, 105: 431-446.

\bibitem{Carrasco19}
Carrasco M, Julio L\'{o}pez, Sebasti\'{a}n Maldonado. Epsilon-nonparallel support vector regression. Applied Intelligence, 2019, 49(12):4223-4236.

\bibitem{QiTianShi12}
Qi Z, Tian Y, Shi Y. Laplacian twin support vector machine for semi-supervised classification. Neural networks, 2012, 35: 46-53.

\bibitem{ChenShaoDeng14}
Chen W J, Shao Y H, Deng N Y, et al. Laplacian least squares twin support vector machine for semi-supervised classification. Neurocomputing, 2014, 145: 465-476.



\bibitem{SunXie16}
Sun S, Xie X. Semisupervised support vector machines with tangent space intrinsic manifold regularization. IEEE Transactions on Neural Networks and Learning Systems, 2016, 27(9): 1827-1839.

\bibitem{YangXu16}
Yang Z, Xu Y. Laplacian twin parametric-margin support vector machine for semi-supervised classification. Neurocomputing, 2016, 171: 325-334.

\bibitem{RastogiPal19}
Rastogi R, Pal A. Fuzzy semi-supervised weighted linear loss twin support vector clustering. Knowledge-Based Systems, 2019, 165: 132-148.

\bibitem{WangShaoBai15}
Wang Z, Shao Y H, Bai L, et al. Twin support vector machine for clustering. IEEE Transactions on Neural Networks and Learning Systems, 2015, 26(10): 2583-2588.

\bibitem{KhemchandaniPal16}
Khemchandani R, Pal A. Weighted linear loss twin support vector clustering. Proceedings of the 3rd IKDD Conference on Data Science, ACM, 2016: 18.

\bibitem{YeZhaoNaiem17}
Ye Q, Zhao H, Naiem M. Fast robust twin support vector clustering. 2017 2nd International Conference on Applied Mechanics, Electronics and Mechatronics Engineering, 2017, 224-230.

\bibitem{YeZhaoLi18}
Ye Q, Zhao H, Li Z, et al. L1-norm distance minimization-based fast robust twin support vector $ k $-plane clustering. IEEE Transactions on Neural Networks and Learning Systems, 2018, 29(9): 4494-4503.

\bibitem{WangChenLi18}
Wang Z, Chen X, Shao Y H, et al. Ramp-based twin support vector clustering. Neural Computing and Applications, 2019, https://doi.org/10.1007/s00521-019-04511-3.

\bibitem{BaiShaoWang19}
Bai L, Shao Y H, Wang Z, et al. Clustering by twin support vector machine and least square twin support vector classifier with uniform output coding. Knowledge-Based Systems, 2019, 163: 227-240.

\bibitem{ChenShaoLi16}
Chen W J, Shao Y H, Li C N, et al. MLTSVM: a novel twin support vector machine to multi-label learning. Pattern Recognition, 2016, 52: 61-74.

\bibitem{Azad-Manjiri19}
Azad-Manjiri M, Amiri A, Sedghpour A S. ML-SLSTSVM: a new structural least square twin support vector machine for multi-label learning. Pattern Analysis and Applications, 2020, 23: 295-308.


\bibitem{KhemchandaniKarpatne13}
Khemchandani R, Karpatne A, Chandra S. Proximal support tensor machines. International Journal of Machine Learning and Cybernetics, 2013, 4(6): 703-712.

\bibitem{ZhaoShiLv14}
Zhao X, Shi H, Lv M, et al. Least squares twin support tensor machine for classification. Journal Of Information \& Computational Science, 2014, 11(12): 4175-4189.

\bibitem{ShiZhaoJing14}
Shi H F, Zhao X B, Jing L. Tensor distance based least square twin support tensor machine. Applied Mechanics and Materials. Trans Tech Publications, 2014, 668: 1170-1173.

\bibitem{ShiZhaoZhen16}
Shi H, Zhao X, Zhen L, et al. Twin bounded support tensor machine for classification. International Journal of Pattern Recognition and Artificial Intelligence, 2016, 30(01): 1650002.

\bibitem{XiangJiangHe18}
Xiang Y, Jiang Q, He J, et al. The advance of support tensor machine. 2018 IEEE 16th International Conference on Software Engineering Research, Management and Applications (SERA). IEEE, 2018: 121-128.

\bibitem{XieSun14}
Xie X, Sun S. Multi-view Laplacian twin support vector machines. Applied intelligence, 2014, 41(4): 1059-1068.


\bibitem{TangLiTian18}
Tang J, Li D, Tian Y, et al. Multi-view learning based on nonparallel support vector machine. Knowledge-Based Systems, 2018, 158: 94-108.

\bibitem{SunXieDong18}
Sun S, Xie X, Dong C. Multiview learning with generalized eigenvalue proximal support vector machines. IEEE Transactions on Cybernetics, 2018 (99): 1-10.

\bibitem{Xie18}
Xie X. Regularized multi-view least squares twin support vector machines. Applied Intelligence, 2018, 48(9): 3108-3115.

\bibitem{XieSun20}
Xie X , Sun S. General Multi-view Semi-supervised Least Squares Support Vector Machines with Multi-manifold Regularization. Information Fusion, 2020, 62: 63-72.

\bibitem{Kressel99}
Kressel U H G. Advances in Kernel methods. Cambridge, MA: MIT Press, 1999.


\bibitem{Vapnik98}
Vapnik V. Statistical learning theory. Wiley, New York, 1998.

\bibitem{PlattCristianini00}
Platt J C, Cristianini N, Shawe-Taylor J. Large margin DAGs for multiclass classification. Advances in Neural Information Processing Systems. 2000: 547-553.

\bibitem{CheongOh04}
Cheong S, Oh S H, Lee S Y. Support vector machines with binary tree architecture for multi-class classification. Neural Information Processing Letters and Reviews, 2004, 2(3): 47-51.

\bibitem{vTWSVM10}
Peng X. A $\nu$-twin support vector machine classifier and its geometric algorithms. Information Sciences, 2010, 180(20): 3863-3875.

\bibitem{TPMSM11}
Peng X. TPMSVM: a novel twin parametric-margin support vector machine for pattern recognition. Pattern Recognition, 2011, 44(10-11): 2678-2692.

\bibitem{CDMTSVM12}
Shao Y H, Deng N Y. A coordinate descent margin based-twin support vector machine for classification. Neural networks, 2012, 25: 114-121.


\bibitem{WLTSVM12}
Ye Q, Zhao C, Gao S, et al. Weighted twin support vector machines with local information and its application. Neural Networks, 2012, 35: 31-39.

\bibitem{WLSTSVM15}
Shao Y H, Chen W J, Wang Z, et al. Weighted linear loss twin support vector machine for large-scale classification. Knowledge-Based Systems, 2015, 73: 276-288.

\bibitem{SharmaRastogi19}
Sharma S, Rastogi R, Chandra S. Large-scale twin parametric support vector machine using pinball loss function. IEEE Transactions on Systems, Man, and Cybernetics: Systems, 2019. DOI: 10.1109/TSMC.2019.2896642.

\bibitem{NSVM2018}
Shao Y H, Yang K L, Liu M Z, et al. From support vector machine to nonparallel support vector machine. Operations Research Transactions, 2018, 22(2): 55-65. (In Chineses)

\bibitem{AVIPC2014}
Shao Y H, Li C N, Wang Z, et al. Proximal classifier via absolute value inequalities. 2014 IEEE International Conference on Data Mining Workshop. IEEE, 2014: 74-79.

\bibitem{FanLuXLi16}
Fan B, Lu X, Li H X. Probabilistic inference-based least squares support vector machine for modeling under noisy environment. IEEE Transactions on Systems, Man, and Cybernetics: Systems, 2016, 46(12): 1703-1710.

\bibitem{WangZhangChoi19}
Wang G, Zhang G, Choi K S, et al. Deep additive least squares support vector machines for classification with model transfer. IEEE Transactions on Systems, Man, and Cybernetics: Systems, 2019, 49(7): 1527-1540.
%


\bibitem{LPC}
Shao Y H, Chen W J, Wang Z, et al. A proximal classifier with positive and negative local regions. Neurocomputing, 2014, 145: 131-139.

\bibitem{LSPCC}
Shao Y H, Wang Z, Li C N, et al. Locality sensitive proximal classifier with consistency for small sample size problem. 2015 IEEE International Conference on Data Mining Workshop (ICDMW). IEEE, 2015: 1163-1170.

\bibitem{FTSVM}
Bai L, Wang Z, Shao Y H, et al. A novel feature selection method for twin support vector machine. Knowledge-Based Systems, 2014, 59: 1-8.

\bibitem{WenChenPong18}
Wen B, Chen X, Pong T K. A proximal difference-of-convex algorithm with extrapolation. Computational Optimization and Applications, 2018, 69(2): 297-324.

\bibitem{BeckTeboulle09}
Beck A, Teboulle M. A fast iterative shrinkage-thresholding algorithm for linear inverse problems. SIAM Journal on Imaging Sciences, 2009, 2: 183-202.

\bibitem{JohnNello04}
John S T, Nello C. Kernel methods for pattern analysis. U.K., Cambridge: Cambridge Univ. Press, 2004.

%
%
%




%


\bibitem{UCI98}
Blake C L, Merz C J. UCI repository of machine learning databases. http://archive.ics.uci.edu/ml/index.php, 1998.

\bibitem{Yale}
Georghiades A S, Belhumeur P N, Kriegman D J. From few to many: illumination cone models for face recognition under variable lighting and pose. IEEE Transactions on Pattern Analysis and Machine Intelligence, 2002, 23(6):643-660.

\bibitem{ORL}
Samaria F S, Harter A C. Harter A. Parameterisation of a stochastic model for human face identification. Proceedings of the Second IEEE Workshop on Applications of Computer Vision, IEEE, 1994: 138-142.

\end{thebibliography}
\end{document}